\documentclass[10pt]{article} % For LaTeX2e
\usepackage[preprint]{tmlr}

\usepackage{hyperref}
\usepackage{url}

\usepackage{amsmath}
\usepackage{amssymb}
\usepackage{amsfonts}
\usepackage{amsthm}
\usepackage{mathtools}
\usepackage{mathrsfs}
\usepackage{cases}
\usepackage{nicefrac}

\usepackage{graphicx}
\usepackage{subcaption}
\usepackage{caption}
\usepackage{booktabs}
\usepackage{multirow}
\usepackage{adjustbox}

\usepackage{tikz-cd}
\tikzcdset{scale cd/.style={every label/.append style={scale=#1},
    cells={nodes={scale=#1}}}}
\usepackage{pgfplots}
\pgfplotsset{compat=1.17}
\usepackage{extpfeil}

\usepackage{algorithm}
\usepackage{algorithmic}
\usepackage{etoolbox}
\AtBeginEnvironment{algorithmic}{\let\AND\relax}

\usepackage{enumitem}

\usepackage{xcolor}
\usepackage{microtype}
\usepackage{fmtcount}
\usepackage{listings}
\usepackage{comment}
\usepackage{filecontents}
\usepackage[most]{tcolorbox}
\usepackage{CJKutf8}
\usepackage{svg}
\usepackage{quiver}

\usepackage[capitalize,noabbrev]{cleveref}

\definecolor{myred}{rgb}{1, 0, 0}
\definecolor{mygreen}{rgb}{0, 0.5, 0}
\definecolor{myblue}{rgb}{0, 0, 1}

\definecolor{c1}{rgb}{1, 0, 0}
\definecolor{c2}{rgb}{0, 0.8, 0}
\definecolor{c3}{rgb}{0, 0, 1}
\definecolor{c4}{rgb}{0.5, 0, 0.5}

\definecolor{c11}{rgb}{1, 0, 0}
\definecolor{c12}{rgb}{0.8, 0.5, 0}
\definecolor{c13}{rgb}{0.8, 0.8, 0}
\definecolor{c14}{rgb}{0, 0.8, 0.4}
\definecolor{c15}{rgb}{0, 0, 1}
\definecolor{c16}{rgb}{0.5, 0, 0.5}

\definecolor{mycolor1}{rgb}{0.5, 0, 0}
\definecolor{mycolor2}{rgb}{1, 0.5, 0}
\definecolor{mycolor3}{rgb}{0.8, 0.8, 0}
\definecolor{mycolor4}{rgb}{0, 0.4, 0.2}
\definecolor{mycolor5}{rgb}{0, 0.8, 0.4}
\definecolor{mycolor6}{rgb}{0, 0.8, 0.8}

\definecolor{dc0}{rgb}{1, 0, 0}
\definecolor{dc1}{rgb}{0.5, 0, 0}
\definecolor{dc2}{rgb}{0.5, 0.5, 0}
\definecolor{dc3}{rgb}{0, 1, 0}
\definecolor{dc4}{rgb}{0, 0.5, 0}
\definecolor{dc5}{rgb}{0, 0.5, 0.5}
\definecolor{dc6}{rgb}{0, 0, 1}
\definecolor{dc7}{rgb}{0, 0, 0.5}
\definecolor{dc8}{rgb}{0.5, 0, 0.5}
\definecolor{dc9}{rgb}{0.5, 0.5, 0.5}
\definecolor{dc10}{rgb}{0.8, 0, 0}
\definecolor{dc11}{rgb}{0.6, 0.2, 0}
\definecolor{dc12}{rgb}{0.2, 0.6, 0}
\definecolor{dc13}{rgb}{0, 0.8, 0}
\definecolor{dc14}{rgb}{0, 0.6, 0.2}
\definecolor{dc15}{rgb}{0, 0.2, 0.6}
\definecolor{dc16}{rgb}{0, 0, 0.8}
\definecolor{dc17}{rgb}{0.2, 0, 0.6}
\definecolor{dc18}{rgb}{0.2, 0, 0.6}
\definecolor{dc19}{rgb}{0.2, 0.2, 0.2}

\definecolor{pc11}{rgb}{1, 0.7, 0.7}
\definecolor{pc12}{rgb}{1, 0.4, 0.4}
\definecolor{pc13}{rgb}{0.4, 0, 0}
\definecolor{pc21}{rgb}{0.1, 1, 0.1}
\definecolor{pc22}{rgb}{0, 0.8, 0}
\definecolor{pc23}{rgb}{0, 0.2, 0}
\definecolor{pc31}{rgb}{0.7, 0.7, 1}
\definecolor{pc32}{rgb}{0.4, 0.4, 1}
\definecolor{pc33}{rgb}{0, 0, 0.4}

\theoremstyle{plain}
\newtheorem{theorem}{Theorem}[section]
\newtheorem{proposition}[theorem]{Proposition}

\newtheorem{corollary}[theorem]{Corollary}
\theoremstyle{definition}
\newtheorem{definition}[theorem]{Definition}
\newtheorem{assumption}[theorem]{Assumption}
\theoremstyle{remark}

\shortcites{radford2019language,abdin2024phi3technicalreporthighly,qwen2025qwen25technicalreport}

\title{RL-STaR: Theoretical Analysis of Reinforcement Learning Frameworks for Self-Taught Reasoner}

\author{\name Fu-Chieh Chang\thanks{Equal contribution.} \email fc-mark.chang@mediatek.com \\
      \addr MediaTek Inc., Hsinchu, Taiwan
      \AND
      \name Yu-Ting Lee\footnotemark[1] \email r14942088@ntu.edu.tw \\
      \addr Graduate Institute of Communication Engineering, National Taiwan University, Taipei, Taiwan
      \AND
      \name Hui-Ying Shih \email hys114061583@gapp.nthu.edu.tw \\
      \addr Department of Electrical Engineering, National Tsing Hua University, Hsinchu, Taiwan
      \AND
      \name Yi Hsuan Tseng \email r12227115@ntu.edu.tw \\
      \addr Department of Psychology, National Taiwan University, Taipei, Taiwan
      \AND
      \name Pei-Yuan Wu \email peiyuanwu@ntu.edu.tw \\
      \addr Graduate Institute of Communication Engineering, National Taiwan University, Taipei, Taiwan}

\def\month{MM}  % Insert correct month for camera-ready version
\def\year{YYYY} % Insert correct year for camera-ready version
\def\openreview{\url{https://openreview.net/forum?id=XXXX}} % Insert correct link to OpenReview for camera-ready version

\begin{document}

\maketitle

\begin{abstract}
The reasoning abilities of large language models (LLMs) have improved with chain-of-thought (CoT) prompting, allowing models to solve complex tasks stepwise. However, training CoT capabilities requires detailed reasoning data, which is often scarce. The self-taught reasoner (STaR) framework addresses this by using reinforcement learning to automatically generate reasoning steps, reducing reliance on human-labeled data. Although STaR and its variants have demonstrated empirical success, a theoretical foundation explaining these improvements is lacking. This work provides a theoretical framework for understanding the effectiveness of reinforcement learning on CoT reasoning and STaR. Our contributions are: (1) criteria for the quality of pre-trained models necessary to initiate effective reasoning improvement; (2) an analysis of policy improvement, showing why LLM reasoning improves iteratively with STaR; (3) conditions for convergence to an optimal reasoning policy; and (4) an examination of STaR's robustness, explaining how it can improve reasoning even when incorporating occasional incorrect steps. We also run RL-STaR on GPT-2, Qwen2.5-0.5B and Phi-3-mini, and the measured return curves follow the ones the analysis predicts. This framework bridges empirical findings with theoretical insights, advancing reinforcement learning approaches for reasoning in LLMs.
\end{abstract}

\section{Introduction}
With the advancement of large language models (LLMs), their reasoning capabilities have become crucial to their success. This progress is mainly attributed to chain-of-thought (CoT) prompting~\citep{wei2022chain}, which allows LLMs to go beyond pattern matching and handle more complex reasoning problems by providing step-by-step guidance. GPT4-o1~\citep{chatgpt2024} exemplifies this success, achieving high scores on various mathematical and programming benchmarks.
However, to train models with CoT capabilities, training data must include detailed reasoning steps~\citep{malach2023auto,prystawski2024think,xiao2024theory}, which are often absent. To address this challenge, the self-taught reasoner (STaR) approach~\citep{zelikman2022star} was proposed, leveraging reinforcement learning to automatically discover reasoning steps. Numerous improvements to STaR have since been introduced~\citep{hosseini2024v,zelikman2024quiet,lin2024lean,andukuri2024star,xiang2025towards}, demonstrating empirically that LLMs can effectively learn reasoning steps via reinforcement learning without human intervention.
Although some theoretical research exists on CoT techniques (e.g.,~\citep{prystawski2024think,malach2023auto,feng2024towards,xiao2024theory,hu2024unveilingstatisticalfoundationschainofthought}), these studies are primarily focused on supervised and auto-regressive learning settings that require detailed reasoning steps included in training data. They do not show how reinforcement learning techniques can enhance reasoning steps. Furthermore, while there are existing reinforcement learning frameworks for theoretical analysis (e.g.,~\citep{jin2018sample,ayoub2020model,jin2021pessimism,jin2019linear,bhandari2021linear,chen2022human,yeh2023sample,lai2024leveraging}), none are designed to analyze the self-improvement of LLMs through reinforcement learning. As a result, no theoretical framework explains how LLMs can enhance their reasoning capabilities via reinforcement learning.  A detailed literature review is given in Sec.~\ref{sec:related_works}.

\subsection{Our Contributions}
\label{sec:contribution}
In this work, we develop a theoretical framework to analyze how the Self-Taught Reasoner (STaR) enhances Chain-of-Thought (CoT) reasoning under a simplified setup such as Markov properties (see Sec.~\ref{sec:more_about_limitation_non_markov}). In this context, we address the following questions:
\begin{enumerate}
    \item \textbf{Conditions of Pre-trained Models for STaR: \emph{How competent does the pre-trained LLM need to be to bootstrap the discovery of reasoning steps in the first iteration?}}
    We show that a pre-trained LLM can initiate effective reasoning improvement if one of the following holds when inferencing on the problems in STaR’s training set.
    \begin{itemize}
    \item The pre-trained LLM performs better than a randomly initialized model at each reasoning step (i.e., its probability of producing the correct step exceeds that of random guessing).
    \item The pre-trained LLM is on par with a randomly initialized model at exactly one reasoning step but exceeds it at all other steps.    
    \end{itemize}
    \item \textbf{Policy Improvement: \emph{Can LLMs improve their reasoning capabilities iteratively through STaR?}}  
    We demonstrate that if the pre-trained model satisfies the conditions we mentioned previously, within each iteration of the STaR algorithm, LLMs can consistently improve the correctness of their reasoning trajectories.  
    \item \textbf{Convergence to the Optimal Policy: \emph{If an optimal reasoning model exists, can STaR eventually find this optimal reasoner?} }
    Given sufficient iterations, we prove that if the pre-trained model satisfies the previously mentioned conditions, LLMs can converge to the optimal reasoner, achieving the highest probability of generating correct reasoning trajectories that lead to correct answers.
    
    \item \textbf{Existence of Incorrect Reasoning Steps in STaR: \emph{Is it possible for a sub-optimal model to generate incorrect reasoning steps that are included in the next iteration of training, while still arriving at the correct final answer?}}
    We show that even when incorrect reasoning steps are included in the training data for a given iteration, the probability of these incorrect reasoning steps being included in the training data will diminish as the number of STaR iterations increases.

\end{enumerate}
Although our theoretical setup simplifies real-world LLM settings, we also discuss how our theoretical results could be applied to more realistic scenarios. In Sec.~\ref{sec:comparison_pg} we explain why the convergence guarantees of policy-gradient methods do not already answer these questions.
%of why LLMs could fix their wrong reasoning path given the CoT corpus, where only the final answers are annotated.

% \section{Related Works}
% We review existing literature on reinforcement learning and chain-of-thought, covering both theoretical and practical aspects. We highlight the contributions and limitations of prior work in comparison to our own contributions. \todo{} The detailed literature review is shown in Sec.~\ref{sec:related_works}

\section{Related Works}\label{sec:related_works}
\subsection{Approaches to Improve Chain-of-Thought}

Chain-of-Thought (CoT) prompting improves reasoning by encouraging models to generate intermediate steps \citep{wei2022chain}. However, scaling model size alone does not strengthen causal understanding \cite{mimichumanBao}, motivating research on selecting effective CoT prompts and correcting flawed reasoning traces (e.g., \cite{PedCOT24}). Recent work has explored multiple strategies for enhancing CoT. Retrieval-Q-CoT \citep{AutoCOTpromting} selects prompts by grouping questions via cosine similarity, while Process Supervision \citep{lightman2023let} trains reward models that stabilize step-by-step reasoning. Other studies focus on discovering optimal instructions \citep{humanlevelprompt} or improving Zero-Shot CoT through automated prompting strategies \citep{planandsolveLei}. Importantly, \citet{cannotselfcorrect} shows that external feedback is critical for self-correction, highlighting the need for explicit feedback-driven mechanisms. Overall, these approaches illustrate a growing effort to refine CoT prompting, enhance reasoning robustness, and reduce reliance on model scale alone.
% \todo{\citep{teng2025atomthoughtsmarkovllm}}

\subsection{Self-Taught Reasoner (STaR) for Boosting Chain-of-Thought}

The Self-Taught Reasoner (STaR) framework \citep{zelikman2022star} improves LLM reasoning by using reinforcement learning to iteratively refine self-generated chain-of-thought (CoT) traces. STaR first produces reasoning steps via in-context learning, retains only those leading to correct answers, and continually expands its training set with successful CoTs. Numerous extensions enhance this paradigm: Quiet-STaR \citep{zelikman2024quiet} adds token-level rationales; V-STaR \citep{hosseini2024v} trains a verifier using DPO to assess both correct and incorrect solutions; Lean-STaR \citep{lin2024lean} introduces informal pre-proof reasoning; STaR-GATE \citep{andukuri2024star} rewards insightful question generation; and Meta-STaR \citep{xiang2025towards} incorporates meta-cognitive CoT. Other variants include B-STaR \citep{zeng2025bstarmonitoringbalancingexploration}, which balances exploration and exploitation, and rStar-Math \citep{guan2025rstarmathsmallllmsmaster}, showing strong reasoning gains even for small models. Despite these advances, the theoretical basis for why STaR enables autonomous reasoning improvement remains largely unexplored.

\subsection{Theories of Reinforcement Learning}

Reinforcement learning theory aims to explain how algorithms iteratively improve a policy toward optimality. Foundational results on Tabular Q-learning, such as \citep{jin2018sample}, establish polynomial-time convergence, with extensions to settings involving linear reward and transition structures \citep{jin2019linear,yang2019sample,he2021logarithmic} and kernel-based function approximation \citep{yang2020function,yeh2023sample}. Additional analyses characterize convergence for policy gradient methods \citep{6796286,bhandari2021linear,bhandari2024global}, human-in-the-loop RL \citep{chen2022human,kong2022provably}, model-based RL \citep{osband2014model,ayoub2020model}, and offline RL \citep{hu2023provable,jin2021pessimism,lai2024leveraging}. Collectively, these works provide deep insight into classical RL algorithms and their sample-efficiency guarantees. However, they do not directly address the distinct challenges involved in improving large language model reasoning—where policies correspond to multi-step generative processes rather than state-action mappings. This gap motivates the need for new theoretical frameworks tailored to RL-driven enhancement of chain-of-thought reasoning in LLMs.

\subsection{Theories of Chain-of-Thought}
Chain-of-Thought (CoT) prompting~\citep{wei2022chain} enables LLMs to solve complex tasks by decomposing solutions into stepwise reasoning. Several theoretical works attempt to explain its effectiveness. \citet{prystawski2024think} models CoT as Bayesian networks linking questions, answers, and intermediate steps, while \citet{xiao2024theory} introduces length generalization, showing that models can extend simple patterns to harder cases. In a PAC auto-regressive framework, \citet{malach2023auto} proves that CoT allows learning linear threshold circuits. CoT also increases computational reach: \citet{feng2024towards} demonstrates transformer ability to emulate dynamic programming; \citet{li2024chainthoughtempowerstransformers} and \citet{kim2024transformersprovablysolveparity} show enhanced expressiveness via sequential intermediate states. Additional analyses address sample complexity~\citep{hu2024unveilingstatisticalfoundationschainofthought,li2024trainingnonlineartransformerschainofthought}, OOD generalization~\citep{yao2025unveilingmechanismsexplicitcot}, and the role of positional embeddings~\citep{wang2025indistributionsuccessscalingcurves}. While these studies form the theoretical basis of CoT, they do not explain why reinforcement learning further improves CoT, nor how to reduce reliance on extensive reasoning-step supervision.

% \todo{\citep{yao2025unveilingmechanismsexplicitcot}, \citep{wang2025indistributionsuccessscalingcurves},
% \citep{li2024trainingnonlineartransformerschainofthought}
% }
\section{Theoretical Framework}
\subsection{Problem Formulation}
\label{sec:problem_settings}
In our problem formulation, we consider a chain of thought (CoT) reasoning process composed of $N $ steps where $N>1$. Let $ s_0 $ denote the initial input string, and $ s_n $ represent the resulting string after the $ n $-th CoT step, where $ 1 \leq n \leq N $. We assume that the chain-of-thought steps satisfy the Markov property: each string $ s_n $ contains sufficient information to derive the next string $ s_{n+1} $, without relying on information from any of the preceding strings $ s_{i}$ where $0\leq i<n$. For instance, in the addition problem $ 1+2+3+4 $, the chain-of-thought steps can be expressed as:
$$
\begin{aligned}
& s_0 = \texttt{1+2+3+4} \Rightarrow s_1 = \texttt{3+3+4} ~\Rightarrow s_2 = \texttt{6+4} \Rightarrow s_3 = \texttt{10}.
\end{aligned}
$$
In this example, obtaining $ s_2 = \texttt{6+4} $ requires only the information in $ s_1 = \texttt{3+3+4} $ and does not depend on $ s_0 = \texttt{1+2+3+4} $. 
Under this Markov assumption, the chain-of-thought (CoT) process can naturally be cast as a Reinforcement Learning (RL) problem, which can be described by a Markov decision process (MDP). Formally, let \(\mathcal{M} = (\mathcal{S}, \mathcal{A}, N, r, \mathcal{P})\) represent the MDP, where:
\begin{itemize}
\item \(\mathcal{S}\) is the space of all possible initial inputs, CoT steps, and  final answers, denoted \(\{s_i \mid 0 \leq i \leq N\}\). In this formulation, \(s_0\) is the CoT input,  \(s_1, \ldots, s_{N-1}\) are the subsequent reasoning steps, and $s_N$ is the final answer.
\item \(\mathcal{A}\) is the action space. Because each action corresponds uniquely to selecting the next state, we identify \(\mathcal{A}\) with \(\mathcal{S}\). That is, choosing \(a\) is equivalent to setting \(s_{n+1}\) as the unique next step associated with \(a\).
\item \(\mathcal{P}\) is the transition function. Given that an action \(a\) uniquely determines the next state \(s_{n+1}\), this transition is deterministic:
  \[
\mathcal{P}\bigl(S_{n+1} = s_{n+1} \big| A = s_{n+1}, S_n = s_n\bigr) = 1.
  \]
\item  \(r(s_0,s_N)\) is the reward function, yielding a nonzero reward solely at the final step if \(s_N\) matches the correct answer. Concretely, consider a dataset \(\mathcal{D}\) composed of question-answer pairs. For each instance \((s_0, s_N^\star) \in \mathcal{D}\), where \(s_0\) represents the question and \(s_N^\star\) the correct final answer, define
\[
r(s_0,s_N) =
\begin{cases}
1, ~\text{if  $s_N = s_N^\star$ and $(s_0,s_N^{\star})\in\mathcal{D}$},\\
0, ~ \text{otherwise}.
\end{cases}
\]
Thus, the agent earns a reward of 1 exclusively when it terminates on \(s_N^\star\), and 0 otherwise. Note that we fixed the number of CoT steps to be \(N\); intermediate states \(s_n\) (for \(1 \le n < N\)) do not produce any positive reward.
\end{itemize}

Given the RL formulation above, a policy \(\pi(A \mid S_n = s_n)\) specifies how the next action \(A = s_{n+1}\) is selected based on the current state \(S_n = s_n\). Although the transition \(\mathcal{P}\bigl(S_{n+1} \mid A = s_{n+1}, S_n = s_n\bigr)\) is deterministic (that is, once \(A = s_{n+1}\) is chosen, \(S_{n+1}=s_{n+1}\) is uniquely determined), the policy \(\pi(A \mid S_n = s_n)\) itself can be stochastic if there is uncertainty about which next step to choose given \(S_n=s_n\).
To streamline notations, we define a stochastic transition \(P(S_{n+1} \mid S_n)\) by combining the stochastic policy \(\pi(A \mid S_n)\) with the deterministic transition \(\mathcal{P}(S_{n+1} \mid A, S_n)\):
\[
P\bigl(S_{n+1} \mid S_n\bigr) \;=\; \mathcal{P}\bigl(S_{n+1} \mid \pi(A \mid S_n), S_n\bigr).
\]
In this setup, the LLM serves as the transition function \(P(S_{n+1} \mid S_n)\), producing the subsequent CoT step \(s_{n+1}\) based on the current CoT step \(s_n\). After the final step \(s_N\) is reached, the reward function \(r(s_0,s_N)\) measures correctness by comparing \(s_N\) to the ground truth answer \(s_N^\star\).

% Given a dataset of reasoning problems $\mathcal{D}$, which consists of pairs of questions and corresponding answers, we represent each problem as a pair $\left(s_0, s_N^{\star}\right) \sim \mathcal{D}$, where $s_0$ is the input question and $s_N^{\star}$ is the ground truth answer. The agent, using policy $\pi$, generates a final state $s_N = \operatorname{RL-CoT}(s_0, \pi)$. The agent receives a reward of one if the generated final state matches the correct answer, i.e., $s_N = s_N^{\star}$; otherwise, the reward is zero. We define an objective function $J$ for policy $\pi$ to measure the expected accuracy in obtaining the true answer:
% $$
% J(\pi) = \mathbb{E}_{\left(s_0, s_N^{\star}\right) \sim \mathcal{D} \text{ and } s_N \sim \operatorname{RL-CoT}(s_0, \pi)} \mathbb{I}[s_N = s_N^{\star}].
% $$

\begin{algorithm}[h!]
\caption{RL-STaR}
\begin{algorithmic}
\STATE \textbf{Input:} A dataset $\mathcal{D}=\{(s_0^{(k)},s_N^{\star(k)}) | k \in [K]\}$, a pre-trained LLM as transition $P_0$.
\STATE \textbf{Output:} A trained LLM as transition $P_T$.
\FOR{$t = 1$ to $T$} 
%\STATE $\mathcal{D}_t \leftarrow 	\varnothing$ $\#$ Initialize $\mathcal{D}_t$ as an empty array.
\STATE \textcolor{gray}{$\#$ Repeat the following procedure for $L$ times where $L \gg K$.}
\FOR{$\ell = 1$ to $L$} 
\STATE $\left(s_0^{(\ell)} s_N^{\star(\ell)}\right) \sim \mathcal{D}$. \textcolor{gray}{$\#$ Uniformly sample a pair $\left(s_0^{(\ell)}, s_N^{\star(\ell)}\right)$ from $\mathcal{D}$. }
% \STATE $\left( s_N^{(\ell)}, \tau^{(\ell)}\right) \leftarrow \operatorname{RL-CoT}\left(s_0^{(\ell)}, \pi_{t-1}\right)$. $\#$ Run RL-CoT to get the final state $s_N$ and the trajectory $\tau$.
\STATE $\tau^{(\ell)} \leftarrow (s_0^{(\ell)},)$ \textcolor{gray}{$\#$ Set $s_0^{(\ell)}$ as the initial state of the trajectory $\tau^{(\ell)}$.}
\FOR{$n = 1$ to $N$}  
\STATE $s_n^{(\ell)} \sim P_{t-1}(S_{n}| S_{n-1} = s_{n-1}^{(\ell)})$ \textcolor{gray}{$\#$ Randomly sample $s_n^{(\ell)}$ with probability $P_{t-1}(S_{n} = s_{n}^{(\ell)}| S_{n-1} = s_{n-1}^{(\ell)})$.}
\STATE $\tau^{(\ell)} \leftarrow (\tau^{(\ell)}_{1:n-1},s_n^{(\ell)})$ \textcolor{gray}{$\#$ Append $s_n^{(\ell)}$ to $\tau^{(\ell)}$.}
\ENDFOR
\ENDFOR
\STATE $\mathcal{D}_t \leftarrow \left\{ \tau^{(\ell)} \mid \ell \in [L] \text{ and } s_N^{(\ell)} = s_N^{\star(\ell)}\right\}$ \textcolor{gray}{$\#$ Add the trajectories whose final state $s_N^{(\ell)}$ matches $s_N^{\star(\ell)}$ to $\mathcal{D}_t$.}
\STATE $P_t \leftarrow \operatorname{Train}\left(P_{t-1}, \mathcal{D}_t \right)$ \textcolor{gray}{$\#$ Use $\mathcal{D}_t$ to update the transition.}
\ENDFOR
\end{algorithmic}
\label{algo:rl_star}
\end{algorithm}
To guide LLMs toward selecting a final state \(s_N\) that maximizes the reward upon completing CoT, we use a modified version of STaR~\citep{zelikman2022star}, termed RL-STaR, as outlined in Algorithm \ref{algo:rl_star}.
This algorithm takes a training dataset $\mathcal{D}$ and a pre-trained LLM as transition $P_0$ as input, and outputs a trained LLM as transition $P_T$, where $\mathcal{D}$  consists of $K$ instances.  
In each iteration \(t\), we repeat the following procedure \(L\) times: we uniformly sample a pair \(\bigl(s_0^{(\ell)}, s_N^{\star(\ell)}\bigr)\) from \(\mathcal{D}\) and generate a trajectory \(\tau^{(\ell)}\) by sequentially sampling states from the current transition \(P_{t-1}\). Specifically, starting from \(s_0^{(\ell)}\), we iteratively draw \(s_n^{(\ell)}\sim P_{t-1}\bigl(S_n\!\mid\!S_{n-1}=s_{n-1}^{(\ell)}\bigr)\) for \(n=1,\dots,N\). If the final state \(s_N^{(\ell)}\) matches the ground truth \(s_N^{\star(\ell)}\), we add the entire trajectory \(\tau^{(\ell)}\) to \(\mathcal{D}_t\). After completing these \(L\) samplings, we update \(P_{t-1}\) to \(P_t\) by retraining on \(\mathcal{D}_t\). In particular, \(\operatorname{Train}(P_{t-1}, \mathcal{D}_t)\) adjusts the transition probabilities \(P_{t-1}(S_n=s_n\!\mid\!S_{n-1}=s_{n-1})\) to better align with the transitions observed in the successful trajectories within \(\mathcal{D}_t\).  After \(T\) iterations, this procedure outputs the final transition model \(P_T\).
%In Sec.~\ref{sec:theoretical_results}, we assume that the trained LLM can perfectly match this transition $P(S_n = s_n | S_{n-1} = s_{n-1})$.
In Sec.~\ref{sec:theoretical_results}, we show that RL-STaR aims to maximize the total expected return \(J(P_t)\), defined by
$$
\begin{aligned}
&J(P_t) = \mathbb{E}_{(s_0,s_N^{\star})\sim \mathcal{D}}\mathbb{E}_{(s_1,\cdots,s_N)\sim P_t(\tau|S_0=s_0) }r(s_0,s_N), \\
&\quad\text{where}~ P_t(\tau|S_0=s_0) = P_t(S_1=s_1|S_0=s_0) \times (\Pi_{n=2}^{N}P_t(S_n=s_n|S_{n-1}=s_{n-1})). 
\end{aligned}
$$
% To demonstrate this, we use the following setup.

\subsection{Settings of Our Theoretical Analysis}
\label{sec:implementation_details}
\paragraph{Markov Assumption of Language Models' Input}
For simplicity in analyzing the RL-STaR algorithm, we impose the Markov assumption by allowing the LLM to accept only the current step \(s_{n-1}\) as input, rather than the complete history \(\bigl(s_0, s_1, \dots, s_{n-2}\bigr)\). This contrasts with standard CoT methods, which include every previous token in the context. Nevertheless, the Markov property remains essential for CoT generalization. Several studies (e.g., \citep{ yao2025unveilingmechanismsexplicitcot, teng2025atomthoughtsmarkovllm, wang2025indistributionsuccessscalingcurves, li2024trainingnonlineartransformerschainofthought,  xiao2024theory,prystawski2024think}) show that CoT succeeds by breaking complex tasks into simpler, reusable subtasks whose outcomes depend only on a limited recent context, thus preserving Markovian structure.
The details are discussed in Sec.~\ref{sec:more_about_limitation_non_markov}.

\paragraph{Simplification of STaR Algorithm}
For simplicity in our theoretical analysis, we exclude the \emph{rationalization} from STaR—that is, we do not correct an LLM’s incorrect outputs using hints derived from the final answer. As noted in \citep{zelikman2022star}, omitting rationalization may reduce STaR’s convergence rate. However, based on their result, rationalization is not a necessary condition for policy improvement.
Hence, we accept this trade-off here since our focus lies on preliminary theoretical examination rather than achieving state-of-the-art performance on reasoning benchmarks.
% This setting allows us to analyze the transitions between each reasoning step in the CoT process. We assume that, for each reasoning problem in the dataset $\mathcal{D}$, there exists a ground-truth reasoner $\bar{\pi}(A|S)$ that can output the sequence of reasoning steps $s_n$ for $1 \leq n \leq N$, leading to the correct answer $s_N^{\star}$ given the initial input $s_0$. In our setup, the output of ground-truth policy $\bar{\pi}(A|S_{n-1} = s_{n-1})$ depends solely on $s_{n-1}$, without relying on previous states $s_{n-2}, \dots, s_0$. This property enables us to use reinforcement learning to approximate $\bar{\pi}(A|S)$ with a policy $\pi(A|S)$. 

\paragraph{Assumption of the Ground-Truth Reasoner $\bar{\pi}$}
We assume that a ground-truth transition \(\bar{P}(S_{n+1} |S_{n})\) exists, which produces the correct sequence \(s_1, \ldots, s_{N-1}\) leading to \(s_N=s_N^\star\) given $s_0$ for every instance $(s_0,s_N^{\star})$ in \(\mathcal{D}\). This property enables us to apply the analysis of RL-STaR which approximates $\bar{P}(S_{n+1}|S_{n})$ with an estimated transition $P_T(s_{n+1}|S_{n})$. It is clear that if an estimated transition $P_T$ can perfectly match the ground-truth reasoning step transitions $\bar{P}$, it would be the optimal estimated transition $P^{\star}$ which maximizes $J(P^{\star})$, 
namely,
$$
\begin{aligned}
& P^{\star}(S_{n+1} =s_{n+1}|S_{n}=s_n) = \bar{P}(S_{n+1}=s_{n+1}|S_{n}=s_n), \\ 
& \text{\quad for all $(s_{n+1},s_n) \in \operatorname{support}{(S_{n+1})} \times \operatorname{support}{(S_{n})}$,\, where } n\in \{0,1,\dots,N-1\}.
\end{aligned}
$$

% Since the transition function $P(S_{n+1}|A, S_n = s_n)$ is essentially an identity function when $A = s_{n+1}$, we can merge the transition $P(S_{n+1}|A, S_n)$ with the ground-truth policy $\bar{\pi}(A|S_n)$. This defines the transition of the ground-truth reasoning steps as $\bar{P}(S_{n+1}|S_n) \equiv P\left(S_{n+1} | \bar{\pi}(A|S_n), S_n\right)$. Similarly, we denote the estimated transition by LLMs as $\hat{P}(S_{n+1}|S_n) \equiv P\left(S_{n+1}|\pi(A|S_n), S_n\right)$, which is trained to approximate the ground-truth reasoning step transition $\bar{P}(S_{n+1}|S_n)$.

% \paragraph{Pre-training:}  
% For pre-training, we create an independent dataset from $\mathcal{D}$ by sampling pairs $\left(s_0^{\star}, s_N^{\star}\right) \sim \mathcal{D}$. For each $(s_0^{\star}, s_N^{\star})$, we use a golden simulator or manual labeling to construct intermediate steps $s_1^{\star}, s_2^{\star}, \dots, s_{N-1}^{\star}$, forming the pre-training dataset $\mathcal{D}_{\text{pre-train}}$ with tuples $(s_0^{\star}, s_1^{\star}, s_2^{\star}, \dots, s_{N-1}^{\star}, s_N^{\star})$, where the superscript $\star$ indicates steps generated by a golden simulator or manual labeling. Similar to training, we first sample a random step $n$ from a discrete uniform distribution over $\{1, 2, \dots, N\}$, and the LLM is trained to output $s_n^{\star}$ when given $s_{n-1}^{\star}$ as input. 
%In Sec.~\ref{sec:theoretical_results}, we assume that after pre-training, the LLMs can perform state transitions with accuracy exceeding that of a uniform distribution.

\section{Theoretical Results}
\label{sec:theoretical_results}
In this section, we present our theoretical analysis addressing the questions outlined in Sec.~\ref{sec:contribution}. We outline conditions under which the RL-STaR algorithm can effectively train an LLM to approximate the optimal estimated transition $P^{\star}$. 
\subsection{Notations}
For clarity, the notations used in the main text are presented below. The additional notation that appears only in the proofs is collected in~\ref{sec:appendix_notation}.
\label{sec:notations}
% Before presenting the additional theorems and proofs, we define the following notation for clarity:
% \begin{table}[htbp]
% \centering
% \caption{Notations \todo{}}\label{table:Notations}
% \hspace{0.3cm}
% \begin{tabular}{|p{1.8cm}|p{5.4cm}|}
% \hline
% \textbf{Notation} & \textbf{Description} \\ 
% \hline
% \(T\) & The number of iterations at RL-STaR. \\ \hline
% \(N\) & The number of CoT steps. \\ \hline
% \(M\) & The number of states at each CoT step. \\ \hline
% \(s_{n, m}\) & The \(m\)-th state at the \(n\)-th CoT step. \\ \hline
% \(s_{n, m' \neq m}\) & A state \(s_{n, m'}\) where \(m' \neq m\) for some \(m' \in [M]\). \\ \hline
% \(\operatorname{support}(S)\) & The support of a random variable \(S\). \\ \hline
% \([n]\) & The set \(\{1, 2, \dots, n\}\). \\ \hline
% \(\tau = (a, b, c)\) & An ordered set containing elements \(a, b, c\) sequentially. \\ \hline
% \(\|P\|_{\infty}\) & The maximum element in the matrix \(P\). \\ \hline
% \((s_i, s_k) \Subset \tau\) & Indicates that \(s_i\) and \(s_k\) are both in the ordered set 
% \(\tau = (s_i, s_j, \dots, s_k, s_l)\), with \(s_i\) preceding \(s_k\). \\ 
% \hline
% \end{tabular}
% \end{table}
% \todo{}

\begin{itemize}[itemsep=1pt, topsep=2pt, parsep=0pt]
\item $T$, $N$, $M$: the number of RL-STaR iterations, the number of CoT steps, and the number of states at each CoT step.
\item $s_{n, m}$: the $m$-th state at the $n$-th CoT step, and $s_{n, m' \neq m}$: any state at the $n$-th CoT step other than $s_{n,m}$.
\item $[n]$: the set $\{1, 2, \dots, n\}$, and $\operatorname{support}(S)$: the support of a random variable $S$.
\item $\tau = (s_0, s_1, \dots, s_N)$: a trajectory, written as a tuple whose entries appear in the order listed.
\item $[P]_{i,j}$: the element with index $(i,j)$ in the matrix $P$; $I_M$: the $M \times M$ identity matrix; and $e_1$: the first standard basis vector.
\item $\|P\|_{\infty}$: the induced $\infty$-norm of the matrix $P$, that is, $\|P\|_{\infty} = \max_{i}\sum_{j}\left|[P]_{i,j}\right|$.
\item $\lceil x \rceil_{+}$: $\operatorname{min}(\{n \mid n \in \mathbb{N} ~\text{and}~ n \geq x\})$, and $\lfloor x \rfloor_{+}$: $\operatorname{max}(\{n \mid n \in \mathbb{N} ~\text{and}~ n \leq x\})$, where $\mathbb{N}=\{0,1,2,\dots\}$.
\item $\mathbb{I}\{a=b\}$ : $\mathbb{I}\{a=b\}=1$ if $a=b$; otherwise $\mathbb{I}\{a=b\}=0$.
% \item  $P_0$: The estimated transition from a pre-trained LLM.
% \item  $P_t$: The estimated transition from the LLM after the $t$-th iteration of RL-STaR.
\end{itemize}

\subsection{The Family $P_n[\delta]$ and Standing Assumptions}
\label{sec:delta_family}
Throughout Sec.~\ref{sec:theoretical_results}, a CoT step $n$ has $M$ states. We write $\bar{P}_n$ for the ground-truth transition at that step, which puts all of its mass on the correct next state, and $P_{u,n}$ for the uniform transition, which puts mass $\frac{1}{M}$ on every state.

\begin{definition}[$\delta$-accurate transition]
\label{def:delta_family}
For $\delta \in [0,1)$, let
$$
P_n[\delta] \;:=\; \delta\,\bar{P}_n + (1-\delta)\,P_{u,n},
$$
so that $P_n[\delta]$ assigns probability $\frac{1+(M-1)\delta}{M}$ to the correct next step and $\frac{1-\delta}{M}$ to each of the $M-1$ incorrect ones. Hence $P_n[0]=P_{u,n}$ is the uniform model, and $P_n[\delta]$ approaches $\bar{P}_n$ as $\delta \rightarrow 1$, so $\delta$ measures how far the model has moved from random guessing toward the ground-truth reasoner.
\end{definition}

The results in this section rest on two assumptions about RL-STaR.

\begin{assumption}[Training set]
\label{assump:dataset}
$\mathcal{D} = \{(s_{0,m}, s_{N,m}) \mid m \in [M]\}$, that is, the training set holds one question--answer pair per ground-truth trajectory.
\end{assumption}

\begin{assumption}[Exact fitting]
\label{assump:exact_fit}
$\operatorname{Train}(P_{t-1},\mathcal{D}_t)$ returns the transition that matches the conditional distribution of $\mathcal{D}_t$ exactly. That is, for every step $n$, $P_{t,n}(S_n \mid S_{n-1})$ equals the conditional probability of observing $(s_{n-1,m},s_{n,m})$ and $(s_{n-1,m},s_{n,m'\neq m})$ among the trajectories $\tau \sim \mathcal{D}_t$.
\end{assumption}

\subsection{A Toy Example}  
\label{sec:toy_example}  
We first illustrate our theoretical results with a toy example. In this scenario, we consider a CoT process with two reasoning steps (i.e., $N = 2$), and each step has two possible states (i.e., $M = 2$). Here, $S_0$ is a random variable representing the initial state, $S_1$ the intermediate state, and $S_2$ the final state. We assume their supports are $\operatorname{support}(S_0) = \{s_{0,1}, s_{0,2}\}$, $\operatorname{support}(S_1) = \{s_{1,1}, s_{1,2}\}$, and $\operatorname{support}(S_2) = \{s_{2,1}, s_{2,2}\}$. The ground-truth reasoning trajectories are defined as $\tau_0^{\star} = (s_{0,1}, s_{1,1}, s_{2,1})$ and $\tau_1^{\star} = (s_{0,2}, s_{1,2}, s_{2,2})$, giving the ground-truth transition $\bar{P}_n(S_n | S_{n-1})$ at step $1\leq n \leq 2$ as
$$
\bar{P}_n(S_n | S_{n-1}) =
\begin{cases}
1 & \text{if } (S_n,S_{n-1}) = (s_{n,m},s_{n-1,m}) \text{ for all } n, m \in [2], \\
0 & \text{otherwise}.
\end{cases}
$$
% The transition $\bar{P}(S_n|S_{n-1})$ can be illustrated as
% $$
% \begin{tikzcd}
% 	{s_{0,1}} &&& {s_{1,1}} &&& {s_{2,1}} \\
% 	\\
% 	\\
% 	{s_{0,2}} &&& {s_{1,2}} &&& {s_{2,2}}
% 	\arrow["{{\bar{P}(s_{1,1}|s_{0,1})=1}}"{description}, from=1-1, to=1-4]
% 	\arrow["{{\bar{P}(s_{1,2}|s_{0,1})=0}}"{description, pos=0.7}, from=1-1, to=4-4]
% 	\arrow["{{\bar{P}(s_{2,1}|s_{1,1})=1}}"{description}, from=1-4, to=1-7]
% 	\arrow["{{\bar{P}(s_{2,2}|s_{1,1})=0}}"{description, pos=0.7}, from=1-4, to=4-7]
% 	\arrow["{{\bar{P}(s_{1,1}|s_{0,2})=0}}"{description, pos=0.7}, from=4-1, to=1-4]
% 	\arrow["{{\bar{P}(s_{1,2}|s_{0,2})=1}}"{description}, from=4-1, to=4-4]
% 	\arrow["{{\bar{P}(s_{2,1}|s_{1,2})=0}}"{description, pos=0.7}, from=4-4, to=1-7]
% 	\arrow["{{\bar{P}(s_{2,2}|s_{1,2})=1}}"{description}, from=4-4, to=4-7]
% \end{tikzcd}
% $$
We define $P_{u,n}$ as a uniform distribution such that
$$
P_{u,n}(S_n|S_{n-1}) = \frac{1}{2}\text{\quad if $(S_{n},S_{n-1})=(s_{n,m'},s_{n-1,m})$} \text{\, for all $n,m,m'\in[2]$}.
$$
% which can be illustrated as 
% $$
% \begin{tikzcd}
% 	{s_{0,1}} &&& {s_{1,1}} &&& {s_{2,1}} \\
% 	\\
% 	\\
% 	{s_{0,2}} &&& {s_{1,2}} &&& {s_{2,2}}
% 	\arrow["{{P_u(s_{1,1}|s_{0,1})=\frac{1}{2}}}"{description}, from=1-1, to=1-4]
% 	\arrow["{{P_u(s_{1,2}|s_{0,1})=\frac{1}{2}}}"{description, pos=0.7}, from=1-1, to=4-4]
% 	\arrow["{{P_u(s_{2,1}|s_{1,1})=\frac{1}{2}}}"{description}, from=1-4, to=1-7]
% 	\arrow["{{P_u(s_{2,2}|s_{1,1})=\frac{1}{2}}}"{description, pos=0.7}, from=1-4, to=4-7]
% 	\arrow["{{P_u(s_{1,1}|s_{0,2})=\frac{1}{2}}}"{description, pos=0.7}, from=4-1, to=1-4]
% 	\arrow["{{P_u(s_{1,2}|s_{0,2})=\frac{1}{2}}}"{description}, from=4-1, to=4-4]
% 	\arrow["{{P_u(s_{2,1}|s_{1,2})=\frac{1}{2}}}"{description, pos=0.7}, from=4-4, to=1-7]
% 	\arrow["{{P_u(s_{2,2}|s_{1,2})=\frac{1}{2}}}"{description}, from=4-4, to=4-7]
% \end{tikzcd}
% $$
With these assumptions in place, we turn to the questions outlined in Sec.~\ref{sec:contribution}. Regarding  the first question, Theorem~\ref{theorem:toy2_nter} demonstrates that conditions for the pre-trained models to bootstrap the RL-STaR algorithm are that the pre-trained LLM captures certain critical features of the ground-truth transition, thus enabling it to surpass a randomly initialized model. The following theorem subsequently verifies this condition.

\begin{theorem}[\textbf{Sufficient Conditions for Pre-trained Models}]
\label{theorem:toy2_nter}
Given the toy example defined in the previous paragraph, 
let Assumptions~\ref{assump:dataset} and~\ref{assump:exact_fit} hold, so that $\mathcal{D} = \{(s_{0,1}, s_{2,1}), (s_{0,2}, s_{2,2})\}$.
Assume that the pre-trained LLM estimates $P_{0,n}=P_n[\delta_{0,n}]$ (Definition~\ref{def:delta_family}) at every CoT step $n\in[2]$, for some $0 \leq \delta_{0,n} < 1$.
If, at the start of RL-STaR iteration $t$, there exist $0\leq\delta_{t-1,n}<1$ with $P_{t-1,n}=P_n[\delta_{t-1,n}]$ for every step $n\in[2]$, then $P_{t,n}=P_n[\delta_{t,n}]$ for every step $n\in[2]$, where
$$
0\leq \delta_{t,1}=\delta_{t,2}=  \frac{\delta_{t-1,1}+\delta_{t-1,2}}{\delta_{t-1,1}\delta_{t-1,2}+1}<1.
$$ 
Besides, conditions on the values of $\delta_{0,1}$ and $\delta_{0,2}$ can be listed as follows:
\begin{enumerate}[label=(\alph*)]
\item If \(0 < \delta_{0,1}\) and  \(0 < \delta_{0,2} \), then for all \(t \geq 1\),
    $$
    \delta_{t-1,1} < \delta_{t,1}  \quad \text{and} \quad \delta_{t-1,2} < \delta_{t,2}.
    $$
\item  If exactly one of \(\delta_{0,1}\) or \(\delta_{0,2}\) is zero—meaning there exist \(n,n' \in \{1,2\}\) with \(\delta_{0,n} = 0\) but \(\delta_{0,n'} > 0\)—then 
\[
\delta_{1,n} = \delta_{1,n'} = \delta_{0,n'} > 0,
\]
and for all \(t > 1\),
\[
\delta_{t-1,1} < \delta_{t,1} 
\quad\text{and}\quad 
\delta_{t-1,2} < \delta_{t,2}.
\]
\item If both $\delta_{0,1}=0$ and $\delta_{0,2}=0$, then for all $t\geq 1$,
    $$
    \delta_{t,1}=\delta_{t,2}=0.
    $$    
\end{enumerate}
% $$
% \delta_{t-1}<\delta_t=\frac{\delta_{t-1}}{2\left(\frac{1}{2^2}+\delta_{t-1}^2\right)}<\frac{1}{2},
% $$ 
% $$
% \delta_{t-1}<
% \delta_t <\frac{1}{2},\quad\text{and}\quad \delta_t=\frac{\left(\frac{\delta_0^{-1} +2}{\delta_0^{-1} -2}\right)^{2^t} -1}{2\left(1 + \left(\frac{\delta_0^{-1} +2}{\delta_0^{-1} -2}\right)^{2^t}\right)} .
% $$
% Besides, the reward $J(P_t)$ is
% $$
% J\left(P_t\right)=2\left(\frac{1}{2^2}+\delta_t^2\right).
% $$
\end{theorem}
\proof The proof is in~\ref{sec:proof_of_theorem:toy2_nter}.
\qedhere

Building upon the preceding theorem, we can establish the convergence rate of \(\delta_{t,n}\). This result is presented in the following theorem
\begin{theorem}[\textbf{Convergence Speed of $\delta_t$}]\label{theorem:toy2_convergence_speed}
Given the toy example defined in Sec.~\ref{sec:toy_example}, if  $\delta_{t,1}=\delta_{t,2} > 0 $ for all $t\geq 1$, we define $\delta_t = \delta_{t,1}=\delta_{t,2}$. Then,
$\delta_t=\left(\left(\frac{\delta_0^{-1}+1}{\delta_0^{-1}-1}\right)^{2^t} -1\right)\Bigg/\left(\left(\frac{\delta_0^{-1}+1}{\delta_0^{-1}-1}\right)^{2^t}+1\right)
$ .
\end{theorem}
\proof The proof is in~\ref{sec:proof_of_theorem:toy2_convergence_speed}.
\qedhere

We also address the remaining questions about the STaR algorithm for this toy example, as detailed in the following.
\begin{corollary}[\textbf{Policy Improvement in the Toy Example}]\label{theorem:toy2_policy_improvement}
Given the toy example defined in Sec.~\ref{sec:toy_example}, let \(P_t\) be the transition model at iteration \(t\) of RL-STaR training.  
If $\delta_{t,1}=\delta_{t,2} > 0 $ for all $t\geq 1$, we define $\delta_t = \delta_{t,1}=\delta_{t,2}$. Then, 
\[
J(P_t) > J(P_{t-1}).
\]
\end{corollary}
\proof The proof is in~\ref{sec:proof_of_theorem:toy2_policy_improvement}.
\qedhere
\begin{corollary}[\textbf{Convergence to Optimal Policy in the Toy Example}]\label{theorem:toy_optimal_policy} 
Given the toy example defined in Sec.~\ref{sec:toy_example},   
If $\delta_{t,1}=\delta_{t,2} > 0 $ for all $t\geq 1$, we define $\delta_t = \delta_{t,1}=\delta_{t,2}$. 
Define $P^{\star}$ as the optimal estimated transition, which maximizes the reward $J(P^{\star})$. This maximum is achieved when
$$
J(P^{\star}) = \lim_{\delta \rightarrow 1}\left(\frac{1+\delta^2}{2}\right) = 1.
$$
\end{corollary}
\proof The proof is in~\ref{sec:proof_of_theorem:toy_optimal_policy}.
\qedhere
\begin{corollary}[\textbf{Diminishing of Incorrect Reasoning Trajectories}]\label{theorem:toy2_diminish_reasonong_path}
Given the toy example defined in Sec.~\ref{sec:toy_example},   
If $\delta_{t,1}=\delta_{t,2} > 0 $ for all $t\geq 1$, we define $\delta_t = \delta_{t,1}=\delta_{t,2}$. 
Denote $\tau'$ as the incorrect reasoning trajectories included in the dataset $\mathcal{D}_t$. There are two types of $\tau'$ in this toy example:
$$
\tau' = (s_{0,1}, s_{1,2}, s_{2,1}), \quad \text{and}\quad \tau' = (s_{0,2}, s_{1,1}, s_{2,2}). 
$$
When $t$ increases, the probability of $\tau'\in D$ diminishes. Specifically, 
$$
\lim_{t\rightarrow \infty}p(\tau' \in \mathcal{D}_t)=0.
$$
\end{corollary}
\proof The proof is in~\ref{sec:proof_of_theorem:toy2_diminish_reasonong_path}.
\qedhere

We move on to a more general case in the next section.
%as detailed in Sec.~\ref{sec:additional_theorems}.
% \subsection{Additional Theorems and Corollaries}\label{sec:additional_theorems}
% \begin{corollary}[\textbf{Policy Improvement in the Toy Example}]\label{theorem:toy_policy_improvement}
% Given the toy example defined in Sec.~\ref{sec:toy_example}, let $P_t$ represent the estimated transition of the model at the $t$-th iteration of RL-STaR training. We aim to show that the training process improves the reward $J(P_t)$, namely,
% $$
% J(P_t) \geq J(P_{t-1}).
% $$
% \end{corollary}
% \proof The proof can be found in Sec.~\ref{sec:proof_of_theorem:toy_policy_improvement}. 
% \qedhere 
% \begin{corollary}[\textbf{Convergence to Optimal Policy in the Toy Example}]\label{theorem:toy_optimal_policy}
% Define $P^{\star}$ as the optimal estimated transition, which maximizes the reward $J(P^{\star})$. This maximum is achieved when
% $$
% J(P^{\star}) = \sup_{\delta \in \left(0, \frac{1}{2}\right)} 2\left(\frac{1}{2^2} + \delta^2\right) = \lim_{\delta \rightarrow \frac{1}{2}} 2\left(\frac{1}{2^2} + \delta^2\right) = 1.
% $$
% \end{corollary}
% \proof The proof can be found in Sec.~\ref{sec:proof_of_theorem:toy_optimal_policy}\qedhere

% \todo{Diminishing of incorrect reasoning steps}

\subsection{Main Theorem}\label{sec:intro_main_theorem}
After introducing a toy example, we now present our main theorem, which can be applied to an arbitrary number of reasoning steps $N$ and an arbitrary number of states $M$ for each step. In this scenario, $S_0$ is a random variable that represents the initial state, $S_1, \dots, S_{N-1}$ represent the intermediate states of steps $1$ to $N-1$ respectively, and $S_N$ represents the final state. Each step $n \in [N]$ contains $M$ possible states, so $\operatorname{support}(S_n) = \{s_{n,1}, s_{n,2}, \dots, s_{n,M}\}$.
Assuming there are $M$ ground-truth reasoning trajectories, we denote these trajectories as $
\{\tau_{m} | m \in [M]\}$ where each trajectory $\tau_m$ has the form of $(s_{0,m}, s_{1,m}, \dots, s_{N,m})$. The transition function for step $n$, denoted as $\bar{P}_{n}(S_n | S_{n-1})$, is defined as
\begin{equation*}
\bar{P}_n(S_n | S_{n-1}) 
= 
\begin{cases}
1 & \text{if $(S_{n},S_{n-1})=(s_{n,m},s_{n-1,m})$} \text{ for all $m\in[M]$},  \\
0  & \text {if  $(S_n,S_{n-1})=(s_{n,m'\neq m},s_{n-1, m})$ } \text { for all  $ m,m' \in [M]$},
\end{cases}
\end{equation*}
and the uniform transition at step $n$, denoted as $P_{u,n}(S_{n} | S_{n-1})$, is defined as
$$
P_{u,n}(S_{n} | S_{n-1}) = \frac{1}{M} \quad \text{for all } (S_n, S_{n-1}).
$$
We now address the questions listed in Sec.~\ref{sec:contribution}. In response to the first question, Theorem~\ref{theorem:main_theorem} indicates that a condition for the pre-trained models to bootstrap the RL-STaR algorithm is that the pre-trained LLM outperforms a randomly initialized model at every reasoning step. %This condition is subsequently verified by the following theory.%
\begin{theorem}[\textbf{Sufficient Conditions of Pre-trained Models}]
\label{theorem:main_theorem}
Given the scenario defined in the previous paragraph, let Assumptions~\ref{assump:dataset} and~\ref{assump:exact_fit} hold, and assume that the pre-trained LLM estimates $P_{0,n}=P_n[\delta_{0,n}]$ at every CoT step $n\in[N]$, for some $0 \leq \delta_{0,n} < 1$.
If, at the start of RL-STaR iteration $t$, there exist $0\leq \delta_{t-1,n} < 1$ with $P_{t-1,n}=P_n[\delta_{t-1,n}]$ for every CoT step $n\in[N]$, then $P_{t,n}=P_n[\delta_{t,n}]$ for all $n\in[N]$, where
$$
\delta_{t,n}=
\frac{(M-2)\prod_{k=1}^{N} \delta_{t-1,k}+\prod_{k\neq n}\delta_{t-1,k}+\delta_{t-1,n}}{(M-1)\prod_{k=1}^{N} \delta_{t-1,k}+1}.
$$
Besides, based on the values of $\delta_{0,n}$, we have the following cases:
\begin{enumerate}[label=(\alph*)]
\item \label{theorem:item_d1} If \(0 < \delta_{0,n} < 1\) for each \(n \in [N]\), then for all \(t \geq 1\),
    $$
    \delta_{t-1,n} < \delta_{t,n} < 1.
    $$
    \item  \label{theorem:item_d2} If there exists a step $n\in[N]$ satisfying $\delta_{0,n}=0$ and for any other $n'\neq n,~n'\in[N]$ we have $\delta_{0,n'}>0$, then when $t=1$,
    $$
    \delta_{1,n} = \prod_{n'\neq n}\delta_{0,n'} > 0
    \quad\text{and}\quad 
    \delta_{1,n'}=\delta_{0,n'} > 0,
    $$
    and for all $t > 1$,
    $$
    \delta_{t-1,n} < \delta_{t,n} < 1. \quad \text{for all $n\in[N]$}.
    $$
\item  \label{theorem:item_d3} 
If there exist two distinct steps $n,n'\in[N]$ such that $\delta_{0,n}= 0 =\delta_{0,n'}$, then for all $t\geq 1$,
    $$
    \delta_{t,n}=\delta_{t-1,n} \quad \text{for all $n\in[N]$}.
    $$
\end{enumerate}

\end{theorem}
% $$
% \begin{aligned}
% \delta_{t-1}<&\delta_{t} <1-\frac{1}{M}. %\quad 
% % \text{and} \\
% %      &\delta_{t} =\frac{d(\delta_{t-1})}{d(\delta_{t-1})+(M-1)d'(\delta_{t-1})} - \frac{1}{M}
% \end{aligned}
% $$
% where
% \begin{multline*}
% d(\delta_{t-1}) = \sum\limits_{k_1+\dots+k_M = N-1}{{N-1}\choose{k_1, \dots, k_M}}\\ 
% (\delta_{t-1} + \frac{1}{M})^{k_1+1}(\frac{1}{M} - \frac{\delta_{t-1}}{M-1})^{k_2 + \cdots+ k_M}
% \end{multline*}
% and
% \begin{multline*}
% d'(\delta_{t-1})=\sum\limits_{k'_1+\dots+k'_M = N-1}{{N-1}\choose{k'_1, \dots, k'_M}}\\ (\delta_{t-1} + \frac{1}{M})^{k'_1}(\frac{1}{M} - \frac{\delta_{t-1}}{M-1})^{1 +k'_2 + \cdots+ k'_M}
% \end{multline*}
% in which $k_i, k'_i \geq 0\  \forall i$, $\big(k_2 + 2 k_3 + \dots + (M-1)k_M\big) \equiv 0 \pmod{M}$, $\big(k'_2 + 2 k'_3 + \dots + (M-1)k'_M\big) \equiv 1 \pmod{M}$ and ${{N-1}\choose{k_1, \dots, k_M}} = \frac{(N-1)!}{k_1! \dots k_M!}$.
% specifically
% $$
% \delta_{t}=
% \begin{cases}
% \frac{\delta_{t-1}\left(\delta_{t-1} M^2-2 \delta_{t-1} M+2 M-2\right)}{\delta_{t-1}^2 M^2+M-1} & \text{when $N=2$}, \\
% \frac{\delta_{t-1}(M^3\delta_{t-1}^2+M^2\delta_{t-1}-2M-2\delta_{t-1}^2M^2-\delta_{t-1}M+M^2+1)}{\delta_{t-1}^3M^3+M^2-2M+1} & \text{when $N=3$}.
% \end{cases}
% $$ 
\proof The proof is in~\ref{sec:proof_of_theorem:main_theorem}.
\qedhere

The above theorem establishes three key points:
\begin{itemize}
\item[\ref{theorem:item_d1}]  If \(\delta_{0,n} > 0\) for all $n\in[N]$, then \(\delta_{t,n}\) is strictly increasing in \(t\).  
\item[\ref{theorem:item_d2}] There can be at most one step \(n\) with \(\delta_{0,n} = 0\). After the first iteration of RL-STaR, we have \(\delta_{1,n} > 0\) for all \(n \in [N]\), and hence by \ref{theorem:item_d1}, \(\delta_{t,k} > 0\) will be strictly increasing in \(t\) for all \(t > 1\) and \(k \in [N]\).  
\item[\ref{theorem:item_d3}]  Conversely, if more than one step satisfies \(\delta_{0,n} = 0\), then \(\delta_{t,k}\) remains unchanged for all \(t \ge 1\).
\end{itemize}
In conclusion, the conditions of a pre-trained model to improve itself through RL-STaR are to satisfy \ref{theorem:item_d1} or at least \ref{theorem:item_d2}. 
The following theorem establishes the convergence speed of $\delta_{t,n}$ toward $1$.
\begin{theorem}[\textbf{Convergence Speed of $\delta_{t,n}$}]\label{theorem:main_theorem_delta_speed}
Given the scenario defined in Theorem~\ref{theorem:main_theorem}~\ref{theorem:item_d1} or \ref{theorem:item_d2}, 
denote 
$$
\gamma=\begin{cases}    
\frac{1-\prod_{k=1}^{N}\delta_{0,k}}{(M-1)\prod_{k=1}^{N}\delta_{0,k}+1}\quad\text{if~\ref{theorem:item_d1} is satisfied}, \\
\frac{1-\prod_{k=1}^{N}\delta_{1,k}}{(M-1)\prod_{k=1}^{N}\delta_{1,k}+1}\quad\text{if~\ref{theorem:item_d2} is satisfied}. 
\end{cases}
$$
Then for any $\varepsilon \in\left(0, \frac{M}{N(M+1)}\right)$, it holds that
\begin{align*}
\delta_{t,n} > 1-\varepsilon,~&\forall t \geq\left\lceil\frac{\log \frac{M+1}{M \gamma}+\log \left(N-\sum_{k=1}^{N}\delta_{0,k}\right)}{\log (1 / \gamma)}\right\rceil_{+} \\
& \quad+\left\lceil\log _{2}\left(\frac{\log M+\log \frac{1-N\varepsilon}{N\varepsilon}}{\log \frac{(M+1)-M \gamma}{\gamma}}\right)\right\rceil_{+} \\
&\quad+\left\lfloor 1-\prod_{k=1}^{N}\delta_{0,k} \right\rfloor_{+}
\quad\text{for every $n\in[N]$}.
\end{align*}
\end{theorem}
\proof The proof is in~\ref{sec:proof_of_theorem:main_theorem_delta_speed}.
\qedhere

% The previous theorem discuss the scenario when $\delta_{t,n}>0$, the followings we discuss when there is one or more CoT steps $n$ satisfying $\delta_{t,n}=0$. 
% \begin{theorem}[\textbf{Boundary Conditions of Pre-trained Models}]
% \label{theorem:main_theorem_boundary_conditions}
% Given the scenario defined in Theorem~\ref{theorem:main_theorem}, if
% \begin{enumerate}[label=(\alph*)]
%     \item If there exist a step $n\in[N]$ satisfying $\delta_{t-1,n}=0$ and for any other $n'\neq n,~n'\in[N]$ satisfying $\delta_{t-1,n'}>0$, then
%     $$
%     \delta_{t,n} = \prod_{n'\neq n}\delta_{t-1,n'} > 0
%     \quad\text{and}\quad 
%     \delta_{t,n'}=\delta_{t-1,n'} > 0.
%     $$
% \item If there exist two steps $n,n'\in[N]$ satisfying $n'\neq n$ and $\delta_{t-1,n}=\delta_{t-1,n'}=0$, then
%     $$
%     \delta_{t,n}=\delta_{t-1,n} \quad \forall n\in[N].
%     $$
% \end{enumerate}
% \end{theorem}
% \proof The proof can be found in Sec.~\ref{sec:proof_of_theorem:main_theorem_boundary_conditions}.
% \qedhere

The remaining questions in Sec.~\ref{sec:contribution} are answered by the subsequent corollaries.% In the first iteration $j=1$, we will obtain the trajectories $\tau$ with the following probabilities 
% $$
% p(\tau|s_N=s_N^{\star})=
% \begin{cases}
%  \alpha_0^N          & \text{if $\tau \in \mathcal{T}$}, \\
% \frac{(N-1)(2M-2)}{M} \beta_0^2\alpha_0^{N-2} & \text{if $\tau=(s_{0,m},\cdots,s_{N,m})$ but  one $s_{n,m'}$ satisfies $m' \neq m $ and $0<n < N$}, \\
% \cdots 
% \end{cases}
% $$

%\subsubsection{Policy Improvement}
\begin{corollary}[\textbf{Policy Improvement}]
\label{theorem:main_theorem_policy_improvement}
Suppose the assumptions of Theorem~\ref{theorem:main_theorem}~\ref{theorem:item_d1} or \ref{theorem:item_d2} hold. Let $P_{t} = \{P_{t,n}\}_{n=1}^{N}$ represent the estimated transition by the model in the $t$-th iteration of RL-STaR training. Then the training process improves $J(P_{t})$. Specifically,
$
J(P_{t+1}) \geq J(P_{t})
$.
\end{corollary}
\proof The proof is in~\ref{sec:proof_of_theorem:main_theorem_policy_improvement}.
\qedhere

%\subsubsection{Convergence to the Optimal Policy} 
\begin{corollary}[\textbf{Convergence to the Optimal Policy}]
\label{theorem:main_theorem_optimal_policy} Suppose the assumptions of Theorem~\ref{theorem:main_theorem}~\ref{theorem:item_d1} or \ref{theorem:item_d2} hold.
%Let the set of ground-truth reasoning paths $\mathcal{T} = \{\tau_{m} | m \in [M]\}$, where each path $\tau_m$ has the form $(s_{0,m}, s_{1,m}, \dots, s_{N,m})$, be given.
If the optimal transition $P^{\star}$ matches the $M \times M$ identity transition $I_{M}$ in every CoT step, 
when the training iteration $t$ of RL-STaR approaches infinity, $P_t = \{P_{t,n}\}_{n=1}^{N}$ will converge to $P^{\star}$. That is,
$
\lim_{t\rightarrow \infty}\| P_{t,n}-I_{M} \|_{\infty}=0 ~ \text{for all}~  n \in [N]
$.
\end{corollary}
\proof The proof is in~\ref{sec:proof_of_theorem:main_theorem_optimal_policy}.
\qedhere

%\paragraph{Properties of pre-training models for STaR:}

\begin{corollary}[\textbf{Diminishing of Incorrect Reasoning Trajectories}]
\label{theorem:main_theorem_incorrect_reasoning_step}
Suppose the assumptions of Theorem~\ref{theorem:main_theorem}~\ref{theorem:item_d1} or \ref{theorem:item_d2} hold. In iterations $t$ of RL-STaR, denote $\tau_{t,k}$ to be a trajectory containing $k$ incorrect reasoning steps in $\mathcal{D}_{t}$. Then, the probability that RL-CoT generates trajectories containing incorrect reasoning steps will diminish as $t$ increases. Specifically,
$\lim_{t\rightarrow\infty}p\left(\bigcup_{k}\tau_{t,k}\right)=0$.
\end{corollary}
\proof The proof is in~\ref{sec:proof_of_theorem:main_theorem_incorrect_reasoning_step}.
\qedhere

\subsection{Comparison with the Convergence Analysis of Policy-Gradient Methods}
\label{sec:comparison_pg}
At first glance, STaR~\citep{zelikman2022star} might appear to be an offline variant of policy-gradient methods in reinforcement learning, suggesting that classical convergence guarantees~\citep{6796286,bhandari2021linear,bhandari2024global} could be directly applied, rendering STaR’s convergence analysis trivial.
However, 
% the original STaR paper~\citep{zelikman2022star} explicitly notes:\begin{quote}
% \textit{STaR can be seen as an approximation to an RL-style
% policy gradient objective... future work should more closely
% investigate the link between STaR and the RL objective above.}
% \end{quote}
% This statement leaves the precise relationship—including both similarities and differences—open for further investigation. 
to provide a clear comparison between STaR and RL-style policy-gradient methods, we first revisit the 
%original claim of STaR:
% \begin{quote}
% \textit{by leveraging the LLM’s pre-existing reasoning ability, we iteratively bootstrap the ability to generate high-quality rationales ... refine the model’s ability further by fine-tuning on those rationales that lead to correct answers.} 
% \end{quote}
%From this description, several distinguishing 
distinguishing characteristics of STaR's convergence:
\begin{enumerate}
\item STaR exploits the inductive biases of a pretrained LLM to bootstrap improved reasoning traces.
\item The reward signal is sparse, depending solely on the final correctness of the answer.
\item Iterative self-training enables progressive refinement of intermediate reasoning steps.
\end{enumerate}
Although both STaR and policy-gradient methods can operate under similar settings—such as receiving a reward only at the final step, or being initialized from a pre-trained model—their training dynamics differ fundamentally.  Besides, from a theoretical perspective, the convergence guarantees of these policy-gradient algorithms focus on the maximization of the expected return, $\mathbb{E}_{\pi_{\theta}}[v_t]$. These guarantees ensure that the parameter $\theta$ converges to a local (or global) optimum of the expected return, but do not provide assurances about the improvement of intermediate reasoning steps.
This distinction creates different learning dynamics---STaR may gradually suppress incorrect intermediate reasoning steps if the base model is biased toward correct ones, even without stepwise rewards. Our analysis formalizes this, showing how, under specific conditions, STaR can refine reasoning paths through this iterative process.

\section{Experiments}
\label{sec:experiment}
We conduct experiments to illustrate our theoretical analysis in Sec.~\ref{sec:intro_main_theorem}.  For the language model, we choose GPT-2 \citep{radford2019language}, Qwen2.5-0.5B~\citep{qwen2025qwen25technicalreport} and Phi-3-mini~\citep{abdin2024phi3technicalreporthighly}. Besides, to satisfy the assumption of Markov property and constant state space $M$, we conduct the following modifications on LLMs: For the Markov property, we restrict the length of the context window of LLMs so that they can only take $s_{n-1}$ as input when the output is $s_n$.  For the uniformity of $\delta_{t,n}$, we reinitialize the weights of LLMs and pre-train them on a dataset comprising noisy trajectories whose transitions between \(s_n\) and \(s_{n-1}\) in this dataset match the transition probability \(P_0(S_n=s_n|S_{n-1}=s_{n-1})\) referenced in Theorem~\ref{theorem:main_theorem}. For the constant $M$, we sample the output of LLMs within $M$ possible valid states for each CoT step.

Sec.~\ref{sec:experiment_A} to Sec.~\ref{sec:experiment_C3} cover the three cases \ref{theorem:item_d1}--\ref{theorem:item_d3} of Theorem~\ref{theorem:main_theorem}; Sec.~\ref{sec:experiment_non_uniform_delta} and Sec.~\ref{sec:experiment_non_constant_M} then drop two of the assumptions discussed in Sec.~\ref{sec:more_about_limitation}. 
%Besides,  we prodives the experiment when Markov assumption is not satisfied in Sec~\ref{sec:exp_non_markov}. 
% Our experiments are conducted on a machine with RTX 4090 GPU card and A100 GPU card, which takes about 50 GPU-hours for 10 iterations of RL-STaR. 
For the scenario where the constant state space $M$ is satisfied, we choose two reasoning tasks as shown below:
\paragraph{Task `zip'}
This task is known as the zip operator
\footnote{\url{https://www.w3schools.com/python/ref_func_zip.asp}}. 
We consider the scenario with binary strings of length three. An example of this operation is as follows:
$$
\begin{aligned}
& s_0 = \texttt{x:101,110}  \\
&\Rightarrow  s_1 = \texttt{x:10,11},\texttt{y:10}   \\
&\Rightarrow  s_2 = \texttt{x:11},\texttt{y:01,10} \\
&\Rightarrow  s_3 = \texttt{y:11,01,10}.
\end{aligned}
$$
Here, the symbols $\texttt{x}$ and $\texttt{y}$ represent the input and output at each step, respectively. At each step, a single token from $\texttt{x}$ is paired with the output $\texttt{y}$. This example has total number of reasoning steps $N=3$, and each step has the equal number of states $M=64$.
%In this task, we restrict the output of LLMs within $M=64$ valid states within each CoT step.
\paragraph{Task `dec2bin'} This is the task of converting a decimal number (that fits within three bits) into its binary representation via long division. For example,
$$
\begin{aligned}
& s_0=\texttt{x:7} \\
&\Rightarrow s_1=\texttt{x:3,y:1} \\
& \Rightarrow s_2=\texttt{x:1,y:11} \\
& \Rightarrow  s_3=\texttt{y:111}.
\end{aligned}
$$
The process consists of three steps $N=3$, and each step includes eight possible states $M=8$.

\begin{figure*}[t!]
    \centering
        \centering
        \begin{tabular}{c c c}
        \resizebox{.31\textwidth}{!}{\begin{tikzpicture}
\begin{axis}[
    width=6cm,
    height=3.5cm,
    legend pos=south east,
    grid=major,
    grid style={dashed,gray!30},
    xmin=0, xmax=9,
    ymin=-0.1, ymax=1.1,
    font=\scriptsize,
    xlabel={$t$},
    ylabel={$J(P_t)$},
    legend cell align={left},
    title style={font=\scriptsize},
    label style={font=\scriptsize},
    tick label style={font=\scriptsize},
    legend style={font=\tiny},
    xlabel style={
        at={(current axis.south east)}, % Relative positioning
        anchor=north east,              % Anchoring at a specific point
        yshift=0pt,                   % Shifting downward
        xshift=10pt                      % Shifting rightward (if needed)
    },
    ylabel style={
        at={(current axis.north west)}, % Relative positioning
        anchor=north east,              % Anchoring at a specific point
        yshift=20pt,                   % Shifting downward
        xshift=0pt                      % Shifting rightward (if needed)
    },
   title={\texttt{zip}, GPT-2},
]
\addplot[c1, solid, thick] table[row sep=\\]{
        x y \\
0    0.0238809838750315 \\
1   0.1387970147215855 \\
2    0.8805859938624284 \\
3    0.9998173897891941 \\
4    0.99999999964703 \\
5    1.0 \\
6    1.0 \\
7    1.0 \\
8    1.0 \\
9    1.0 \\
};
\addlegendentry{$\delta_0=0.2$, Theory}
\addplot[c1, dashed, thick] table[row sep=\\]{
        x y \\
0 0.028106689453125 \\
1 0.4643058776855469 \\
2 0.9990997314453125 \\
3 1.0 \\
4 1.0 \\
5 1.0 \\
6    1.0 \\
7    1.0 \\
8    1.0 \\
9    1.0 \\
};
\addlegendentry{$\delta_0=0.2$, Exp.}
 \addplot[c2, solid, very thick] table[row sep=\\]{
        x y \\
0    0.01665699798437894 \\
1    0.020123459753581558 \\
2    0.06557845281185505 \\
3    0.6649908499462328 \\
4    0.9978254678481339 \\
5    0.9999999497974984 \\
6    1.0 \\
7    1.0 \\
8    1.0 \\
9    1.0 \\
% 10    1.0 \\
};
\addlegendentry{$\delta_0=0.1$, Theory}
 \addplot[c2, dashed, thick] table[row sep=\\]{
        x y \\
% 0 0.016307830810546875 \\
0 0.016307830810546875 \\
1 0.03815460205078125 \\
2 0.4627265930175781 \\
3 0.9492 \\
4 0.9957 \\
5 1.0 \\
6    1.0 \\
7    1.0 \\
8    1.0 \\
9    1.0 \\
% 5 1.0 \\
% 6 1.0 \\
};
\addlegendentry{$\delta_0=0.1$, Exp.}
 \end{axis}
    \end{tikzpicture}
} & 
            \resizebox{.31\textwidth}{!}{\begin{tikzpicture}
\begin{axis}[
    width=6cm,
    height=3.5cm,
    legend pos=south east,
    grid=major,
    grid style={dashed,gray!30},
    xmin=0, xmax=9,
    ymin=-0.1, ymax=1.1,
    font=\scriptsize,
    xlabel={$t$},
    ylabel={$J(P_t)$},
    legend cell align={left},
    title style={font=\scriptsize},
    label style={font=\scriptsize},
    tick label style={font=\scriptsize},
    legend style={font=\tiny},
    xlabel style={
        at={(current axis.south east)}, % Relative positioning
        anchor=north east,              % Anchoring at a specific point
        yshift=0pt,                   % Shifting downward
        xshift=10pt                      % Shifting rightward (if needed)
    },
    ylabel style={
        at={(current axis.north west)}, % Relative positioning
        anchor=north east,              % Anchoring at a specific point
        yshift=20pt,                   % Shifting downward
        xshift=0pt                      % Shifting rightward (if needed)
    },
   title={\texttt{dec2bin}, GPT-2},
]
%  \addplot[c1, solid, thick] table[row sep=\\]{
%         x y \\
% 0    0.0238809838750315 \\
% 1   0.1387970147215855 \\
% 2    0.8805859938624284 \\
% 3    0.9998173897891941 \\
% 4    0.99999999964703 \\
% 5    1.0 \\
% 6    1.0 \\
% 7    1.0 \\
% 8    1.0 \\
% 9    1.0 \\
% };
% \addlegendentry{$\delta_0=0.2$, Theory}
% \addplot[c1, dashed, thick] table[row sep=\\]{
%         x y \\
% 0 0.028106689453125 \\
% 1 0.4643058776855469 \\
% 2 0.9990997314453125 \\
% 3 1.0 \\
% 4 1.0 \\
% 5 1.0 \\
% 6    1.0 \\
% 7    1.0 \\
% 8    1.0 \\
% 9    1.0 \\
% };
% \addlegendentry{$\delta_0=0.2$, Exp.}
%  \addplot[c2, solid, very thick] table[row sep=\\]{
%         x y \\
% 0    0.01665699798437894 \\
% 1    0.020123459753581558 \\
% 2    0.06557845281185505 \\
% 3    0.6649908499462328 \\
% 4    0.9978254678481339 \\
% 5    0.9999999497974984 \\
% 6    1.0 \\
% 7    1.0 \\
% 8    1.0 \\
%  9    1.0 \\
% % 10    1.0 \\
% };
% \addlegendentry{$\delta_0=0.1$, Theory}
%  \addplot[c2, dashed, thick] table[row sep=\\]{
%         x y \\
% % 0 0.016307830810546875 \\
% 0 0.016307830810546875 \\
% 1 0.03815460205078125 \\
% 2 0.4627265930175781 \\
% 3 0.9492 \\
% 4 0.9957 \\
% 5 1.0 \\
% 6    1.0 \\
% 7    1.0 \\
% 8    1.0 \\
% 9    1.0 \\
% % 5 1.0 \\
% % 6 1.0 \\
% };
% \addlegendentry{$\delta_0=0.1$, Exp.}
\addplot[c1, solid, thick] table[row sep=\\]{
x y \\ 
 0  0.18100000000000005 \\ 
 1  0.36744785733474694 \\ 
 2  0.8331764349350943 \\ 
 3  0.9964781633461554 \\ 
 4  0.9999988121640007 \\ 
 5  0.9999999999998654 \\ 
 6  0.9999999999999997 \\ 
 7  1.0000000000000007 \\ 
 8  1.0 \\ 
 9  1.0 \\ 
};
\addlegendentry{$\delta_0=0.4$, Theory}

    \addplot[c1, dashed, thick] table[row sep=\\]{
x y \\ 
 0  0.391632080078125 \\ 
 1  0.996429443359375 \\ 
 2  1.0 \\ 
 3  1.0 \\ 
 4  1.0 \\ 
 5  1.0 \\ 
 6  1.0 \\ 
 7  1.0 \\ 
 8  1.0 \\ 
 9  1.0 \\ 
};
\addlegendentry{$\delta_0=0.4$, Exp.}
\addplot[c2, solid, thick] table[row sep=\\]{
x y \\ 
 0  0.13200000000000003 \\ 
 1  0.14274981217129978 \\ 
 2  0.18553841667881618 \\ 
 3  0.387923843958137 \\ 
 4  0.8548557319137602 \\ 
 5  0.9974377636730782 \\ 
 6  0.9999993722311926 \\ 
 7  0.9999999999999624 \\ 
 8  1.0000000000000007 \\ 
 9  1.0 \\ 
};
\addlegendentry{$\delta_0=0.2$, Theory}

\addplot[c2, dashed, thick] table[row sep=\\]{
x y \\ 
 0  0.16339111328125 \\ 
 1  0.70050048828125 \\ 
 2  0.999908447265625 \\ 
 3  1.0 \\ 
 4  1.0 \\ 
 5  1.0 \\ 
 6  1.0 \\ 
 7  1.0 \\ 
 8  1.0 \\ 
 9  1.0 \\ 
};
\addlegendentry{$\delta_0=0.2$, Exp.}
 \end{axis}
    \end{tikzpicture}
} & 
        \resizebox{.31\textwidth}{!}{\begin{tikzpicture}
\begin{axis}[
    width=6cm,
    height=3.5cm,
    legend pos=south east,
    grid=major,
    grid style={dashed,gray!30},
    xmin=0, xmax=5,
    ymin=-0.1, ymax=1.1,
    font=\scriptsize,
    xlabel={$t$},
    ylabel={$J(P_t)$},
    legend cell align={left},
    title style={font=\scriptsize},
    label style={font=\scriptsize},
    tick label style={font=\scriptsize},
    legend style={font=\tiny},
    xlabel={$t$},    
    xlabel style={
        at={(current axis.south east)}, % Relative positioning
        anchor=north east,              % Anchoring at a specific point
        yshift=0pt,                   % Shifting downward
        xshift=10pt                      % Shifting rightward (if needed)
    },
    ylabel style={
        at={(current axis.north west)}, % Relative positioning
        anchor=north east,              % Anchoring at a specific point
        yshift=20pt,                   % Shifting downward
        xshift=0pt                      % Shifting rightward (if needed)
    },
   title={\texttt{zip2}, $\delta_0=0.2$},
]
 \addplot[c11, solid, thick] table[row sep=\\]{
        x y \\
0    0.0238809838750315 \\
1   0.1387970147215855 \\
2    0.8805859938624284 \\
3    0.9998173897891941 \\
4    0.99999999964703 \\
5    1.0 \\
6    1.0 \\
7    1.0 \\
8    1.0 \\
9    1.0 \\
};
\addlegendentry{Theory}
\addplot[c13, dashed, thick] table[row sep=\\]{
        x y \\
0 0.028106689453125 \\
1 0.4643058776855469 \\
2 0.9990997314453125 \\
3 1.0 \\
4 1.0 \\
5 1.0 \\
6    1.0 \\
7    1.0 \\
8    1.0 \\
9    1.0 \\
};
\addlegendentry{GPT-2, Exp.}
\addplot[c14, dotted, very thick] table[row sep=\\]{
        x y \\
0 0.0272369384765625 \\
1 0.3861732482910156 \\
2 0.9961929321289062 \\
3 1.0 \\
4 1.0 \\
5 1.0 \\
6    1.0 \\
7    1.0 \\
8    1.0 \\
9    1.0 \\
};
\addlegendentry{QWen2.5-0.5B, Exp.}
 \addplot[c15, loosely dotted, very thick] table[row sep=\\]{
        x y \\
0    0.026519775390625 \\
1    0.3972015380859375 \\
2    0.9970169067382812 \\
3    1.0 \\
4    1.0 \\
5    1.0 \\
6    1.0 \\
7    1.0 \\
8    1.0 \\
9    1.0 \\
% 10    1.0 \\
};
\addlegendentry{Phi3-mini, Exp.}
 \end{axis}
    \end{tikzpicture}
} 
        \end{tabular}
        \vspace{3mm}
        \begin{tabular}{c c}
        \resizebox{.42\textwidth}{!}{\begin{tikzpicture}
\begin{axis}[
    width=4cm,
    height=3cm,
    legend pos=outer north east,
    grid=major,
    grid style={dashed,gray!30},
    xmin=1, xmax=4,
    ymin=-0.1, ymax=1.1,
    font=\footnotesize,
    xlabel={$t$},
    ylabel={$100\%$},
    legend cell align={left},
    title style={font=\scriptsize},
    label style={font=\scriptsize},
    tick label style={font=\tiny},
    legend style={font=\tiny},
    xlabel={$t$},    
    xlabel style={
        at={(current axis.south east)}, % Relative positioning
        anchor=north east,              % Anchoring at a specific point
        yshift=0pt,                   % Shifting downward
        xshift=10pt                      % Shifting rightward (if needed)
    },
    ylabel style={
        at={(current axis.north west)}, % Relative positioning
        anchor=north east,              % Anchoring at a specific point
        yshift=20pt,                   % Shifting downward
        xshift=10pt                      % Shifting rightward (if needed)
    },
   title={\texttt{zip}, GPT-2},
]
\addplot[c1, thick, dashed] table[row sep=\\]{
x y \\ 
 1  0.18186753528773072 \\ 
 2  0.955313642525572 \\ 
 3  1.0 \\ 
 4  1.0 \\ 
 5  1.0 \\ 
 6  1.0 \\ 
};
\addlegendentry{$\delta_0=0.2$, Correct}
\addplot[c1, very thick, loosely dotted] table[row sep=\\]{
x y \\ 
 1  0.8181324647122693 \\ 
 2  0.044686357474427965 \\ 
 3  0.0 \\ 
 4  0.0 \\ 
 5  0.0 \\ 
 6  0.0 \\ 
};
\addlegendentry{$\delta_0=0.2$, Incorrect}
    \addplot[c2, thick, dashed] table[row sep=\\]{
x y \\ 
 1  0.013342535081665517 \\ 
 2  0.26304739052189563 \\ 
 3  0.9266782631635353 \\ 
 4  0.9946707178856495 \\ 
 5  0.9999731828508162 \\ 
 6  1.0 \\ 
};
\addlegendentry{$\delta_0=0.1$, Correct}
\addplot[c2, very thick, loosely dotted] table[row sep=\\]{
x y \\ 
 1  0.9866574649183345 \\ 
 2  0.7369526094781044 \\ 
 3  0.07332173683646466 \\ 
 4  0.005329282114350479 \\ 
 5  2.681714918380091e-05 \\ 
 6  0.0 \\ 
};
\addlegendentry{$\delta_0=0.1$, Incorrect}
 \end{axis}
    \end{tikzpicture}
} & 
            \resizebox{.42\textwidth}{!}{\begin{tikzpicture}
\begin{axis}[
    width=4cm,
    height=3cm,
    legend pos=outer north east,
    grid=major,
    grid style={dashed,gray!30},
    xmin=1, xmax=4,
    ymin=-0.1, ymax=1.1,
    font=\footnotesize,
    xlabel={$t$},
    ylabel={$100\%$},
    legend cell align={left},
    title style={font=\scriptsize},
    label style={font=\scriptsize},
    tick label style={font=\tiny},
    legend style={font=\tiny},
    xlabel={$t$},    
    xlabel style={
        at={(current axis.south east)}, % Relative positioning
        anchor=north east,              % Anchoring at a specific point
        yshift=0pt,                   % Shifting downward
        xshift=10pt                      % Shifting rightward (if needed)
    },
    ylabel style={
        at={(current axis.north west)}, % Relative positioning
        anchor=north east,              % Anchoring at a specific point
        yshift=20pt,                   % Shifting downward
        xshift=10pt                      % Shifting rightward (if needed)
    },
   title={\texttt{dec2bin}, GPT-2},
]
\addplot[c1, thick, dashed] table[row sep=\\]{
x y \\ 
 1  0.9146731083924258 \\ 
 2  1.0 \\ 
 3  1.0 \\ 
 4  1.0 \\ 
 5  1.0 \\ 
 6  1.0 \\ 
};
\addlegendentry{$\delta_0=0.4$, Correct}
\addplot[c1, very thick, loosely dotted] table[row sep=\\]{
x y \\ 
 1  0.08532689160757423 \\ 
 2  0.0 \\ 
 3  0.0 \\ 
 4  0.0 \\ 
 5  0.0 \\ 
 6  0.0 \\ 
};
\addlegendentry{$\delta_0=0.4$, Incorrect}
    \addplot[c2, thick, dashed] table[row sep=\\]{
x y \\ 
 1  0.45853567426223385 \\ 
 2  0.9870610786790973 \\ 
 3  1.0 \\ 
 4  1.0 \\ 
 5  1.0 \\ 
 6  1.0 \\ 
};
\addlegendentry{$\delta_0=0.2$, Correct}
\addplot[c2, very thick, loosely dotted] table[row sep=\\]{
x y \\ 
 1  0.5414643257377662 \\ 
 2  0.012938921320902675 \\ 
 3  0.0 \\ 
 4  0.0 \\ 
 5  0.0 \\ 
 6  0.0 \\ 
};
\addlegendentry{$\delta_0=0.2$, Incorrect}
 \end{axis}
    \end{tikzpicture}
} 
        \end{tabular}
        \caption{The first row shows the comparison of theoretical values (solid lines) and empirical values (dashed and dotted lines) of $J(P_t)$. The first two plots illustrate the effects of varying $\delta_0$ and of different task settings, respectively, while the third plot compares results across various LLMs. The second row shows the ratio of correct trajectories and incorrect trajectories in training dataset $\mathcal{D}_t$.
        }
        \label{subfig:main_figure_exp_binary_reward}
    % \caption{Combined visualization of theoretical and experimental comparisons (top), and transitions derived from datasets and learned by LLMs (bottom).}
    % \label{fig:combined_figures}
\end{figure*}

\subsection{Experiment of Theoretical and Empirical Values of $J(P_t)$}\label{sec:experiment_A}
In this experiment, we focus on evaluating \(J(P_t)\) to compare its theoretically predicted values with those observed in practice. For the value of $\delta_{t,n}$, we set \(\delta_{t,3} = \delta_{t,2} = \delta_{t,1} = \delta_t.\) where $\delta_{t}\in\{0.1,0.2\}$ for \texttt{zip} and $\delta_{t}\in\{0.2,0.4\}$ for \texttt{dec2bin}. 
To derive the empirical values, we train LLMs using the training dataset \( \mathcal{D} \) by the procedures described in Algorithm~\ref{algo:rl_star}. To derive the theoretical values, we start from Theorem~\ref{theorem:main_theorem}. For the special case \(N=3\) and \(\delta_{t,n} = \delta_t\) for all \(n\), the value of \(\delta_t\) satisfies:
$$
\delta_t 
= \frac{(M-2) \, \delta_{t-1}^3 + \delta_{t-1}^2 + \delta_{t-1}}
       {(M-1) \, \delta_{t-1}^3 + 1}.
$$
To estimate \(J(P_t)\), it can be shown that when \(N=3\),  
a direct calculation yields:
$$
\begin{aligned}
J(P_t) & = e_1^T P_{t,3} P_{t,2}  P_{t,1} e_1 \\
& = \alpha_t^3 + 3(M-1)\alpha_t\beta_t^2 + (M-1)(M-2)\beta_t^3
\end{aligned}
$$
where $\alpha_t=\frac{1+(M - 1)\delta_t}{M}$ and $\beta_t=\frac{1-\delta_t}{M}$.
\paragraph{Results} 
The results are presented in Fig.~\ref{subfig:main_figure_exp_binary_reward}. In the first row, the three figures illustrate that our theoretical model approximately matches the empirical values obtained from training LLMs across different task settings, $\delta_0$ variations, and model architectures.  While the gap between theory and experiment differs substantially across tasks, different LLM architectures achieve similar empirical values.
Additionally, LLMs exhibit faster convergence compared to the theoretical predictions. This discrepancy arises because Assumption~\ref{assump:exact_fit} requires the transitions learned by LLMs, denoted as $ P_t $, to match the transitions in the dataset $ \mathcal{D}_{t}  $ exactly. In practice, however, this assumption is not always valid. 
The second-row plots reveal a decreasing proportion of incorrect reasoning trajectories in $\mathcal{D}_t$ as $t$ grows, confirming Corollary~\ref{theorem:main_theorem_incorrect_reasoning_step} that erroneous steps’ inclusion probability falls with each STaR iteration.
In the following paragraph, we discuss the cause of the discrepancy between theoretical and empirical values.

% \subsection{Additional Experiments}\label{sec:additional_experiments}
\begin{figure}[t!]
    \centering
        \centering
        \begin{tabular}{c c c}
            \resizebox{0.3\textwidth}{!}{\includegraphics[width=\textwidth]{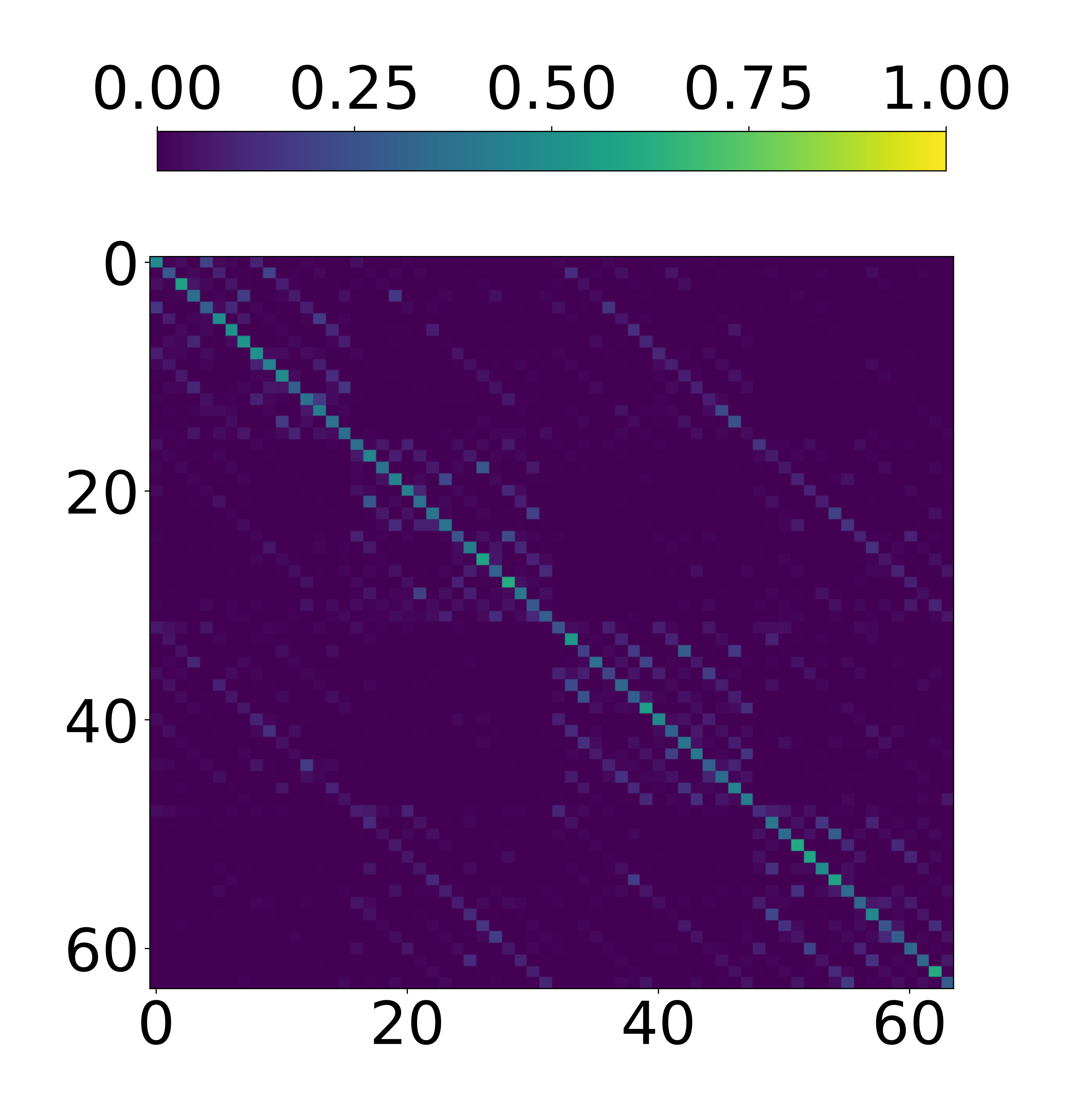}} &
            \resizebox{0.3\textwidth}{!}{\includegraphics[width=\textwidth]{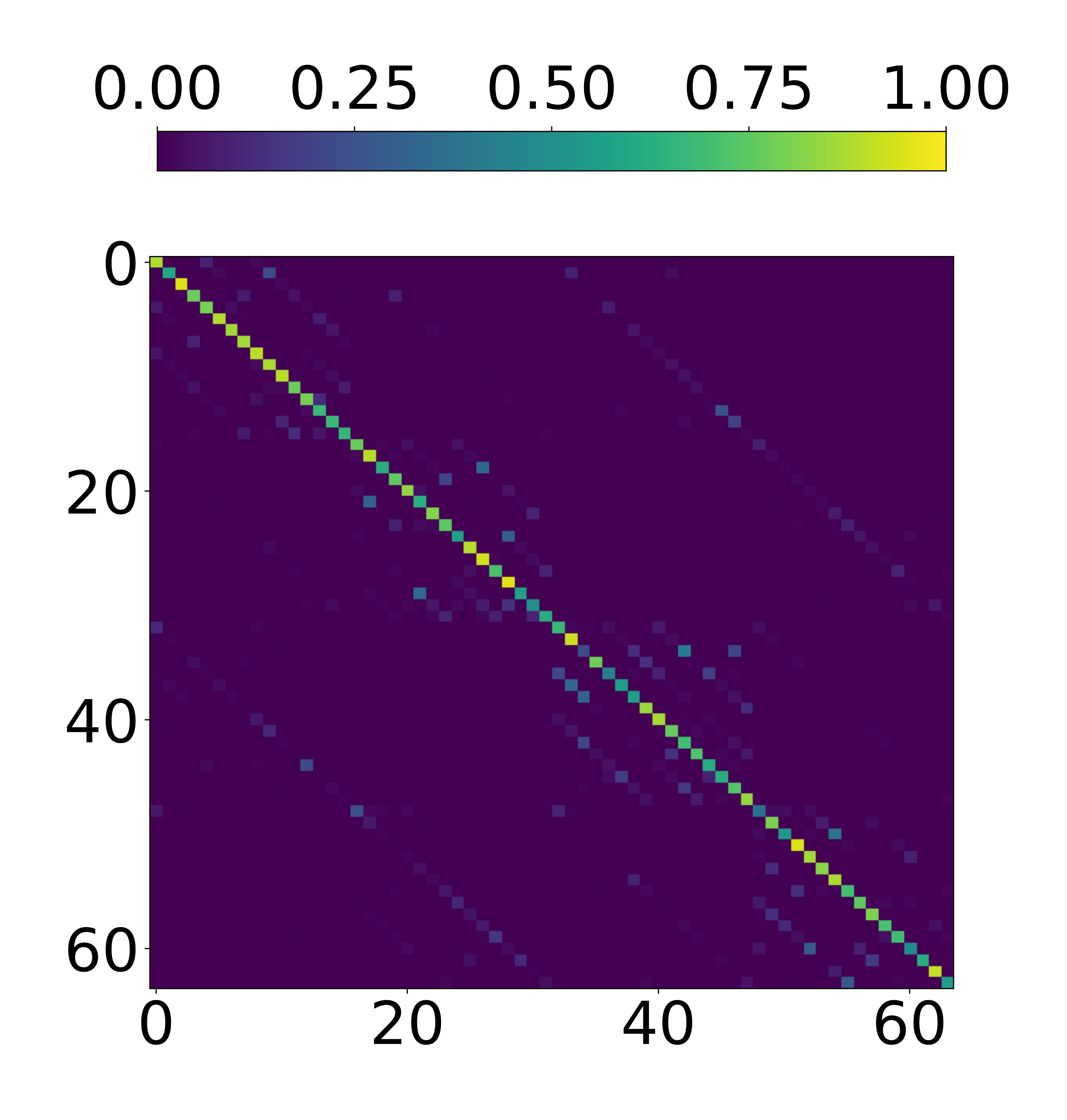}} &
            \resizebox{0.3\textwidth}{!}{\begin{tikzpicture}
\begin{axis}[
    width=5cm,
    height=4cm,
    legend pos=south east,
    grid=major,
    grid style={dashed,gray!30},
    xmin=0, xmax=9,
    ymin=-0.6, ymax=1.1,
    font=\small,
    xlabel={$t$},
    ylabel={$J(P_t)$},
    legend cell align={left},
    title style={font=\small},
    label style={font=\small},
    tick label style={font=\small},
    legend style={font=\tiny},
    xlabel style={
        at={(current axis.south east)}, % Relative positioning
        anchor=north east,              % Anchoring at a specific point
        yshift=0pt,                   % Shifting downward
        xshift=10pt                      % Shifting rightward (if needed)
    },
    ylabel style={
        at={(current axis.north west)}, % Relative positioning
        anchor=north east,              % Anchoring at a specific point
        yshift=20pt,                   % Shifting downward
        xshift=20pt                      % Shifting rightward (if needed)
    },
   title={\texttt{zip}, GPT-2, $\delta_0=0.2$},
]
 \addplot[c1, solid, thick] table[row sep=\\]{
        x y \\
0    0.0238809838750315 \\
1   0.1387970147215855 \\
2    0.8805859938624284 \\
3    0.9998173897891941 \\
4    0.99999999964703 \\
5    1.0 \\
6    1.0 \\
7    1.0 \\
8    1.0 \\
9    1.0 \\
};
\addlegendentry{Theory}
\addplot[c2, dashed, thick] table[row sep=\\]{
        x y \\
 0  0.01685333251953125 \\ 
 1  0.019710540771484375 \\ 
 2  0.02744293212890625 \\ 
 3  0.05300140380859375 \\ 
 4  0.13409805297851562 \\ 
 5  0.2924461364746094 \\ 
 6  0.48504638671875 \\ 
 7  0.6735191345214844 \\ 
 8  0.7896652221679688 \\ 
 9  0.8543815612792969 \\ 
};
\addlegendentry{$\mathcal{T}=1$, Exp.}
\addplot[c3, dashed, thick] table[row sep=\\]{
        x y \\
0 0.028106689453125 \\
1 0.4643058776855469 \\
2 0.9990997314453125 \\
3 1.0 \\
4 1.0 \\
5 1.0 \\
6    1.0 \\
7    1.0 \\
8    1.0 \\
9    1.0 \\
};
\addlegendentry{$\mathcal{T}=0.5$, Exp.}
\addplot[c4, dashed, thick] table[row sep=\\]{
        x y \\
 0  0.13710784912109375 \\ 
 1  0.9998817443847656 \\ 
 2  1.0 \\ 
 3  1.0 \\ 
 4  1.0 \\ 
 5  1.0 \\ 
 6  1.0 \\ 
 7  1.0 \\ 
 8  1.0 \\ 
 9  1.0 \\ 
};
\addlegendentry{$\mathcal{T}=0.25$, Exp.}
 \end{axis}
    \end{tikzpicture}
}
        \end{tabular}
        \caption{The comparison of transitions \( P(S_1|S_0) \), directly extracted from dataset \( \mathcal{D}_1 \) (left), and the transitions \( P_{1,1}(S_1|S_0) \) learned by LLMs during the RL-STaR algorithm (middle). The right plot shows $J(P_t)$ under different sampling temperatures $\mathcal{T}$.}
        \label{subfig:figure_gap_theory}
    % \caption{Combined visualization of theoretical and experimental comparisons (top), and transitions derived from datasets and learned by LLMs (bottom).}
    % \label{fig:combined_figures}
\end{figure}

% \subsubsection{Gap between Theoretical Values and Empirical Values Matrix}\label{sec:vis_transition_matrix}
\paragraph{Visualization of Transition Matrix}
This gap arises when LLMs put excessive probability mass on the instances that appear more frequently in the training dataset. An example is provided in the first two figures in Fig.~\ref{subfig:figure_gap_theory}, which illustrates the difference between the transitions from the dataset and learned by GPT-2 in the task of \texttt{zip} when $\delta_{0,1}=\delta_{0,2}=\delta_{0,3}=0.22$. The left-hand heatmap represents the transitions directly derived from the dataset $ \mathcal{D}_1 $, while the right-hand heatmap depicts the transitions $P_1$ learned by the LLMs. Notably, the LLM-learned transitions' diagonal elements exhibit higher probabilities than those in the original dataset.
\paragraph{Varying the Sampling Temperature} 
We also discover that this gap depends on the sampling temperature. 
We varied the sampling temperature $\mathcal{T}$ around the default value of $0.5$ to assess its influence. The last figure in Fig.~\ref{subfig:figure_gap_theory} demonstrates that higher $\mathcal{T}$ flattens the output distribution—slowing convergence—while lower $\mathcal{T}$ concentrates probability mass and speeds it up. This highlights the importance of tuning $\mathcal{T}$ to strike a balance between convergence rate and output diversity.

% \subsubsection{Experiment of Non-Markov Transitions}\label{sec:exp_non_markov}
% \todo{}

\subsection{Experiment of Convergence of $J(P_t)$ with One $\delta_{0,n}=0$}\label{sec:experiment_C2}
In this experiment, we examine the convergence behavior of $J(P_t)$ when one $\delta_{0,n}=0$. We apply the setting of $\delta_{0,1}=0$ but $\delta_{0,n'}=0.2$ for $n'>1$. 
We use GPT-2 to perform both the \texttt{zip} and \texttt{dec2bin} tasks. However, in the \texttt{dec2bin} experiment, inputs are limited to decimal numbers representable in 5 bits (i.e. $N=5$). All other configurations remain as detailed in Sec.~\ref{sec:experiment_A}.

\begin{figure}[hbt!]
    \centering
    \begin{tabular}{c c}
    \resizebox{0.22\textwidth}{!}{
    \begin{tikzpicture}
\begin{axis}[
width=6cm,
    height=5cm,
legend pos=south east,
    grid=major,
    grid style={dashed,gray!30},
    xmin=0, xmax=8,
    ymin=-0.1, ymax=1.1,
    font=\small,
    legend cell align={left},
    title style={font=\small},
    label style={font=\small},
    tick label style={font=\small},
    legend style={font=\scriptsize},
    xlabel={$t$},    
    xlabel style={
        at={(current axis.south east)}, % Relative positioning
        anchor=north east,              % Anchoring at a specific point
        yshift=0pt,                   % Shifting downward
        xshift=10pt                      % Shifting rightward (if needed)
    },
    ylabel={$J(P_t)$},
    ylabel style={
        at={(current axis.north west)}, % Relative positioning
        anchor=north east,              % Anchoring at a specific point
        yshift=15pt,                   % Shifting downward
        xshift=20pt                      % Shifting rightward (if needed)
    },
   title={\texttt{zip}, GPT-2, $N=3$},
]
\addplot[c1, dashed, thick] table[row sep=\\]{
        x y \\
0 0.01561737060546875 \\
1 0.04375457763671875 \\
2 0.60333251953125 \\
3 0.975555419921875 \\
4 0.984466552734375 \\
5 0.984466552734375 \\
6 0.984466552734375 \\
7 0.984466552734375 \\
8 0.984466552734375 \\
};
\addlegendentry{$J(P_t)$}
 \addplot[c2, dotted, very thick] table[row sep=\\]{
        x y \\
0 0.015625 \\
1 0.109375 \\
2 0.75 \\
3 0.984375 \\
4 0.96875 \\
5 0.96875 \\
6 0.96875 \\
7 0.96875 \\
8 0.96875 \\
};
\addlegendentry{$P_{t,1}(s_{1,m}|s_{0,m})$}
 \addplot[c3, dotted, very thick] table[row sep=\\]{
        x y \\
    0 0.140625 \\
1  0.28125 \\
2 0.796875 \\
3 0.984375 \\
4 0.984375 \\
5 0.984375 \\
6 0.984375 \\
7 0.984375 \\
8 0.984375 \\
};
\addlegendentry{$P_{t,2}(s_{2,m}|s_{1,m})$}
 \addplot[c4, dotted, very thick] table[row sep=\\]{
        x y \\
    0 0.203125 \\
1 0.515625 \\
2 0.953125 \\
3 1.0 \\
4 0.984375 \\
5 0.984375 \\
6 0.984375 \\
7 0.984375 \\
8 0.984375 \\
};
\addlegendentry{$P_{t,3}(s_{3,m}|s_{2,m})$}
\end{axis}
    \end{tikzpicture}
} &
    \resizebox{0.22\textwidth}{!}{
    \begin{tikzpicture}
    \begin{axis}[
    width=6cm,
    height=5cm,
    legend pos=south east,
    grid=major,
    grid style={dashed,gray!30},
    xmin=0, xmax=8,
    ymin=-0.1, ymax=1.1,
    font=\small,
    legend cell align={left},
    title style={font=\small},
    label style={font=\small},
    tick label style={font=\small},
    legend style={font=\scriptsize},
    xlabel={$t$},
    ylabel={$J(P_t)$},
    xlabel style={
        at={(current axis.south east)}, % Relative positioning
        anchor=north east,              % Anchoring at a specific point
        yshift=0pt,                   % Shifting downward
        xshift=10pt                      % Shifting rightward (if needed)
    },
    ylabel style={
        at={(current axis.north west)}, % Relative positioning
        anchor=north east,              % Anchoring at a specific point
        yshift=15pt,                   % Shifting downward
        xshift=20pt                      % Shifting rightward (if needed)
    },
   title={\texttt{dec2bin}, GPT-2, $N=5$},
    ]
    \addplot[c11, dashed, thick] table[row sep=\\]{
x y \\ 
 0  0.0318756103515625 \\ 
 1  0.0362548828125 \\ 
 2  0.2811126708984375 \\ 
 3  0.8050079345703125 \\ 
 4  0.8755950927734375 \\ 
 5  0.96875 \\ 
 6  1.0 \\ 
 7  1.0 \\ 
 8  1.0 \\ 
 9  1.0 \\ 
};
\addlegendentry{$J(P_t)$}
\addplot[c12, dotted, very thick] table[row sep=\\]{
x y \\ 
 0  0.0312 \\ 
 1  0.0625 \\ 
 2  0.4062 \\ 
 3  0.75 \\ 
 4  0.9062 \\ 
 5  0.9688 \\ 
 6  1.0 \\ 
 7  1.0 \\ 
 8  1.0 \\ 
 9  1.0 \\ 
};
\addlegendentry{$P_{t,1}(s_{1,m}|s_{0,m})$}
\addplot[c13, dotted, very thick] table[row sep=\\]{
x y \\ 
 0  0.1875 \\ 
 1  0.5312 \\ 
 2  0.8438 \\ 
 3  0.9375 \\ 
 4  0.9688 \\ 
 5  1.0 \\ 
 6  1.0 \\ 
 7  1.0 \\ 
 8  1.0 \\ 
 9  1.0 \\ 
};
\addlegendentry{$P_{t,2}(s_{2,m}|s_{1,m})$}
\addplot[c14, dotted, very thick] table[row sep=\\]{
x y \\ 
 0  0.3125 \\ 
 1  0.5 \\ 
 2  0.9688 \\ 
 3  0.9375 \\ 
 4  1.0 \\ 
 5  1.0 \\ 
 6  1.0 \\ 
 7  1.0 \\ 
 8  1.0 \\ 
 9  1.0 \\ 
};
\addlegendentry{$P_{t,3}(s_{3,m}|s_{2,m})$}
\addplot[c15, dotted, very thick] table[row sep=\\]{
x y \\ 
 0  0.2188 \\ 
 1  0.5 \\ 
 2  0.8438 \\ 
 3  0.9688 \\ 
 4  0.9688 \\ 
 5  1.0 \\ 
 6  1.0 \\ 
 7  1.0 \\ 
 8  1.0 \\ 
 9  1.0 \\ 
};
\addlegendentry{$P_{t,4}(s_{4,m}|s_{3,m})$}
\addplot[c16, dotted, very thick] table[row sep=\\]{
x y \\ 
 0  0.375 \\ 
 1  0.6562 \\ 
 2  0.9375 \\ 
 3  1.0 \\ 
 4  1.0 \\ 
 5  1.0 \\ 
 6  1.0 \\ 
 7  1.0 \\ 
 8  1.0 \\ 
 9  1.0 \\ 
};
\addlegendentry{$P_{t,5}(s_{5,m}|s_{4,m})$}
\end{axis}
\end{tikzpicture}
}
    \end{tabular}
    \caption{Empirical values of $J(P_t)$ when $\delta_{0,1}=0$ but $\delta_{0,n'}=0.2$ for $n'>1$.}
\label{fig:main_experiment_b} 
\end{figure}
\begin{figure*}[hbt!]
    \centering
    \begin{tabular}{c c}
    \resizebox{0.46\textwidth}{!}{
    \begin{tikzpicture}
\begin{axis}[
width=5.5cm,
    height=3.5cm,
    legend pos=outer north east,
    grid=major,
    grid style={dashed,gray!30},
    xmin=0, xmax=9,
    ymin=-0.1, ymax=1.1,
    font=\footnotesize,
    legend cell align={left},
    title style={font=\scriptsize},
    label style={font=\scriptsize},
    tick label style={font=\scriptsize},
    legend style={font=\tiny},
    xlabel={$t$},
    ylabel={$J(P_t)$},
    xlabel style={
        at={(current axis.south east)}, % Relative positioning
        anchor=north east,              % Anchoring at a specific point
        yshift=0pt,                   % Shifting downward
        xshift=0pt                      % Shifting rightward (if needed)
    },
    ylabel style={
        at={(current axis.north west)}, % Relative positioning
        anchor=north east,              % Anchoring at a specific point
        yshift=15pt,                   % Shifting downward
        xshift=20pt                      % Shifting rightward (if needed)
    },
   title={\texttt{zip}, GPT-2, $N=3$},
    ]
    \addplot[c1, dashed, thick] table[row sep=\\]{
x y \\ 
 0  0.0153656005859375 \\ 
 1  0.016574859619140625 \\ 
 2  0.02802276611328125 \\ 
 3  0.18509674072265625 \\ 
 4  0.3858299255371094 \\ 
 5  0.3455047607421875 \\ 
 6  0.3613319396972656 \\ 
 7  0.3485145568847656 \\ 
 8  0.3851127624511719 \\ 
 9  0.3891181945800781 \\ 
};
\addlegendentry{$J(P_t)$}
\addplot[c2, dotted, very thick] table[row sep=\\]{
x y \\ 
 0  0.0 \\ 
 1  0.0156 \\ 
 2  0.0 \\ 
 3  0.0 \\ 
 4  0.0312 \\ 
 5  0.0312 \\ 
 6  0.0312 \\ 
 7  0.0312 \\ 
 8  0.0625 \\ 
 9  0.0625 \\ 
};
\addlegendentry{$P_{t,1}(s_{1,m}|s_{0,m})$}
\addplot[c3, dotted, very thick] table[row sep=\\]{
x y \\ 
 0  0.0156 \\ 
 1  0.0 \\ 
 2  0.0 \\ 
 3  0.0156 \\ 
 4  0.0469 \\ 
 5  0.0312 \\ 
 6  0.0312 \\ 
 7  0.0469 \\ 
 8  0.0938 \\ 
 9  0.0781 \\ 
};
\addlegendentry{$P_{t,2}(s_{2,m}|s_{1,m})$}
\addplot[c4, dotted, very thick] table[row sep=\\]{
x y \\ 
 0  0.1562 \\ 
 1  0.4219 \\ 
 2  0.7188 \\ 
 3  0.8438 \\ 
 4  0.8594 \\ 
 5  0.9062 \\ 
 6  0.9219 \\ 
 7  0.8438 \\ 
 8  0.8906 \\ 
 9  0.875 \\ 
};
\addlegendentry{$P_{t,3}(s_{3,m}|s_{2,m})$}
\end{axis}
\end{tikzpicture}
} &
    \resizebox{0.46\textwidth}{!}{
    \begin{tikzpicture}
\begin{axis}[
    width=5.5cm,
    height=3.5cm,
    legend pos=outer north east,
    grid=major,
    grid style={dashed,gray!30},
    xmin=0, xmax=9,
    ymin=-0.1, ymax=1.1,
    font=\footnotesize,
    legend cell align={left},
    title style={font=\scriptsize},
    label style={font=\scriptsize},
    tick label style={font=\scriptsize},
    legend style={font=\tiny},
    xlabel={$t$},
    ylabel={$J(P_t)$},
    xlabel style={
        at={(current axis.south east)}, % Relative positioning
        anchor=north east,              % Anchoring at a specific point
        yshift=0pt,                   % Shifting downward
        xshift=0pt                      % Shifting rightward (if needed)
    },
    ylabel style={
        at={(current axis.north west)}, % Relative positioning
        anchor=north east,              % Anchoring at a specific point
        yshift=15pt,                   % Shifting downward
        xshift=20pt                      % Shifting rightward (if needed)
    },
   title={\texttt{dec2bin}, GPT-2, $N=5$},
    ]
    \addplot[c11, dashed, thick] table[row sep=\\]{
x y \\ 
 0  0.03133392333984375 \\ 
 1  0.0320892333984375 \\ 
 2  0.05628204345703125 \\ 
 3  0.3006591796875 \\ 
 4  0.541473388671875 \\ 
 5  0.6111679077148438 \\ 
 6  0.6251602172851562 \\ 
 7  0.6251449584960938 \\ 
 8  0.624786376953125 \\ 
 9  0.62506103515625 \\ 
};
\addlegendentry{$J(P_t)$}
\addplot[c12, dotted, very thick] table[row sep=\\]{
x y \\ 
 0  0.0312 \\ 
 1  0.0312 \\ 
 2  0.0312 \\ 
 3  0.0312 \\ 
 4  0.0312 \\ 
 5  0.0312 \\ 
 6  0.0312 \\ 
 7  0.0312 \\ 
 8  0.0312 \\ 
 9  0.0312 \\ 
};
\addlegendentry{$P_{t,1}(s_{1,m}|s_{0,m})$}
\addplot[c13, dotted, very thick] table[row sep=\\]{
x y \\ 
 0  0.0 \\ 
 1  0.0312 \\ 
 2  0.0625 \\ 
 3  0.0625 \\ 
 4  0.0625 \\ 
 5  0.0625 \\ 
 6  0.0625 \\ 
 7  0.0938 \\ 
 8  0.0625 \\ 
 9  0.0625 \\ 
};
\addlegendentry{$P_{t,2}(s_{2,m}|s_{1,m})$}
\addplot[c14, dotted, very thick] table[row sep=\\]{
x y \\ 
 0  0.3438 \\ 
 1  0.5625 \\ 
 2  0.9062 \\ 
 3  0.9688 \\ 
 4  1.0 \\ 
 5  1.0 \\ 
 6  0.9375 \\ 
 7  1.0 \\ 
 8  1.0 \\ 
 9  0.9375 \\ 
};
\addlegendentry{$P_{t,3}(s_{3,m}|s_{2,m})$}
\addplot[c15, dotted, very thick] table[row sep=\\]{
x y \\ 
 0  0.2188 \\ 
 1  0.4062 \\ 
 2  0.875 \\ 
 3  0.9688 \\ 
 4  0.9688 \\ 
 5  0.9688 \\ 
 6  0.9688 \\ 
 7  1.0 \\ 
 8  1.0 \\ 
 9  1.0 \\ 
};
\addlegendentry{$P_{t,4}(s_{4,m}|s_{3,m})$}
\addplot[c16, dotted, very thick] table[row sep=\\]{
x y \\ 
 0  0.4375 \\ 
 1  0.6875 \\ 
 2  0.9688 \\ 
 3  1.0 \\ 
 4  1.0 \\ 
 5  1.0 \\ 
 6  1.0 \\ 
 7  1.0 \\ 
 8  1.0 \\ 
 9  1.0 \\ 
};
\addlegendentry{$P_{t,5}(s_{5,m}|s_{4,m})$}
\end{axis}
\end{tikzpicture}
}
    \end{tabular}
    \caption{Empirical values of $J(P_t)$ when $\delta_{0,1}=\delta_{0,2}=0$ but $\delta_{0,n'}=0.2$ for $n'>2$.}
\label{fig:main_experiment_c} 
\end{figure*}
\begin{figure*}[hbt!]
    \centering
    \begin{tabular}{c c c}
    \resizebox{0.30\textwidth}{!}{
    \begin{tikzpicture}
\begin{axis}[
    width=6cm,
    height=3.8cm,
    legend pos=south east,
    grid=major,
    grid style={dashed,gray!30},
    xmin=0, xmax=7,
    ymin=-0.1, ymax=1.1,
    font=\scriptsize,
    legend cell align={left},
    title style={font=\scriptsize},
    label style={font=\scriptsize},
    tick label style={font=\scriptsize},
    legend style={font=\tiny},
    xlabel={$t$},
    ylabel={$J(P_t)$},
    xlabel style={
        at={(current axis.south east)}, % Relative positioning
        anchor=north east,              % Anchoring at a specific point
        yshift=0pt,                   % Shifting downward
        xshift=0pt                      % Shifting rightward (if needed)
    },
    ylabel style={
        at={(current axis.north west)}, % Relative positioning
        anchor=north east,              % Anchoring at a specific point
        yshift=15pt,                   % Shifting downward
        xshift=20pt                      % Shifting rightward (if needed)
    },
   title={\texttt{zip}, GPT-2, $\delta_0=0.1$},
    ]
 \addplot[c1, solid, very thick] table[row sep=\\]{
        x y \\
0    0.01665699798437894 \\
1    0.020123459753581558 \\
2    0.06557845281185505 \\
3    0.6649908499462328 \\
4    0.9978254678481339 \\
5    0.9999999497974984 \\
6    1.0 \\
7    1.0 \\
8    1.0 \\
9    1.0 \\
% 10    1.0 \\
};
\addlegendentry{Unif. $\delta_0$, Theory}
 \addplot[c2, dashed, thick] table[row sep=\\]{
        x y \\
% 0 0.016307830810546875 \\
0 0.016307830810546875 \\
1 0.03815460205078125 \\
2 0.4627265930175781 \\
3 0.9492 \\
4 0.9957 \\
5 1.0 \\
6    1.0 \\
7    1.0 \\
8    1.0 \\
9    1.0 \\
% 5 1.0 \\
% 6 1.0 \\
};
\addlegendentry{Unif. $\delta_0$, Exp.}
 \addplot[c3, dashed, thick] table[row sep=\\]{
        x y \\
% 0 0.016307830810546875 \\
%0 0.020603179931640625 \\ 
%1 0.25815460205078125 \\ 
0 0.020603179931640625 \\
1 0.16170883178710938 \\
2 0.9560050964355469 \\
3 1.0 \\
4 1.0 \\
5 1.0 \\
6 1.0 \\
7 1.0 \\
};
\addlegendentry{Non-Unif. $\delta_{0}$, Exp.}
\end{axis}
\end{tikzpicture}
} &
    \resizebox{0.30\textwidth}{!}{
    \begin{tikzpicture}
\begin{axis}[
    width=6cm,
    height=3.8cm,
    legend pos=south east,
    grid=major,
    grid style={dashed,gray!30},
    xmin=0, xmax=7,
    ymin=-0.1, ymax=1.1,
    font=\scriptsize,
    legend cell align={left},
    title style={font=\scriptsize},
    label style={font=\scriptsize},
    tick label style={font=\scriptsize},
    legend style={font=\tiny},
    xlabel={$t$},
    ylabel={$J(P_t)$},
    xlabel style={
        at={(current axis.south east)}, % Relative positioning
        anchor=north east,              % Anchoring at a specific point
        yshift=0pt,                   % Shifting downward
        xshift=0pt                      % Shifting rightward (if needed)
    },
    ylabel style={
        at={(current axis.north west)}, % Relative positioning
        anchor=north east,              % Anchoring at a specific point
        yshift=15pt,                   % Shifting downward
        xshift=20pt                      % Shifting rightward (if needed)
    },
   title={\texttt{zip}, GPT-2, $\delta_0=0.2$},
    ]
 \addplot[c1, solid, thick] table[row sep=\\]{
        x y \\
0    0.0238809838750315 \\
1   0.1387970147215855 \\
2    0.8805859938624284 \\
3    0.9998173897891941 \\
4    0.99999999964703 \\
5    1.0 \\
6    1.0 \\
7    1.0 \\
8    1.0 \\
9    1.0 \\
};
\addlegendentry{Unif. $\delta_0$, Theory}
\addplot[c2, dotted, very thick] table[row sep=\\]{
        x y \\
0 0.028106689453125 \\
1 0.4643058776855469 \\
2 0.9990997314453125 \\
3 1.0 \\
4 1.0 \\
5 1.0 \\
6    1.0 \\
7    1.0 \\
8    1.0 \\
9    1.0 \\
};
\addlegendentry{Unif. $\delta_0$, Exp.}
 \addplot[c3, dashed, thick] table[row sep=\\]{
        x y \\
% 0 0.016307830810546875 \\
0 0.06591033935546875 \\
1 0.8716621398925781 \\
2 0.9999923706054688 \\ 
3 1.0 \\ 
4 1.0 \\ 
5 1.0 \\ 
6    1.0 \\
7    1.0 \\
8    1.0 \\
9    1.0 \\
% 5 1.0 \\
% 6 1.0 \\
};
\addlegendentry{Non-Unif. $\delta_{0}$, Exp.}
\end{axis}
\end{tikzpicture}
} &
    \resizebox{0.30\textwidth}{!}{
    \begin{tikzpicture}
\begin{axis}[
    width=6cm,
    height=3.8cm,
    legend pos=south east,
    grid=major,
    grid style={dashed,gray!30},
    xmin=0, xmax=7,
    ymin=-0.1, ymax=1.1,
    font=\scriptsize,
    legend cell align={left},
    title style={font=\scriptsize},
    label style={font=\scriptsize},
    tick label style={font=\scriptsize},
    legend style={font=\tiny},
    xlabel={$t$},
    ylabel={$J(P_t)$},
    xlabel style={
        at={(current axis.south east)}, % Relative positioning
        anchor=north east,              % Anchoring at a specific point
        yshift=0pt,                   % Shifting downward
        xshift=0pt                      % Shifting rightward (if needed)
    },
    ylabel style={
        at={(current axis.north west)}, % Relative positioning
        anchor=north east,              % Anchoring at a specific point
        yshift=15pt,                   % Shifting downward
        xshift=20pt                      % Shifting rightward (if needed)
    },
   title={\texttt{zip}, GPT-2, $\delta_0=0.3$},
    ]
 \addplot[c1, solid, thick] table[row sep=\\]{
        x y \\
0    0.042203124999999994 \\
1    0.4548791787691011 \\
2    0.990080056214512 \\
3    0.9999989429078555 \\
4    0.9999999999999883 \\
5    0.9999999999999997 \\
6    1.0 \\
7    1.0 \\
8    1.0 \\
9    1.0 \\
};
\addlegendentry{Unif. $\delta_0$, Theory}
\addplot[c2, dotted, very thick] table[row sep=\\]{
        x y \\
0 0.06693649291992188 \\ 
1 0.894866943359375 \\
2 1.0 \\
3 1.0 \\
4 1.0 \\
5 1.0 \\
6 1.0 \\
7 1.0 \\
% 1 0.6643058776855469 \\
};
\addlegendentry{Unif. $\delta_0$, Exp.}
 \addplot[c3, dashed, thick] table[row sep=\\]{
        x y \\
% 0 0.016307830810546875 \\
%0 0.20191574096679688 \\
% 1 0.8716621398925781 \\
0 0.20191574096679688 \\
1 0.9924507141113281 \\
2 1.0 \\
3 1.0 \\
4 1.0 \\
5 1.0 \\
6 1.0 \\
7 1.0 \\
};
\addlegendentry{Non-Unif. $\delta_{0}$, Exp.}
\end{axis}
\end{tikzpicture}
}
    \end{tabular}
    \caption{Empirical value of $J(P_t)$ under non-uniform $\delta_{0}$ compared to the value of $J(P_t)$ under uniform settings.}
\label{fig:exp_non_uniform_delta} 
\end{figure*}
\begin{figure}[hbt!]
    \centering
    \begin{tabular}{c c}
    % \resizebox{0.3\textwidth}{!}{
    % \input{acc_exp5_delta_01}} &
    \resizebox{0.22\textwidth}{!}{
    \begin{tikzpicture}
\begin{axis}[
    width=5.5cm,
    height=4.5cm,
    legend pos=south east,
    grid=major,
    grid style={dashed,gray!30},
    xmin=0, xmax=7,
    ymin=-0.1, ymax=1.1,
    font=\small,
    legend cell align={left},
    title style={font=\small},
    label style={font=\small},
    tick label style={font=\small},
    legend style={font=\scriptsize},
    xlabel={$t$},
    ylabel={$J(P_t)$},
    xlabel style={
        at={(current axis.south east)}, % Relative positioning
        anchor=north east,              % Anchoring at a specific point
        yshift=0pt,                   % Shifting downward
        xshift=0pt                      % Shifting rightward (if needed)
    },
    ylabel style={
        at={(current axis.north west)}, % Relative positioning
        anchor=north east,              % Anchoring at a specific point
        yshift=15pt,                   % Shifting downward
        xshift=20pt                      % Shifting rightward (if needed)
    },
   title={\texttt{binadd}, GPT-2, $\delta_0=0.2$},
    ]
 \addplot[c1, solid, thick] table[row sep=\\]{
        x y \\
0    0.0238809838750315 \\
1   0.1387970147215855 \\
2    0.8805859938624284 \\
3    0.9998173897891941 \\
4    0.99999999964703 \\
5    1.0 \\
6    1.0 \\
7    1.0 \\
8    1.0 \\
9    1.0 \\
};
\addlegendentry{$\texttt{const-M}$, Theory}
\addplot[c2, dotted, very thick] table[row sep=\\]{
        x y \\
0 0.00324249267578125 \\
1 0.3540992736816406 \\
2 0.8908615112304688 \\
3 0.9427757263183594 \\
4 0.9809684753417969 \\
5 0.9693222045898438 \\
6 0.9534568786621094 \\
7 0.975738525390625 \\
};
\addlegendentry{$\texttt{const-M}$, Exp.}
 \addplot[c3, dashed, thick] table[row sep=\\]{
        x y \\
% 0 0.016307830810546875 \\
0 0.10503387451171875 \\
1 0.3212165832519531 \\ 
2 0.7312355041503906 \\ 
3 0.937164306640625 \\ 
4 0.9836578369140625 \\ 
5 0.984375 \\ 
6 0.984375 \\ 
7 0.984375 \\ 
};
\addlegendentry{$\texttt{non-const-M}$, Exp.}
\end{axis}
\end{tikzpicture}
} &
        \resizebox{0.22\textwidth}{!}{
    \begin{tikzpicture}
\begin{axis}[
    width=5.5cm,
    height=4.5cm,
    legend pos=south east,
    grid=major,
    grid style={dashed,gray!30},
    xmin=0, xmax=7,
    ymin=-0.1, ymax=1.1,
    font=\small,
    legend cell align={left},
    title style={font=\small},
    label style={font=\small},
    tick label style={font=\small},
    legend style={font=\scriptsize},
    xlabel={$t$},
    ylabel={$J(P_t)$},
    xlabel style={
        at={(current axis.south east)}, % Relative positioning
        anchor=north east,              % Anchoring at a specific point
        yshift=0pt,                   % Shifting downward
        xshift=0pt                      % Shifting rightward (if needed)
    },
    ylabel style={
        at={(current axis.north west)}, % Relative positioning
        anchor=north east,              % Anchoring at a specific point
        yshift=15pt,                   % Shifting downward
        xshift=20pt                      % Shifting rightward (if needed)
    },
   title={\texttt{binadd}, GPT-2, $\delta_0=0.3$},
    ]
 \addplot[c1, solid, thick] table[row sep=\\]{
        x y \\
0    0.042203124999999994 \\
1    0.4548791787691011 \\
2    0.990080056214512 \\
3    0.9999989429078555 \\
4    0.9999999999999883 \\
5    0.9999999999999997 \\
6    1.0 \\
7    1.0 \\
8    1.0 \\
9    1.0 \\
};
\addlegendentry{$\texttt{const-M}$, Theory}
\addplot[c2, dotted, very thick] table[row sep=\\]{
        x y \\
% 0 0.028106689453125 \\ % Fake
% 1 0.4643058776855469 \\ % Fake
0 0.0129127502441406251 \\
1 0.891754150390625 \\
2 0.9994583129882812  \\
3 1.0  \\
4 1.0  \\
5 1.0  \\
6    1.0 \\
7    1.0 \\
8    1.0 \\
9    1.0 \\
};
\addlegendentry{$\texttt{const-M}$, Exp.}
 \addplot[c3, dashed, thick] table[row sep=\\]{
        x y \\
% 0 0.016307830810546875 \\
0 0.169464111328125 \\
1 0.7858238220214844 \\
2 0.9994392395019531 \\
3 1.0  \\
4 1.0  \\
5 1.0  \\
6    1.0 \\
7    1.0 \\
8    1.0 \\
9    1.0 \\
};
\addlegendentry{$\texttt{non-const-M}$, Exp.}
\end{axis}
\end{tikzpicture}
}
    \end{tabular}
    \caption{Empirical value of $J(P_t)$ under non-constant $M$  compared to the value of $J(P_t)$ under constant $M$ settings.}
\label{fig:exp_non_constant_M} 
\end{figure}

\paragraph{Results}
Fig.~\ref{fig:main_experiment_b} presents the results. In this figure, only one \(\delta_{0,n}\) is zero, and \(J(P_t)\) converges to a nearly optimal value, roughly consistent with Theorem~\ref{theorem:main_theorem}\ref{theorem:item_d2}. Besides \(J(P_t)\), we show the probability of pre-trained model $P_t$ generating the correct transition $(s_{n,m}, s_{n-1,m})$ for each CoT step \(n\), denoted by \(P_{t,n}(s_{n,m}|s_{n-1,m})\), across iterations \(t\) of RL-STaR.
Notably, in both tasks, the probability of transition generated by the pre-trained model \( P_{0,n}(s_{n,m}|s_{n-1,m})\) does not perfectly align with the distributions in the pre-training dataset. For instance, \(P_{0,2}(s_{2,m}|s_{1,m})\) is lower than \(P_{0,3}(s_{3,m}|s_{2,m})\) even though \(\delta_{0,2}\) and \(\delta_{0,3}\) share the same value in the pre-training dataset. Moreover, under the condition of Theorem~\ref{theorem:main_theorem}\ref{theorem:item_d2}, the values of \(P_{t,2}(s_{2,m}|s_{1,m})\) and \(P_{t,3}(s_{3,m}|s_{2,m})\) should remain unchanged between \(t=0\) and \(t=1\). In practice, however, they increase on the first iteration. These differences reflect the discrepancy between the theoretical and empirical values observed in the previous experiment.

\subsection{Experiment of the Convergence of $J(P_t)$ with Two $\delta_{0,n}=0$ and $\delta_{0,n'}=0$}\label{sec:experiment_C3}
In this experiment, we examine the convergence behavior of $J(P_t)$ when two $\delta_{0,n}=0$ and $\delta_{0,n'}=0$. We apply the setting of $\delta_{0,1}=\delta_{0,2}=0$ but $\delta_{0,n'}=0.2$ for $n'>2$. 
We use GPT-2 to perform both the \texttt{zip} and \texttt{dec2bin} tasks.  All other configurations remain as detailed in Sec.~\ref{sec:experiment_C2}.

\paragraph{Results}
Fig.~\ref{fig:main_experiment_c} shows that when $\delta_{0,1} = \delta_{0,2} = 0$, $J(P_t)$ fails to converge to its near-optimal value, in agreement with Theorem~\ref{theorem:main_theorem}\ref{theorem:item_d3}. Besides $J(P_t)$, we also plot
$P_{t,n}\bigl(s_{n,m}\mid s_{n-1,m}\bigr)$
—the probability that the pre-trained policy $P_t$ correctly generates the transition $(s_{n,m},s_{n-1,m})$ at each CoT step $n$—over RL-STaR iterations $t$.
Although Theorem~\ref{theorem:main_theorem}\ref{theorem:item_d3} predicts that all $\delta_{t,n}$ remain constant (implying both $J(P_t)$ and each $P_{t,n}$ should be invariant in $t$), the empirical curves for both tasks show that if $\delta_{0,n}>0$, then $P_{t,n}(s_{n,m}\mid s_{n-1,m})$ can still approach 1, causing $J(P_t)$ to increase. The cause is the one identified in Sec.~\ref{sec:experiment_A}: $P_{t,n}$ does not reproduce $\mathcal{D}_t$ exactly, but puts extra probability mass on the transitions that occur most often in $\mathcal{D}_t$. This lifts $\delta_{t,n}$ at the steps where $\delta_{0,n}>0$, whereas Assumption~\ref{assump:exact_fit} requires exact matching, so Theorem~\ref{theorem:main_theorem}\ref{theorem:item_d3} predicts no change.

\subsection{Experiment of Non-Uniform $\delta_{t,n}$}\label{sec:experiment_non_uniform_delta}
In this experiment, we follow the setup in Sec.~\ref{sec:experiment_A}, but for uniform case we choose 
$
\delta_{0,n}\in\{0.1,0.2,0.3\}$ for all $n\in[N].
$
For the non-uniform case, each off-diagonal transition \((s_{n-1,m},s_{n,m'})\), \(m'\neq m\), uses
\[
\delta_{0,n,m,m'}\sim \mathrm{Uniform}[\delta_{0,n},2\delta_{0,n}],
\]
ensuring \(\delta_{0,n}=\min_{m,m'}\delta_{0,n,m,m'}\). 
\paragraph{Results}
Fig.~\ref{fig:exp_non_uniform_delta} shows that \(J(P_t)\) converges faster under non-uniform \(\delta_0\) than both the empirical and theoretical uniform-case curves for $\delta_0=\{0.1,0.2,0.3\}$. Thus, the uniform-case convergence rate provides a valid lower bound for the non-uniform scenario.

\subsection{Experiment of Non-Constant $M$}\label{sec:experiment_non_constant_M}
In this experiment, we follow the setup in Sec.~\ref{sec:experiment_A}, but use different tasks. We choose $\delta_0=\{0.2,0.3\}$ for these tasks. To have different number of states at each step, we use the following task.
\paragraph{Task `binadd-non-const-M'}
This task is a task of binary addition.
$$
\begin{aligned}
s_0 = \texttt{x:101+110} 
& \Rightarrow  s_1 = \texttt{x:10+11},\texttt{z:0},\texttt{y:1}  \\
& \Rightarrow  s_2 = \texttt{x:1+1},\texttt{z:0},\texttt{y:11} \\
& \Rightarrow  s_3 = \texttt{y:1011}.
\end{aligned}
$$
The input consists of two binary numbers $x=101,110$. In the first step, the last digit $1,1$ produces a carry $z=0$ and a sum $y=1$. In the second step, the second digit $0,1$ and carry $z=0$ yield a new $z=0$ and the result of the second digit $y=11$, continuing in this manner until the final step results in $y=1011$. It can be shown that $M_0=64,M_1=48,M_2=28,M_3=15$.

\paragraph{Task `binadd-const-M'}
This task is a task of binary addition, but with constant $M$. We create this task by concatenating $s_0$ to $s_1,s_2$ and $s_3$, as follows:
$$
\begin{aligned}
 s_0 &= \texttt{x:101+110} \\
&\Rightarrow  s_1 = \texttt{x:101+110,x:10+11},\texttt{z:0},\texttt{y:1}   \\
&\Rightarrow  s_2 = \texttt{x:101+110,x:1+1},\texttt{z:0},\texttt{y:11} \\ 
&\Rightarrow  s_3 = \texttt{x:101+110,y:1011}.
\end{aligned}
$$
We compare the convergence of $J(P_t)$ between these two tasks, and verify whether the task $\texttt{binadd-const-M}$ can be an approximator of the task $\texttt{binadd-non-const-M}$.

\paragraph{Results}
 In Fig.~\ref{fig:exp_non_constant_M}, we observe that the convergence curve  $J(P_t)$ in the case of $\texttt{non-constant-M}$ can be approximated by the $J(P_t)$ of $\texttt{constant-M}$ in both empirical values and theoretical values. This validates our hypothesis that the modified case can be an approximator.
\section{Limitations and Possible Extensions}
\label{sec:more_about_limitation}
While our theorem provides useful theoretical guarantees, it also has limitations when applied to real-world LLM settings. These limitations include the \textit{Markov assumption of state transitions}, the \textit{uniformity of $\delta_{t,n}$}, and the \textit{fixed state count $M$}. Below, we discuss these restrictions and outline how our framework could be extended to better align with practical LLM behavior.

\subsection{Markov Assumption of State Transitions}
\label{sec:more_about_limitation_non_markov}  
We simplify our analysis by modeling the reasoning process as Markovian, even though this assumption may not strictly hold for real-world LLMs, whose context windows can span the entire trajectory and thus violate Markov independence. In this section, we first show that the Markov assumption is essential for filtering out incorrect reasoning steps, and we discuss how practical LLMs can approximate this behavior.
% Moreover, prior studies (e.g., \citep{prystawski2024think,xiao2024theory,li2024trainingnonlineartransformerschainofthought,yao2025unveilingmechanismsexplicitcot, teng2025atomthoughtsmarkovllm, wang2025indistributionsuccessscalingcurves}) also demonstates that the approximate Markov property is fundamental to CoT generalization.
\paragraph{Toy Example without Markov Assumption}
Suppose the pre-trained transition at step $2$ of the toy example in Sec.~\ref{sec:toy_example} reads both $S_1$ and $S_0$ instead of $S_1$ alone. Working through one iteration of Algorithm~\ref{algo:rl_star} (Appendix~\ref{sec:non_markov_toy_example}), the transition $P_{1,2}$ learned in the first iteration is already deterministic, so $\mathcal{D}_T=\mathcal{D}_{T-1}=\cdots=\mathcal{D}_1$. Trajectories of the form $\tau=(s_{0,m},s_{1,m^{\prime}\neq m},s_{2,m})$, which contain an incorrect reasoning step, therefore stay in the dataset at every iteration and RL-STaR cannot remove them.

\paragraph{How Real-world LLMs Approximate Markov Property}  
Although real-world Chain-of-Thought trajectories include the entire context \((S_0, S_1, \dots, S_n)\), predicting \(S_{n+1}\) typically relies only on a few recent steps. Several studies (e.g., \citep{yao2025unveilingmechanismsexplicitcot,teng2025atomthoughtsmarkovllm,wang2025indistributionsuccessscalingcurves,li2024trainingnonlineartransformerschainofthought,xiao2024theory,prystawski2024think}) demonstrate that CoT generalization stems from decomposing complex tasks into simpler, reusable subtasks. Each subtask’s outcome depends on a limited window of prior reasoning steps, so the training data effectively exhibits a Markovian structure. Consequently, when LLMs focus on these local dependencies and ignore unnecessary earlier context, they approximate the Markov property in practice.

% \todo{Read \citep{}}
% \todo{}
% LLMs' attention selectively exclude 

% $$
% p(\tau|\tau\in \mathcal{D}_1)=
% \begin{cases}
%  \frac{(1+\delta_{0,1})(1+\delta_{0,2})}{4(1+\delta_{0,1}\delta_{0,2})} & \text{if }\tau=(s_{0,1},s_{1,1},s_{2,1}), \\
%  \frac{(1-\delta_{0,1})(1-\delta_{0,2})}{4(1+\delta_{0,1}\delta_{0,2})} & \text{if }\tau=(s_{0,1},s_{1,2},s_{2,1}),  \\
%  \frac{(1-\delta_{0,1})(1-\delta_{0,2})}{4(1+\delta_{0,1}\delta_{0,2})} & \text{if }\tau=(s_{0,2},s_{1,1},s_{2,2}),  \\
%  \frac{(1+\delta_{0,1})(1+\delta_{0,2})}{4(1+\delta_{0,1}\delta_{0,2})} & \text{if }\tau=(s_{0,2},s_{1,2},s_{2,2}).  \\
% \end{cases}
% $$

\subsection{Uniformity of $\delta_{t,n}$}
\label{sec:more_about_limitation_non_uniform_delta}
We assume that \(\delta_{0,n}\) is uniform across all states at step \(n\), but real pre-trained LLMs rarely satisfy this idealization. Allowing \(\delta_{0,n,m,m'}\) for each transition from $s_{n-1,m}$ to $s_{n,m'}$ to take various values complicates the analysis substantially. However, we will show that it is possible to extend our theorem to non-uniform cases.
\paragraph{Non-Uniform \(\delta_{0,n}\)}
In this section, we show that even though our theorem does not directly apply, we can still derive a valid lower bound by comparing the original, non-uniform transition $P_{0,n}$ to a worst-case uniform model $\underline{P}_{0,n}$. We begin by defining
$$
\begin{cases}
P_{0,n}\bigl(S_n=s_{n,m'}\mid S_{n-1}=s_{n-1,m}\bigr) =\frac{1-\delta_{0,n,m,m'}}{M} \text{for all $m'\neq m$, and $m,m'\in[M]$,} \\
P_{0,n}\bigl(S_n=s_{n,m}\mid S_{n-1}=s_{n-1,m}\bigr) = \frac{1+\sum_{m'\in[M] ~\text{and}~m' \neq m}\delta_{0,n,m,m'}}{M}.
\end{cases}
$$
Let
$\underline{\delta}_{0,n}
=\min_{m, m'\in[M]}\delta_{0,n,m,m'}$,
and define a uniform transition
$$
\begin{cases}
\underline{P}_{0,n}\bigl(S_n=s_{n,m'}\mid S_{n-1}=s_{n-1,m}\bigr) =\frac{1-\underline{\delta}_{0,n}}{M}
\quad \text{for all $m'\neq m$, and $m,m'\in[M]$}, \\
\underline{P}_{0,n}\bigl(S_n=s_{n,m}\mid S_{n-1}=s_{n-1,m}\bigr)  =\frac{1+(M-1)\,\underline{\delta}_{0,n}}{M}.
\end{cases}
$$
By construction, $\underline{P}_{0,n}$ is no more accurate than $P_{0,n}$ at selecting the correct step but does satisfy the uniform-$\delta$ condition.  Applying Theorem~\ref{theorem:main_theorem_delta_speed} to $\underline{P}_{0,n}$ gives the convergence rate of $\delta_{t,n}$ under uniform transitions, which therefore serves as a lower bound on the convergence speed for the original, non-uniform case. We validate this lower bound with the experimental results shown in Sec.~\ref{sec:experiment_non_uniform_delta}.

% \todo{Add Experiment}
% There are many cases for the convergence under non-uniformity of $\delta_{t,n}$. 
% For instance, consider the following example, where the ground-truth trajectories are \((s_{0,1}, s_{1,1}, s_{2,1})\) and \((s_{0,2}, s_{1,2}, s_{2,2})\). The pre-trained model’s transition probabilities (shown on each edge) still enable RL-STaR to discover the optimal policy:
% \[
% \begin{tikzcd}
% 	{s_{0,1}} &&& {s_{1,1}} &&& {s_{2,1}} \\
% 	{s_{0,2}} &&& {s_{1,2}} &&& {s_{2,2}}
% 	\arrow["{{0.1}}"{description}, from=1-1, to=1-4]
% 	\arrow["{{0.9}}"{description, pos=0.3}, from=1-1, to=2-4]
% 	\arrow["{{0.6}}"{description}, from=1-4, to=1-7]
% 	\arrow["{{0.4}}"{description, pos=0.3}, from=1-4, to=2-7]
% 	\arrow["{{0.4}}"{description, pos=0.3}, from=2-1, to=1-4]
% 	\arrow["{{0.6}}"{description}, from=2-1, to=2-4]
% 	\arrow["{{0.01}}"{description, pos=0.3}, from=2-4, to=1-7]
% 	\arrow["{{0.99}}"{description}, from=2-4, to=2-7].
% \end{tikzcd}\]
% Analyzing non-uniform \(\delta_{t,n}\) in general is significantly more challenging, so establishing the corresponding conditions for pre-trained models in this settings remains future

\subsection{Fixed Number of States $M$}
In our framework, each reasoning step is assumed to take one of \(M\) possible states. While this finite-state assumption is common in theoretical analyses of Chain-of-Thought reasoning (e.g., \citep{xiao2024theory,feng2024towards,yao2025unveilingmechanismsexplicitcot,wang2025indistributionsuccessscalingcurves}), it does not always hold in practice. Large language models can freely generate arbitrary strings rather than being confined to a fixed state set. Even if the model’s outputs are drawn from a bounded vocabulary, the effective number of states available at each step may still vary substantially.
% \subsubsection{Non-Constant  $M$}
% \label{sec:more_about_limitation_various_m}
\paragraph{LLMs Generate Arbitrary Strings}  In real-world deployments, LLMs can emit any text, not just tokens from a fixed set. For example, in an arithmetic task, the model might output Python code instead of a numerical expression. A plausible solution for this scenario is to introduce an extra state \(s'\) that represents all out-of-set outputs and redefine the state count as \(M' = M + 1\).  However, since \(s'\) does not lead to a valid terminal state, it complicates convergence and correctness guarantees. Extending the theoretical framework to accommodate this setting is an interesting direction for future work.

\paragraph{Different Number of States at Each Reasoning Step}
Some reasoning tasks may feature a different number of candidate states at each step. Let \(M_n\) denote the number of states at step \(n\in[N]\). We consider two scenarios:
\begin{itemize}
    \item \(M_n\) is nonincreasing: 
      $M_0 \ge M_1 \ge \cdots \ge M_N.$
    \item There exist indices \(n < n'\) such that 
    $M_n < M_{n'}$.
\end{itemize}

First, we focus on the first scenario. In this setting, we reformulate the problem by creating composite states: each state in the step $n$ of ground-truth trajectory as  
\[
\begin{aligned}
& s_{n,m_n}\,\in\,\operatorname{support}(S_n), \text{where}~
|\operatorname{support}(S_n)|=M_n,~\text{and}~
m_n\in[M_n],
\end{aligned}
\]
is paired with its initial state \(s_{0,m_0}\) to form a new state 
\(
\bigl(s_{0,m_0},\,s_{n,m_n}\bigr).
\) This construction is illustrated in the diagram below:
\[\begin{tikzcd}[sep=tiny]
	{s_{0,m_0}} && {s_{n,m_n}} && {s_{0,m_0}} && {(s_{0,m_0},s_{n,m_n})} \\
	{s_{0,m'_0}} && {} && {s_{0,m'_0}} && {(s_{0,m'_0},s_{n,m_n})} \\
	{s_{0,m''_0}} && {s_{n,m''_n}} && {s_{0,m''_0}} && {(s_{0,m''_0},s_{n,m''_n})}
	\arrow[dashed, from=1-1, to=1-3]
	\arrow[dashed, from=1-5, to=1-7]
	\arrow[dashed, from=2-1, to=1-3]
	\arrow[Rightarrow, from=2-3, to=2-5]
	\arrow[dashed, from=2-5, to=2-7]
	\arrow[dashed, from=3-1, to=3-3]
	\arrow[dashed, from=3-5, to=3-7]
\end{tikzcd}\]
Before concatenation, step 0 has \(M_0\) states and step \(n\) has \(M_n\). After concatenation, each composite state \((s_{0,m_0},s_{n,m_n})\) yields exactly \(M_0\) states at step \(n\), so we set \(M=M_0\) for all \(n\). By applying Theorem~\ref{theorem:main_theorem_delta_speed} to this constant-$M$ setting, we obtain a convergence rate that we conjecture approximates that of the original varying-$M$ problem.  In Sec.~\ref{sec:experiment_non_constant_M}, we empirically compare this approximated rate with the convergence behavior observed in the original task. However, owing to the challenges of analyzing the non-constant $M$ case, we present only empirical results and leave the theoretical analysis as future work.

For the second scenario, the ground-truth reasoning paths form a DAG. In this case, one state at step $n$ leads to multiple ground-truth transitions with non-zero probability at step $n+1$, which complicates the analysis. However, to illustrate how our framework can handle certain forms of trajectory multiplicity, consider the case where multiple correct intermediate steps exist but ultimately converge to the same final answer. For instance, suppose we are given two input questions, $s_{0,0} = 1+2+3$ and $s_{0,1} = 2+3+4$, each with multiple valid reasoning paths leading to their respective correct answers. In the following, we show that these intermediate trajectories can be merged into single abstract states $s_{1,0}$ and $s_{1,1}$, allowing our framework to operate over equivalence classes of states. In this example, we define the ground-truth reasoning steps and final answers as follows:
$$
\begin{aligned}
&s_{0,0}=1+2+3 \Rightarrow s_{1,0,0}=3+3 \Rightarrow s_{2,0}=6 \\
&s_{0,0}=1+2+3 \Rightarrow s_{1,0,1}=1+5 \Rightarrow s_{2,0}=6 \\
&s_{0,1}=2+3+4 \Rightarrow s_{1,1,0}=5+4 \Rightarrow s_{2,1}=9 \\
&s_{0,1}=2+3+4 \Rightarrow s_{1,1,1}=2+7 \Rightarrow s_{2,1}=9 \\
\end{aligned}
$$
And the transition probability is as follows:
$$
\begin{aligned}
& P(S_1|S_{0}) =\begin{cases}
    \frac{1+\delta_2}{4} &\text{if $S_1\in \{s_{1,m,0},s_{1,m,1}\}$ and $S_0=s_{0,m}$},\\
    \frac{1+\delta_2}{4} &\text{if $S_1\in \{s_{1,m,0},s_{1,m,1}\}$ and $S_0=s_{0,m'\neq m}$}, \\
\end{cases}
\end{aligned}
$$
and 
$$
\begin{aligned}
& P(S_2|S_{1}) =\begin{cases}
    \frac{1+\delta_1}{2} &\text{if $S_2=s_{2,m}$ and $S_1\in \{s_{1,m,0},s_{1,m,1}\}$},\\
    \frac{1+\delta_1}{2} &\text{if $S_2=s_{2,m'\neq m}$ and $S_1\in \{s_{1,m,0},s_{1,m,1}\}$}. \\
\end{cases}
\end{aligned}
$$
In the above case, we can merge $s_{1,1,0},s_{1,1,1}$ into $s_{1,1}$ and $s_{1,0,0},s_{1,0,1}$ into $s_{1,0}$, and the ground-truth trajectories with merged states become:
$$
\begin{aligned}
&s_{0,0}=1+2+3 \Rightarrow s_{1,0}=(3+3,1+5) \Rightarrow s_{2,0}=6 \\
&s_{0,1}=2+3+4 \Rightarrow s_{1,1}=(5+4,2+7) \Rightarrow s_{2,1}=9 \\
\end{aligned}
$$
And the transition probability becomes:
$$
\begin{aligned}
P(S_1|S_{0})=\begin{cases}
    \frac{1+2\delta_2}{2} &\text{if $S_1=s_{1,m}$ and $S_0=s_{0,m}$}\\
    \frac{1+2\delta_2}{2} &\text{if $S_1=s_{1,m}$ and $S_0=s_{0,m'\neq m}$} \\
\end{cases}
\end{aligned}
$$
and 
$$
\begin{aligned}
P(S_2|S_{1})=\begin{cases}
    \frac{1+\delta_1}{2} &\text{if $S_2=s_{2,m}$ and $S_1=s_{1,m}$}\\
    \frac{1+\delta_1}{2} &\text{if $S_2=s_{2,m'\neq m}$ and $S_1=s_{1,m}$} \\
\end{cases}
\end{aligned}
$$
This setting can still be accommodated within our framework. However, when the ground-truth trajectories differ in their number of reasoning steps $N$—for instance, when certain solution paths omit intermediate steps—our current analysis no longer applies directly. Such scenarios are more complex to handle and are left for future work.

% For example, if the ground-truth trajectories are:
% $$
% \begin{aligned}
% &s_{0,0}=1+2+3 \Rightarrow s_{1,0,0}=3+3 \Rightarrow s_{2,0}=6 \\
% &s_{0,0}=1+2+3 \Rightarrow s_{1,0,1}=1+5 \Rightarrow s_{2,0}=6 \\
% &s_{0,0}=1+2+3 \Rightarrow s_{2,0}=6 \\
% &s_{0,1}=2+3+4 \Rightarrow s_{1,1,0}=5+4 \Rightarrow s_{2,1}=9 \\
% &s_{0,1}=2+3+4 \Rightarrow s_{1,0,1}=2+7 \Rightarrow s_{2,1}=9 \\
% &s_{0,1}=2+3+4 \Rightarrow s_{2,1}=9 \\
% \end{aligned}
% $$

% Initially, step 0 comprises $M_0$ states while step $n$ comprises $M_n$ states.  After concatenation, each composite state $(s_{0,m_0},s_{n,m_n})$ yields exactly $M_0$ states at step $n$, so we fix $M = M_0$ for all $n$.  By applying Theorem\~\ref{theorem\:main\_theorem\_delta\_speed} to this constant-$M$ setting, we obtain a convergence rate that we conjecture approximates that of the original varying-$M$ problem.  In Section\~\ref{sec\:experiment\_non\_constant\_M}, we empirically compare this approximated rate with the convergence behaviour observed in the original task.
\section{Conclusion}
We propose RL-STaR, a theoretical framework for analyzing the core properties of the Self-Taught Reasoner (STaR) algorithm. Assuming suitably bootstrapped pre-trained models, we prove that RL-STaR yields policy improvement and converges toward the optimal policy---even when the training data include erroneous reasoning trajectories. Although our analysis abstracts away many complexities of real-world LLM deployments, it offers a rigorous account of why STaR works and lays the groundwork for extending this theory to more realistic settings in future work.

\bibliography{main}
\bibliographystyle{tmlr}

\appendix
\section{Appendix}

\subsection{Additional Notation}
\label{sec:appendix_notation}
The proofs in this appendix use the following notation in addition to that of Sec.~\ref{sec:notations}.
\begin{itemize}[itemsep=1pt, topsep=2pt, parsep=0pt]
\item $(s_i, s_k) \Subset \tau$: the elements $s_i$ and $s_k$ both occur in the tuple $\tau = (s_i, s_j, \dots, s_k, s_l)$, with $s_i$ preceding $s_k$.
\item $\mathbb{I}\{a=b\}$: $\mathbb{I}\{a=b\}=1$ if $a=b$; otherwise $\mathbb{I}\{a=b\}=0$.
\item $\mathbf{1}$: the all-ones vector.
\end{itemize}

% \subsection{Additional Theorems and Corollaries}\label{sec:additional_theorems}
% \begin{corollary}[\textbf{Policy Improvement in the Toy Example}]\label{theorem:toy_policy_improvement}
% Given the toy example defined in Sec.~\ref{sec:toy_example}, let $P_t$ represent the estimated transition of the model in the $t$-th iteration of RL-STaR training. We aim to show that the training process improves the reward $J(P_t)$, namely,
% $$
% J(P_t) \geq J(P_{t-1}).
% $$
% \end{corollary}
% \proof The proof can be found in Sec.~\ref{sec:proof_of_theorem:toy_policy_improvement}. 
% \qedhere 
% \begin{corollary}[\textbf{Convergence to Optimal Policy in the Toy Example}]\label{theorem:toy_optimal_policy}
% Define $P^{\star}$ as the optimal estimated transition, which maximizes the reward $J(P^{\star})$. This maximum is achieved when
% $$
% J(P^{\star}) = \sup_{\delta \in \left(0, \frac{1}{2}\right)} 2\left(\frac{1}{2^2} + \delta^2\right) = \lim_{\delta \rightarrow \frac{1}{2}} 2\left(\frac{1}{2^2} + \delta^2\right) = 1.
% $$
% \end{corollary}
% \proof The proof can be found in Sec.~\ref{sec:proof_of_theorem:toy_optimal_policy}\qedhere

\subsection{Proof of Theorems}
\label{sec:proof_of_theorem}

% \subsubsection{Additional Theorems for Toy Examples}\label{sec:additional_theorem_toy_example}
\subsubsection{Proof of Theorem \ref{theorem:toy2_nter}} 
\label{sec:proof_of_theorem:toy2_nter}
\proof Without loss of generality, we prove the case when $t=1$. 
This is the case of the first iteration of RL-STaR Algorithm. 
First, we illustrate the transition probability of the pre-trained LLM $P_{0,n}$ as shown in Fig.~\ref{fig:supp_pretrained_transition}.
% https://q.uiver.app/#q=WzAsNixbMCwwLCJzX3swLDF9Il0sWzYsMCwic197MSwxfSJdLFsxMiwwLCJzX3syLDF9Il0sWzAsNCwic197MCwyfSJdLFs2LDQsInNfezEsMn0iXSxbMTIsNCwic197MiwyfSJdLFswLDEsInt7e3tQXzAoc197MSwxfXxzX3swLDF9KT1cXGZyYWN7MX17Mn0rXFxkZWx0YV8wfX19fSIsMV0sWzAsNCwie3t7e1BfMChzX3sxLDJ9fHNfezAsMX0pPVxcZnJhY3sxfXsyfS1cXGRlbHRhXzB9fX19IiwxLHsibGFiZWxfcG9zaXRpb24iOjIwfV0sWzEsMiwie3t7e1BfMChzX3syLDF9fHNfezEsMX0pPVxcZnJhY3sxfXsyfStcXGRlbHRhXzB9fX19IiwxXSxbMSw1LCJ7e3t7UF8wKHNfezIsMn18c197MSwxfSk9XFxmcmFjezF9ezJ9LVxcZGVsdGFfMH19fX0iLDEseyJsYWJlbF9wb3NpdGlvbiI6MjB9XSxbMywxLCJ7e3t7UF8wKHNfezEsMX18c197MCwyfSk9XFxmcmFjezF9ezJ9LVxcZGVsdGFfMH19fX0iLDEseyJsYWJlbF9wb3NpdGlvbiI6MjB9XSxbMyw0LCJ7e3t7UF8wKHNfezEsMn18c197MCwyfSk9XFxmcmFjezF9ezJ9K1xcZGVsdGFfMH19fX0iLDFdLFs0LDIsInt7e3tQXzAoc197MiwxfXxzX3sxLDJ9KT1cXGZyYWN7MX17Mn0tXFxkZWx0YV8wfX19fSIsMSx7ImxhYmVsX3Bvc2l0aW9uIjoyMH1dLFs0LDUsInt7e3tQXzAoc197MiwyfXxzX3sxLDJ9KT1cXGZyYWN7MX17Mn0rXFxkZWx0YV8wfX19fSIsMV1d
\begin{figure*}[ht]
\centering
\begin{adjustbox}{max width=\textwidth}
\begin{tikzcd}
	{s_{0,1}} &&&&&& {s_{1,1}} &&&&&& {s_{2,1}} \\
	\\
	\\
	\\
	{s_{0,2}} &&&&&& {s_{1,2}} &&&&&& {s_{2,2}}
    \arrow["{{{{{P_{0,1}(s_{1,1}|s_{0,1})=\frac{1+\delta_{0,1}}{2}}}}}}"{description}, from=1-1, to=1-7]
    \arrow["{{{{{P_{0,1}(s_{1,2}|s_{0,1})=\frac{1-\delta_{0,1}}{2}}}}}}"{description, pos=0.2}, from=1-1, to=5-7]
    \arrow["{{{{{P_{0,2}(s_{2,1}|s_{1,1})=\frac{1+\delta_{0,2}}{2}}}}}}"{description}, from=1-7, to=1-13]
    \arrow["{{{{{P_{0,2}(s_{2,2}|s_{1,1})=\frac{1-\delta_{0,2}}{2}}}}}}"{description, pos=0.2}, from=1-7, to=5-13]
    \arrow["{{{{{P_{0,1}(s_{1,1}|s_{0,2})=\frac{1-\delta_{0,1}}{2}}}}}}"{description, pos=0.2}, from=5-1, to=1-7]
    \arrow["{{{{{P_{0,1}(s_{1,2}|s_{0,2})=\frac{1+\delta_{0,1}}{2}}}}}}"{description}, from=5-1, to=5-7]
    \arrow["{{{{{P_{0,2}(s_{2,1}|s_{1,2})=\frac{1-\delta_{0,2}}{2}}}}}}"{description, pos=0.2}, from=5-7, to=1-13]
    \arrow["{{{{{P_{0,2}(s_{2,2}|s_{1,2})=\frac{1+\delta_{0,2}}{2}}}}}}"{description}, from=5-7, to=5-13]
\end{tikzcd}
\end{adjustbox}
\caption{Transition probabilities of the pre-trained LLM $P_{0,n}$ in the toy example used in the proof of Theorem~\ref{theorem:toy2_nter}.}
\label{fig:supp_pretrained_transition}
\end{figure*}
To begin with, we have an equal probability of selecting either sample $(s_{0,1}, s_{2,1})$ or $(s_{0,2}, s_{2,2})$ from $\mathcal{D}$. Consequently, we obtain the trajectories $\tau$ from the above transition for $m \in \{1,2\}$, with the following probabilities
\begin{equation}\label{eq:toy2_trajectories}
p(\tau)=
\begin{cases}
 \frac{1}{2}\left(\frac{1+\delta_{0,1}}{2}\right)\left(\frac{1+\delta_{0,2}}{2}\right) & \text{if }\tau=(s_{0,1},s_{1,1},s_{2,1}), \\
 \frac{1}{2}\left(\frac{1+\delta_{0,1}}{2}\right)\left(\frac{1-\delta_{0,2}}{2}\right) & \text{if }\tau=(s_{0,1},s_{1,1},s_{2,2}), \\
 \frac{1}{2}\left(\frac{1-\delta_{0,1}}{2}\right)\left(\frac{1-\delta_{0,2}}{2}\right) & \text{if }\tau=(s_{0,1},s_{1,2},s_{2,1}),  \\
 \frac{1}{2}\left(\frac{1-\delta_{0,1}}{2}\right)\left(\frac{1+\delta_{0,2}}{2}\right) & \text{if }\tau=(s_{0,1},s_{1,2},s_{2,2}),  \\
 \frac{1}{2}\left(\frac{1-\delta_{0,1}}{2}\right)\left(\frac{1+\delta_{0,2}}{2}\right) & \text{if }\tau=(s_{0,2},s_{1,1},s_{2,1}),  \\
 \frac{1}{2}\left(\frac{1-\delta_{0,1}}{2}\right)\left(\frac{1-\delta_{0,2}}{2}\right) & \text{if }\tau=(s_{0,2},s_{1,1},s_{2,2}),  \\
 \frac{1}{2}\left(\frac{1+\delta_{0,1}}{2}\right)\left(\frac{1-\delta_{0,2}}{2}\right) & \text{if }\tau=(s_{0,2},s_{1,2},s_{2,1}),  \\
 \frac{1}{2}\left(\frac{1+\delta_{0,1}}{2}\right)\left(\frac{1+\delta_{0,2}}{2}\right) & \text{if }\tau=(s_{0,2},s_{1,2},s_{2,2}),  \\
\end{cases}
\end{equation}
where $m,n\in[2]$ and $m\neq m'$. In the first iteration of the RL-STaR algorithm, the trajectories $\tau$ that satisfy $(s_{0,m}, s_{2,m}) \Subset \tau$ can be exclusively collected in the dataset $\mathcal{D}_1$. Therefore, the conditional probability for $\tau$ such that  $\tau \in \mathcal{D}_1$ is 
% \begin{cases}
%  \frac{\left(\frac{1}{2}+\delta_0\right)^2}{2\left(\frac{1}{2}+\delta_0\right)^2+2\left(\frac{1}{2}-\delta_0\right)^2}          & \text{if }\tau=(s_{0,1},s_{1,1},s_{2,1}) \\
%  \frac{\left(\frac{1}{2}-\delta_0\right)^2}{2\left(\frac{1}{2}+\delta_0\right)^2+2\left(\frac{1}{2}-\delta_0\right)^2}          & \text{if }\tau=(s_{0,1},s_{1,2},s_{2,1})  \\
%  \frac{\left(\frac{1}{2}-\delta_0\right)^2}{2\left(\frac{1}{2}+\delta_0\right)^2+2\left(\frac{1}{2}-\delta_0\right)^2}         & \text{if }\tau=(s_{0,2},s_{1,1},s_{2,2})  \\
%  \frac{\left(\frac{1}{2}+\delta_0\right)^2}{2\left(\frac{1}{2}+\delta_0\right)^2+2\left(\frac{1}{2}-\delta_0\right)^2}           & \text{if }\tau=(s_{0,2},s_{1,2},s_{2,2})  \\
% \end{cases}
% =
\begin{equation}\label{eq:toy2_traj_in_d1}
p(\tau|\tau\in \mathcal{D}_1)=
\begin{cases}
 \frac{(1+\delta_{0,1})(1+\delta_{0,2})}{4(1+\delta_{0,1}\delta_{0,2})} & \text{if }\tau=(s_{0,1},s_{1,1},s_{2,1}), \\
 \frac{(1-\delta_{0,1})(1-\delta_{0,2})}{4(1+\delta_{0,1}\delta_{0,2})} & \text{if }\tau=(s_{0,1},s_{1,2},s_{2,1}),  \\
 \frac{(1-\delta_{0,1})(1-\delta_{0,2})}{4(1+\delta_{0,1}\delta_{0,2})} & \text{if }\tau=(s_{0,2},s_{1,1},s_{2,2}),  \\
 \frac{(1+\delta_{0,1})(1+\delta_{0,2})}{4(1+\delta_{0,1}\delta_{0,2})} & \text{if }\tau=(s_{0,2},s_{1,2},s_{2,2}).  \\
\end{cases}
\end{equation}
Based on this dataset, we assume that the LLMs can perfectly learn the conditional transition $P(S_{n+1,m}|S_{n,m})$ based on the probabilities of $(s_{n,m}, s_{n+1,m}) \Subset \tau$ and
$(s_{n,m},s_{n+1,m'\neq m}) \Subset \tau$ from the $\tau \sim p(\tau|\tau\in \mathcal{D}_1)$. For example, $P(S_{1,1}|s_{0,1})$ can be obtained from
\begin{align*}
&P(s_{1,1}|s_{0,1}) \\
&=\frac{p((s_{0,1},s_{1,1})\Subset\tau|\tau\in \mathcal{D}_1)}{\sum_{j\in[2]} p((s_{0,1},s_{1,j})\Subset\tau|\tau\in \mathcal{D}_1)} \\
&= \frac{\frac{(1+\delta_{0,1})(1+\delta_{0,2})}{4(1+\delta_{0,1}\delta_{0,2})}}{\frac{(1+\delta_{0,1})(1+\delta_{0,2})}{4(1+\delta_{0,1}\delta_{0,2})}+\frac{(1-\delta_{0,1})(1-\delta_{0,2})}{4(1+\delta_{0,1}\delta_{0,2})}} \\
&= \frac{(1+\delta_{0,1})(1+\delta_{0,2})}{2(1+\delta_{0,1}\delta_{0,2})}.
\end{align*}
Hence, the transition $P_1$ is
\[
P_1(S_n|S_{n-1})=
\begin{cases}
 \frac{(1+\delta_{0,1})(1+\delta_{0,2})}{2(1+\delta_{0,1}\delta_{0,2})} \\
 \quad \text{if $S_{n-1}=s_{n-1,m}$, $S_{n}=s_{n,m}$,} \\
 \frac{(1-\delta_{0,1})(1-\delta_{0,2})}{2(1+\delta_{0,1}\delta_{0,2})} \\
 \quad \text{if $S_{n-1}=s_{n-1,m}$, $S_{n}=s_{n,m'\neq m}$,}
\end{cases}
\]
for all $n,m,m'\in[2]$,
which can be illustrated as in Fig.~\ref{fig:supp_iter1_transition}.
\begin{figure*}[ht]
\centering
\begin{adjustbox}{max width=\textwidth}
\begin{tikzcd}
	{s_{0,1}} &&&&&&& {s_{1,1}} &&&&&&& {s_{2,1}} \\
	\\
	\\
	\\
	{s_{0,2}} &&&&&&& {s_{1,2}} &&&&&&& {s_{2,2}}
	\arrow["{{{P_1(s_{1,1}|s_{0,1})= \frac{(1+ \delta_{0,1})(1 + \delta_{0,2})}{2(1 + \delta_{0,1}\delta_{0,2})} }}}"{description}, from=1-1, to=1-8]
	\arrow["{{{P_1(s_{1,2}|s_{0,1})= \frac{(1- \delta_{0,1})(1 - \delta_{0,2})}{2(1 + \delta_{0,1}\delta_{0,2})} }}}"{description, pos=0.7}, from=1-1, to=5-8]
	\arrow["{{{P_1(s_{2,1}|s_{1,1})= \frac{(1+ \delta_{0,1})(1 + \delta_{0,2})}{2(1 + \delta_{0,1}\delta_{0,2})} }}}"{description}, from=1-8, to=1-15]
	\arrow["{{{P_1(s_{2,2}|s_{1,1})=\frac{(1- \delta_{0,1})(1 - \delta_{0,2})}{2(1 + \delta_{0,1}\delta_{0,2})} }}}"{description, pos=0.7}, from=1-8, to=5-15]
	\arrow["{{{P_1(s_{1,1}|s_{0,2})= \frac{(1- \delta_{0,1})(1 - \delta_{0,2})}{2(1 + \delta_{0,1}\delta_{0,2})} }}}"{description, pos=0.7}, from=5-1, to=1-8]
	\arrow["{{{P_1(s_{1,2}|s_{0,2})= \frac{(1+ \delta_{0,1})(1 + \delta_{0,2})}{2(1 + \delta_{0,1}\delta_{0,2})} }}}"{description}, from=5-1, to=5-8]
	\arrow["{{{P_1(s_{2,2}|s_{1,2})=\frac{(1- \delta_{0,1})(1 - \delta_{0,2})}{2(1 + \delta_{0,1}\delta_{0,2})} }}}"{description, pos=0.7}, from=5-8, to=1-15]
	\arrow["{{{P_1(s_{2,1}|s_{1,2})= \frac{(1+ \delta_{0,1})(1 + \delta_{0,2})}{2(1 + \delta_{0,1}\delta_{0,2})} }}}"{description}, from=5-8, to=5-15]
\end{tikzcd}
\end{adjustbox}
\caption{Transition probabilities $P_1$ after the first iteration of RL-STaR in the toy example used in the proof of Theorem~\ref{theorem:toy2_nter}.}
\label{fig:supp_iter1_transition}
\end{figure*}
If RL-STaR improves the transition probabilities in each iteration, then the probabilities of transitions matching the ground-truth trajectories will increase. Specifically, we have $P_1(S_n = s_{n, m} | S_{n-1} = s_{n-1, m}) > P_1(S_n = s_{n, m'\neq m} | S_{n-1} = s_{n-1, m})$. Now we need to prove that 
$$
\delta_{t-1, m} < \delta_{t, m} <1.
%,\quad\text{and}\quad \delta_{t, m}=\frac{\left(\frac{\delta_{0, m}^{-1} +2}{\delta_{0, m}^{-1} -2}\right)^{2^t} -1}{2\left(1 + \left(\frac{\delta_{0, m}^{-1} +2}{\delta_{0, m}^{-1} -2}\right)^{2^t}\right)} \quad\text{$\forall  m \in [2]$}.
$$
We can get  $\delta_{1,1}=\frac{\delta_{0,1}+\delta_{0,2}}{1+\delta_{0,1}\delta_{0,2}}$ by
\[
\adjustbox{max width=\linewidth}{$\displaystyle
\begin{aligned}
\frac{(1+ \delta_{0,1})(1+ \delta_{0,2})}{2(1 + \delta_{0,1}\delta_{0,2})}
&= \frac{(1 + \delta_{0,1}\delta_{0,2}) + \big( (1+ \delta_{0,1})(1 + \delta_{0,2}) - (1+ \delta_{0,1}\delta_{0,2}) \big)}{2(1+ \delta_{0,1}\delta_{0,2})} \\
&= \frac{1}{2} + \frac{(1+ \delta_{0,1})(1 + \delta_{0,2}) - (1+ \delta_{0,1}\delta_{0,2})}{2(1+ \delta_{0,1}\delta_{0,2})} \\
&= \frac{1}{2} + \frac{ \delta_{0,1} + \delta_{0,2}}{2(1 + \delta_{0,1}\delta_{0,2})} = \frac{1}{2} + \frac{\delta_{1,1}}{2}.
\end{aligned}
$}
\]
We can also show that $\delta_{1,1} < 1$, since
\begin{align*}
\delta_{1,1} - 1 &= \frac{ \delta_{0,1} + \delta_{0,2}}{1 + \delta_{0,1}\delta_{0,2}} -1 \\
&=  \frac{( \delta_{0,1} + \delta_{0,2}) - (1 + \delta_{0,1}\delta_{0,2})}{1 + \delta_{0,1}\delta_{0,2}} \\
&=  \frac{-( \delta_{0,1} - 1)( \delta_{0,2} - 1)}{1 + \delta_{0,1}\delta_{0,2}} < 0.
\end{align*}

% Note that $$
% \begin{aligned}
% \frac{(\frac{1}{2} + \delta_0)^2}{2(\frac{1}{2^2} + \delta_0^2)} &= \frac{(\frac{1}{2^2} + \delta_0^2) + \left( (\frac{1}{2} + \delta_0)^2 - (\frac{1}{2^2} + \delta_0^2) \right)}{2(\frac{1}{2^2} + \delta_0^2)} \\
% &= \frac{1}{2} + \frac{(\frac{1}{2} + \delta_0)^2 - (\frac{1}{2^2} + \delta_0^2)}{2(\frac{1}{2^2} + \delta_0^2)} \\
% &= \frac{1}{2} + \frac{\delta_0}{2(\frac{1}{2^2} + \delta_0^2)} = \frac{1}{2} + \delta_1.
% \end{aligned}
% $$
Besides, it is straightforward that the above analysis is true if we replace $\delta_{0,1}$ by $\delta_{t-1,1}$, or replace $\delta_{t-1,1}$ by $\delta_{t-1,2}$.
%This concludes the proof of $\delta_{t-1, 1}<\delta_{t,1}<1$. By the symmetric property, this also gives us  $\delta_{t-1, 2}<\delta_{t,2}<1$.

Now we consider the three conditions on the values of $\delta_{0,1}$ and $\delta_{0,2}$ as follows. 
\begin{enumerate}[label=(\alph*)]
\item If both $\delta_{0,1} > 0$ and $\delta_{0,2}>0$,  we claim \(\delta_{1,1} - \delta_{0,1} > 0\). Indeed,
\[
\begin{aligned}
\delta_{1,1} - \delta_{0,1}
&= \frac{\delta_{0,1} + \delta_{0,2}}{1 + \delta_{0,1}\,\delta_{0,2}} \;-\; \delta_{0,1} \\
&= \frac{\delta_{0,1} + \delta_{0,2}}{1 + \delta_{0,1}\,\delta_{0,2}}
  \;-\;
  \frac{\delta_{0,1}\bigl(1 + \delta_{0,1}\,\delta_{0,2}\bigr)}{1 + \delta_{0,1}\,\delta_{0,2}} \\
&= \frac{\delta_{0,2}\,\bigl(1 - \delta_{0,1}^2\bigr)}{1 + \delta_{0,1}\,\delta_{0,2}}
\;>\;0,
\end{aligned}
\]
because \(\delta_{0,1}, \delta_{0,2} > 0\) and \(1 - \delta_{0,1}^2 > 0\) (note that \(\delta_{0,1} < 1\)).  
A similar argument holds for \(\delta_{1,2} - \delta_{0,2}\), ensuring \(\delta_{1,2} > \delta_{0,2}\) whenever  \(\delta_{0,1} > 0\) and \(\delta_{0,2} > 0\).
\item If exactly one of \(\delta_{0,1}\) or \(\delta_{0,2}\) is zero, 
we write \(\delta_{0,n'} > 0\) and \(\delta_{0,n} = 0\). Then,
\[
\delta_{1,1} \;=\;\delta_{1,2}\;=\;\frac{0 + \delta_{0,n'}}{0+1} \;=\;\delta_{0,n'} \;>\;0.
\]
Thus, starting from \(t=1\), both \(\delta_{1,1}\) and \(\delta_{1,2}\) are strictly positive, reducing this scenario to the first case above for all subsequent iterations that $\delta_{t,1}>0$ and $\delta_{t,2}>0$.
\item If both \(\delta_{0,1}=0\) and \(\delta_{0,2}=0\),    then
\[
\delta_{t,1} \;=\; \delta_{t,2} 
\;=\; \frac{0 + 0}{0 + 1} 
\;=\; 0,
\]
for all \(t > 0\). In other words, once both \(\delta_{0,1}\) and \(\delta_{0,2}\) are zero, they remain zero for every iteration \(t\).
% $$
% \delta_1 = \frac{\delta_0}{2(\frac{1}{2^2} + \delta_0^2)} > \frac{\delta_0}{\frac{1}{2} + 2(\frac{1}{2})^2} = \delta_0,
% $$
\end{enumerate}

\qedhere

\subsubsection{Proof of Theorem \ref{theorem:toy2_convergence_speed}}
\label{sec:proof_of_theorem:toy2_convergence_speed}
\label{proof_for_theorem:toy2_convergence_speed}
\proof
Let $f(x) = \frac{2x}{1 + x^2}$. Using Taylor's theorem to expand about $x = 1$, one has
\[  
f(\delta_{t}) - 1 = f'(1)(\delta_t - 1) + \frac{f''(1)}{2!}(\delta_t-1)^2 + \frac{f^{(3)}(\xi_t)}{3!}(\delta_t-1)^3,
\]
for each $t \geq 0$ and some $\xi_t \in (\delta_t, 1)$. It's straightforward to see that $f'(1) = 0 $, $f''(1) = -1$ and $f^{(3)}(x) =-\frac{12 \left(x^{4} - 6x^{2} + 1\right)}{\left(x^{2} + 1\right)^{4}}$. Hence,
\[
\lim_{t \to \infty}\frac{|\delta_{t+1} - 1|}{|\delta_t - 1|^2} = 0.5,
\]
which shows quadratic convergence.\\
To find its closed-form expression, put $x_t = \delta_t^{-1}$ and rewrite our recurrence as $ x_{t+1} = ({1 + x_t^2})/{2x_t}.$ Observe that
\begin{align*}
    &x_{t+1} +1 = \frac{(x_t + 1)^2}{2x_t}, \quad \text{and} \\
    &x_{t+1} -1 = \frac{(x_t - 1)^2}{2x_t}.
\end{align*}
One now sees that
\[\frac{x_t +1}{x_t -1} = \left(\frac{x_0 +1}{x_0 -1}\right)^{2^t}.\]
Directly solving for $x_t$ and $\delta_t$, we obtain
$$
x_t = \frac{\left(\frac{x_0+1}{x_0-1}\right)^{2^t}+1}{\left(\frac{x_0+1}{x_0-1}\right)^{2^t}- 1} \quad{\text{and}}\quad \delta_t = \frac{\left(\frac{\delta_0^{-1}+1}{\delta_0^{-1}-1}\right)^{2^t} -1}{ \left(\frac{\delta_0^{-1}+1}{\delta_0^{-1}-1}\right)^{2^t}+1}. 
$$
\qedhere

\subsubsection{Proof of Theorem \ref{theorem:toy2_policy_improvement}}
\label{sec:proof_of_theorem:toy2_policy_improvement}
\label{proof_of_theorem:toy2_policy_improvement}
\proof
From Eq.~\eqref{eq:toy2_trajectories}, the reward \(J(P_0)\) is the probability that \(p(\tau)\) satisfies \((s_{0,m}, s_{2,m}) \Subset \tau\). Hence,
\begin{align*}
J(P_0)
&=
\left(\frac{1 + \delta_{0,1}}{2}\right)\left(\frac{1 + \delta_{0,2}}{2}\right) \\
&\quad +
\left(\frac{1 - \delta_{0,1}}{2}\right)\left(\frac{1 - \delta_{0,2}}{2}\right) \\
&=
\frac{1 + \delta_{0,1}\,\delta_{0,2}}{2}.
\end{align*}
This expression remains valid for all \(t \ge 0\). Moreover, since $\delta_t = \delta_{t,1}=\delta_{t,2}$ when $t\geq 1$, it is obvious that \(J(P_t)\) is an increasing function. Therefore,
$$
J(P_t) 
\;=\; 
\frac{1 + \delta_{t}^2}{2} 
\;>\; 
\frac{1 + \delta_{t-1}^2}{2}
\;=\; 
J(P_{t-1}),
\quad\text{ for all $t\geq 1$},
$$
which completes the proof.
\qedhere

\subsubsection{Proof of Theorem \ref{theorem:toy_optimal_policy}}
\label{sec:proof_of_theorem:toy_optimal_policy}
\label{proof_of_theorem:toy_optimal_policy}
\proof We need to show that, for any $0 < \delta_t < 1$, the limit of $\delta_t$ approaches $1$ as $t$ tends to infinity. Since $0<\delta_t < 1$, we have $(\delta_0^{-1} +1)\big/(\delta_0^{-1} -1) > 1$. It is straightforward that by applying Theorem~\ref{theorem:toy2_convergence_speed}, we have
$$
\lim_{t \rightarrow \infty} \delta_t  =\frac{\left(\frac{\delta_0^{-1}+1}{\delta_0^{-1}-1}\right)^{2^t} -1}{ \left(\frac{\delta_0^{-1}+1}{\delta_0^{-1}-1}\right)^{2^t}+1} =1 .
$$
\qedhere

\subsubsection{Proof of Theorem \ref{theorem:toy2_diminish_reasonong_path}}
\label{sec:proof_of_theorem:toy2_diminish_reasonong_path}
\label{proof_of_theorem:toy2_diminish_reasonong_path}
\proof
Note that $\delta_t = \delta_{t,1}=\delta_{t,2}$ when $t \geq 1$. Apply Eq.\eqref{eq:toy2_traj_in_d1} and we have
\begin{equation}
p(\tau' \in \mathcal{D}_t)=
 \frac{(1-\delta_{t})^2}{4(1+\delta_{t}^2)},
\end{equation}
if $\tau=(s_{0,1},s_{1,2},s_{2,1})$ or $\tau=(s_{0,2},s_{1,1},s_{2,2})$.
We complete the proof by applying Theorem~\ref{theorem:toy2_convergence_speed}.
\qedhere

% \subsection{Proof of Theorems for Toy example}

\subsubsection{Propositions for Proving the Main Theorem}
%\subsubsection{Self-Taught Reasoner}
\begin{proposition}\label{proposition:epsilon_decreasing}
Consider a non-homogeneous Markov sequence $S=\left(S_n\right)_{n=0}^{N}$ with state space $\mathcal{S}=[M]$, initial distribution $\mu$, and transition matrices $P_n \in \mathbb{R}^{M \times M}$. More specifically, for each $n \in [N]$
$$
S_0 \sim \mu, \quad\left[P_n\right]_{i, j}=P\left[S_{n}=j \mid S_{n-1}=i\right].
$$
Denote by $\mathcal{C}=\left\{\left(s_n\right)_{n=0}^{N} \in \mathcal{S}^{N+1}: {s_0=s_{N}}\right\}$ the set of correct trajectories, let $\mu_i=\mu(\{i\})$, and
\[
\adjustbox{max width=\linewidth}{$\displaystyle
\begin{aligned}
{\big[\tilde{P}_n\big]_{i, j} }
& =P\left[S_{n}=j \mid S \in \mathcal{C}, S_{n-1}=i\right] \\
& =\frac{\sum_{m=1}^M P\left[S_0=S_{N}=m, S_{n-1}=i, S_{n}=j\right]}{\sum_{m=1}^M P\left[S_0=S_{N}=m, S_{n-1}=i\right]} \\
& =\frac{\sum_{m=1}^M \mu_m\left[\prod_{k=1}^{n-1} P_k\right]_{m, i}\left[P_n\right]_{i, j}\left[\prod_{k=n+1}^{N} P_k\right]_{j, m}}{\sum_{m=1}^M \mu_m\left[\prod_{k=1}^{n-1} P_k\right]_{m, i}\left[\prod_{k=n}^{N} P_k\right]_{i, m}} .
\end{aligned}
$}
\]
Suppose $\mu$ is uniform and that $\big[P_n\big]_{i, j}=\alpha\left(\delta_n\right) \mathbb{I}\{i=j\}+\beta\left(\delta_n\right)  \mathbb{I}\{i \neq$ $j\}$ with $\delta_n \in[0,1]$ for each $n \in [N]$, where
$$
\alpha(\delta)=\frac{1+(M-1) \delta}{M}, \beta(\delta)=\frac{1-\delta}{M} .
$$
Then $\big[\tilde{P}_n\big]_{i, j}=\alpha(\tilde{\delta}_n) \mathbb{I}\{i=j\}+\beta(\tilde{\delta}_n) \mathbb{I}\{i \neq j\}$ satisfies
\begin{equation}\label{eq:relation_epsilon}
\tilde{\epsilon}_n=\frac{1-\prod_{k \neq n} \delta_k}{1+(M-1) \prod_{k=1}^{N} \delta_k} \epsilon_n,
\end{equation}
where $\tilde{\epsilon}_n=1-\tilde{\delta}_n$ and $\epsilon_n=1-\delta_n$.
% Furthermore, denote $\vartheta=1-\prod_{k=1}^{N} \delta_k \leq 1 \wedge \sum_{k=1}^{N} \epsilon_k$, one has
% \begin{equation*}
% \sum_{k=1}^{N} \tilde{\epsilon}_k \leq \frac{\vartheta}{M-(M-1) \vartheta} \sum_{k=1}^{N} \epsilon_k .
% \end{equation*}
\end{proposition}
\proof Let $\left\{u_m\right\}_{m=1}^M$ be an orthonormal basis of $\mathbb{R}^M$ where $u_1=\frac{1}{\sqrt{M}} \mathbf{1}$. Then
\begin{align*}
P_n &=\delta_n I_{M}+\tfrac{1-\delta_n}{M} \mathbf{1}\mathbf{1}^{\top} \\
&=\delta_n \sum_{m=1}^M u_m u_m^{\top}+\left(1-\delta_n\right) u_1 u_1^{\top} \\
&=u_1 u_1^{\top}+\delta_n \sum_{m=2}^M u_m u_m^{\top}=U \Lambda_n U^{\top},
\end{align*}
where $U=\left[u_1 \cdots u_M\right]$ and $\Lambda_n=\operatorname{diag}\left(1, \delta_n, \cdots, \delta_n\right)$. Denote $\delta_{k: l}=\prod_{t=k}^l \delta_t$ and $\Lambda_{k: l}=$ $\prod_{t=k}^l \Lambda_t=\delta_{k: l} I_M+\left(1-\delta_{k: l}\right) e_1 e_1^{\top}$. One has
\begin{equation}\label{eq:pn_tilde_ij}
\adjustbox{max width=\linewidth}{$\displaystyle
\begin{aligned}
& {\big[\tilde{P}_n\big]_{i, j}=\left[P_n\right]_{i, j} \frac{\sum_{m=1}^M \mu_m e_m^{\top} U \Lambda_{1: n-1} U^{\top} e_i e_j^{\top} U \Lambda_{n+1: N} U^{\top} e_m}{\sum_{m=1}^M \mu_m e_m^{\top} U \Lambda_{1: n-1} U^{\top} e_i e_i^{\top} U \Lambda_{n: N} U^{\top} e_m}} \\
& =\left[P_n\right]_{i, j} \frac{\sum_{m=1}^M e_m^{\top}\left(\delta_{1: n-1} I_M+\left(1-\delta_{1: n-1}\right) u_1 u_1^{\top}\right) e_i e_j^{\top}\left(\delta_{n+1: N} I_M+\left(1-\delta_{n+1: N}\right) u_1 u_1^{\top}\right) e_m}{\sum_{m=1}^M e_m^{\top}\left(\delta_{1: n-1} I_M+\left(1-\delta_{1: n-1}\right) u_1 u_1^{\top}\right) e_i e_i^{\top}\left(\delta_{n: N} I_M+\left(1-\delta_{n: N}\right) u_1 u_1^{\top}\right) e_m} \\
& =\left[P_n\right]_{i, j} \frac{\sum_{m=1}^M\left(\frac{1-\delta_{1: n-1}}{M}+\delta_{1: n-1} e_m^{\top} e_i\right)\left(\frac{1-\delta_{n+1: N}}{M}+\delta_{n+1: N} e_j^{\top} e_m\right)}{\sum_{m=1}^M\left(\frac{1-\delta_{1: n-1}}{M}+\delta_{1: n-1} e_m^{\top} e_i\right)\left(\frac{1-\delta_{n: N}}{M}+\delta_{n: N} e_i^{\top} e_m\right)}.
\end{aligned}
$}
\end{equation}
Note that
\[
\adjustbox{max width=\linewidth}{$\displaystyle
\big[\tilde{P}_n\big]_{i, j}=\beta\left(\delta_n\right) \frac{(M-2) \beta\left(\delta_{1: n-1}\right) \beta\left(\delta_{n+1: N}\right)+\beta\left(\delta_{1: n-1}\right) \alpha\left(\delta_{n+1: N}\right)+\alpha\left(\delta_{1: n-1}\right) \beta\left(\delta_{n+1: N}\right)}{(M-1) \beta\left(\delta_{1: n-1}\right) \beta\left(\delta_{n: N}\right)+\alpha\left(\delta_{1: n-1}\right) \alpha\left(\delta_{n, N}\right)},
$}
\]
are identical for all $i \neq j$. Hence there uniquely exists a $\tilde{\delta}_n \in[0,1]$ such that $\big[\tilde{P}_n\big]_{i, j}=\beta\big(\tilde{\delta}_n\big)$ and $\big[\tilde{P}_n\big]_{i, i}=\alpha\big(\tilde{\delta}_n\big)$ for all $i \neq j$. Moreover, (cf. \eqref{eq:pn_tilde_ij})
\[
\adjustbox{max width=\linewidth}{$\displaystyle
\begin{aligned}
\alpha\big(\tilde{\delta}_n\big) & =\alpha\left(\delta_n\right) \frac{(M-1) \beta\left(\delta_{1: n-1}\right) \beta\left(\delta_{n+1: N}\right)+\alpha\left(\delta_{1: n-1}\right) \alpha\left(\delta_{n+1, N}\right)}{(M-1) \beta\left(\delta_{1: n-1}\right) \beta\left(\delta_{n: N}\right)+\alpha\left(\delta_{1: n-1}\right) \alpha\left(\delta_{n, N}\right)} \\
& =\alpha\left(\delta_n\right) \frac{\left(1-\alpha\left(\delta_{1: n-1}\right)\right) \beta\left(\delta_{n+1: N}\right)+\alpha\left(\delta_{1: n-1}\right) \alpha\left(\delta_{n+1, N}\right)}{\left(1-\alpha\left(\delta_{1: n-1}\right)\right) \beta\left(\delta_{n: N}\right)+\alpha\left(\delta_{1: n-1}\right) \alpha\left(\delta_{n, N}\right)} \\
& =\alpha\left(\delta_n\right) \frac{\beta\left(\delta_{n+1: N}\right)+\alpha\left(\delta_{1: n-1}\right) \delta_{n+1: N}}{\beta\left(\delta_{n: N}\right)+\alpha\left(\delta_{1: n-1}\right) \delta_{n: N}} \\
& =\alpha\left(\delta_n\right) \frac{1+(M-1) \delta_{1: n-1} \delta_{n+1: N}}{1+(M-1) \delta_{1: N}}.
\end{aligned}
$}
\]

Hence,
\[
\adjustbox{max width=\linewidth}{$\displaystyle
\begin{aligned}
& (M-1) \beta\big(\tilde{\delta}_n\big)=1-\left(1-(M-1) \beta\left(\delta_n\right)\right) \frac{1+(M-1) \delta_{1: n-1} \delta_{n+1: N}}{1+(M-1) \delta_{1: N}} \\
& =\frac{(M-1)\left(\delta_{1: N}-\delta_{1: n-1} \delta_{n+1: N}\right)+(M-1) \beta\left(\delta_n\right)\left(1+(M-1) \delta_{1: n-1} \delta_{n+1: N}\right)}{1+(M-1) \delta_{1: N}} \\
& =\frac{(M-1)\left(-M \beta\left(\delta_n\right) \delta_{1: n-1} \delta_{n+1: N}\right)+(M-1) \beta\left(\delta_n\right)\left(1+(M-1) \delta_{1: n-1} \delta_{n+1: N}\right)}{1+(M-1) \delta_{1: N}} \\
& =(M-1) \beta\left(\delta_n\right) \frac{1-\delta_{1: n-1} \delta_{n+1: N}}{1+(M-1) \delta_{1: N}} .
\end{aligned}
$}
\]

Note that $\epsilon_n=M \beta\left(\delta_n\right)$ and $\tilde{\epsilon}_n=M \beta\big(\tilde{\delta}_n\big)$ yield Eq.~\eqref{eq:relation_epsilon}.
\qedhere

\begin{proposition}\label{proposition:convergence_speed}
Consider the scenario defined in Proposition~\ref{proposition:epsilon_decreasing}. Let $\left(\theta_{t}\right)_{t=0}^{\infty},\left(\vartheta_{t}\right)_{t=0}^{\infty}$ be two non-negative non-increasing sequences satisfying $\vartheta_t \leq \theta_t$,  $\vartheta_{0} \in (0,1)$, and
\[\theta_{t+1} \leq \frac{\vartheta_t}{M - (M-1)\vartheta_t}\theta_t.\] We have the following:
% Denote $\gamma=\frac{\vartheta_{0}}{M-(M-1) \vartheta_{0}} \in(0,1)$, then
\begin{enumerate}[label=(\alph*)]
\item Denote $\theta=\sum_{k=1}^{N} (1- \delta_k)$ and $\vartheta=1-\prod_{k=1}^{N} \delta_k$. Then, $\vartheta \leq \theta$.
\item %Define $\theta_{t} = \sum_{k=1}^{N} \epsilon_{t,k}$ and $\vartheta_{t} = 1 - \prod_{k=1}^N \delta_{t, k}$.% %Let 
Following $(a)$, if we further assume $\delta_n > 0$ for all $n$, we have
$$
    \sum_{k=1}^N \tilde{\epsilon}_k \leq \frac{\vartheta \theta}{M - (M-1)\vartheta}.
%\theta_{t+1} \leq \frac{\vartheta_{t}}{M-(M-1) \vartheta_{t}} \theta_{t}.%
$$
\item  Denote $\gamma=\frac{\vartheta_{0}}{M-(M-1) \vartheta_{0}} \in(0,1)$. Then $\theta_{t} \leq \gamma^{t} \theta_{0}$. That is, for $\varepsilon>0$, it holds that $\theta_{t} \leq \varepsilon$ whenever $t \geq \frac{\log \left(\theta_{0} / \varepsilon\right)}{\log (1 / \gamma)} \vee 0$.
\item  If $\theta_{0}<\frac{M}{M+1}$, then
\[
\theta_{t} \leq \frac{M\left(\theta_{0}\right)^{2^{t}}}{M\left(\theta_{0}\right)^{2^{t}}+\left(M\left(1-\theta_{0}\right)\right)^{2^{t}}} .
\]

Furthermore, for $\varepsilon \in\left(0, \frac{M}{M+1}\right)$, it holds that $\theta_{t} \leq \varepsilon$ whenever
\[
t \geq \operatorname{max}\left(\log _{2}\left(\frac{\log M+\log \frac{1-\varepsilon}{\varepsilon}}{\log M+\log \frac{1-\theta_{0}}{\theta_{0}}}\right),\, 0\right).
\]
\item  For $\varepsilon \in\left(0, \frac{M}{M+1}\right)$, we have $\theta_{t} \leq \varepsilon$ whenever
\[
\adjustbox{max width=\linewidth}{$\displaystyle
t \geq\left\lceil\frac{\log \frac{M+1}{M \gamma}+\log \left(\theta_{0}\right)}{\log (1 / \gamma)}\right\rceil_{+}+\left\lceil\log _{2}\left(\frac{\log M+\log \frac{1-\varepsilon}{\varepsilon}}{\log \frac{(M+1)-M \gamma}{\gamma}}\right)\right\rceil_{+} .
$}
\]
\end{enumerate}
\end{proposition}
\proof \mbox{}
\begin{enumerate}[label=(\alph*)]
\item Because \(0 \le \delta_k \le 1\) for all \(k\in[N]\), we can use a probability space \((\Omega, \mathcal{F}, \mu)\) to show \(\vartheta \le \theta\). Let \(\{E_1, E_2, \dots, E_N\} \subset \mathcal{F}\) be events with \(\mu(E_k) = \delta_k\). Then the complements \(E_k^c\) satisfy \(\mu(E_k^c) = 1 - \delta_k\). We can define
\[
\vartheta 
= 
\mu\!\Bigl(\bigcup_{k=1}^{N} E_k^c\Bigr)
\quad\text{and}\quad
\theta 
= 
\sum_{k=1}^{N} \mu(E_k^c).
\]
If \(E_1^c, E_2^c, \dots, E_N^c\) are disjoint, \(\vartheta = \theta\) by countable additivity. In general, \(E_k^c\) may overlap, so 
\[
\mu\Bigl(\bigcup_{k=1}^N E_k^c\Bigr)
\leq
\sum_{k=1}^N \mu(E_k^c),
\]
implying \(\vartheta \leq \theta\).
\item From the proof of Proposition~\ref{proposition:epsilon_decreasing}, we can derive
\[
\adjustbox{max width=\linewidth}{$\displaystyle
\sum_{k=1}^{N} \tilde{\epsilon}_k=\frac{\theta-\delta_{1: N} \sum_{k=1}^{N} \frac{\epsilon_k}{1-\epsilon_k}}{1+(M-1) \delta_{1: N}} \leq \frac{\theta-(1-\vartheta) \frac{N \theta}{N-\theta}}{M-(M-1) \vartheta}=\theta \frac{(N \vartheta-\theta) /(N-\theta)}{M-(M-1) \vartheta} \leq \frac{\vartheta \theta}{M-(M-1) \vartheta},
$}
\]
where the inequalities follow by noting that $\vartheta \leq \theta$ and
\[
\adjustbox{max width=\linewidth}{$\displaystyle
\begin{aligned}
& \theta^2=\left(\sum_{k=1}^{N} \frac{\epsilon_k}{\sqrt{1-\epsilon_k}} \cdot \sqrt{1-\epsilon_k}\right)^2 \leq\left(\sum_{k=1}^{N} \frac{\epsilon_k^2}{1-\epsilon_k}\right)\left(\sum_{k=1}^{N}\left(1-\epsilon_k\right)\right)=(N-\theta) \sum_{k=1}^{N} \frac{\epsilon_k^2}{1-\epsilon_k} \\
\Rightarrow & \sum_{k=1}^{N} \frac{\epsilon_k}{1-\epsilon_k}=\sum_{k=1}^{N}\left(\epsilon_k+\frac{\epsilon_k^2}{1-\epsilon_k}\right) \geq \theta+\frac{\theta^2}{N-\theta}=\frac{N \theta}{N-\theta} .
\end{aligned}
$}
\]
\item Since $\vartheta_{t} \leq \vartheta_{0}$, it is clear that $\theta_{t+1} \leq \gamma \theta_{t}$ and the rest is trivial.
\item Denote $f: x \in[0,1] \mapsto \frac{x^{2}}{M-(M-1) x}$ and $g: x \in[0,1) \mapsto\left(\varphi \circ h \circ \varphi^{-1}\right)(x)$ where
\[
\adjustbox{max width=\linewidth}{$\displaystyle
\varphi: x \in[0, \infty) \mapsto \frac{M x}{M x+1}, \quad h: x \in[0, \infty) \mapsto x^{2}, \quad \varphi^{-1}: y \in[0,1) \mapsto \frac{y}{M(1-y)} .
$}
\]
Note that
\[
\adjustbox{max width=\linewidth}{$\displaystyle
g(x)=\frac{M\left(\frac{x}{M(1-x)}\right)^{2}}{M\left(\frac{x}{M(1-x)}\right)^{2}+1}=\frac{x^{2}}{M-2 M x+(M+1) x^{2}} \geq \frac{x^{2}}{M-2 M x+(M+1) x}=f(x),
$}
\]
for all $x \in[0,1)$. Since $\vartheta_{t} \leq \theta_{t}<1$, one has $\theta_{t+1} \leq f\left(\theta_{t}\right) \leq g\left(\theta_{t}\right)$, which further implies $\theta_{t} \leq g^{(t)}\left(\theta_{0}\right)$ due to $g$ being monotonically increasing. Thus
\[
\theta_{t} \leq g^{(t)}\left(\theta_{0}\right)=\left(\varphi \circ h^{(t)} \circ \varphi^{-1}\right)\left(\theta_{0}\right)=\frac{M\left(\frac{\theta_{0}}{M\left(1-\theta_{0}\right)}\right)^{2^{t}}}{M\left(\frac{\theta_{0}}{M\left(1-\theta_{0}\right)}\right)^{2^{t}}+1} .
\]
We complete the proof by noting that (provided $t \geq 0$)
\[
\adjustbox{max width=\linewidth}{$\displaystyle
\begin{aligned}
g^{(t)}\left(\theta_{0}\right) \leq \varepsilon & \Leftrightarrow\left(\varphi \circ h^{(t)} \circ \varphi^{-1}\right)\left(\theta_{0}\right) \leq \varepsilon \Leftrightarrow\left(\varphi^{-1}\left(\theta_{0}\right)\right)^{2^{t}}=\left(h^{(t)} \circ \varphi^{-1}\right)\left(\theta_{0}\right) \leq \varphi^{-1}(\varepsilon) \\
& \Leftrightarrow 2^{t} \log \left(1 / \varphi^{-1}\left(\theta_{0}\right)\right) \geq \log \left(1 / \varphi^{-1}(\varepsilon)\right) \Leftrightarrow t \geq \log _{2}\left(\frac{\log \left(1 / \varphi^{-1}(\varepsilon)\right)}{\log \left(1 / \varphi^{-1}\left(\theta_{0}\right)\right)}\right) .
\end{aligned}
$}
\]
\item Take
\[
\adjustbox{max width=\linewidth}{$\displaystyle
\tau_{1}=\left\lceil\frac{\log \frac{M+1}{M \gamma}+\log \left(\theta_{0}\right)}{\log (1 / \gamma)}\right\rceil_{+}\!\!,~ \tau_{2}=\left\lceil\log _{2}\left(\frac{\log M+\log \frac{1-\varepsilon}{\varepsilon}}{\log \frac{M+1-M \gamma}{\gamma}}\right)\right\rceil_{+}\!\!.
$}
\]
Then (a) implies $\theta_{\tau_{1}} \leq \frac{M \gamma}{M+1}$, and (b) further implies $\theta_{\tau_{1}+\tau_{2}} \leq \varepsilon$.
\end{enumerate}
\qedhere

\subsubsection{Proof of Theorem~\ref{theorem:main_theorem}}
\label{sec:proof_of_theorem:main_theorem}
\proof The proof of this theorem is based primarily on Proposition~\ref{proposition:epsilon_decreasing}.
Without loss of generality, we assume that \(s_{0,m} = s_{1,m} = \cdots = s_{N,m} = m\) for every \(m \in [M]\) and only consider the first iteration of RL-STaR (i.e., $t=1$ and $t-1=0$). We omit subscripts $(0,n)$ and simply write $P_{0,n} = P_n$, $\delta_{0,n} = \delta_n$, and $S_{0,n} = S_n$, etc. Similarly, let $P_{1,n} = \tilde{P}_n$, $\delta_{1,n} = \tilde{\delta}_n$, \(\tilde{\epsilon}_n = 1 - \delta_{1,n}\), and \(\epsilon_n = 1 - \delta_{n}\), etc. Under this formulation, the scenario given in Theorem~\ref{theorem:main_theorem} can be seen as the Markov process introduced in Proposition~\ref{proposition:epsilon_decreasing}. Applying Eq.~\eqref{eq:relation_epsilon}, we have
\begin{equation}\label{eq:delta_relations_t}
\delta_{t,n}
=\frac{(M-2)\prod_{k=1}^{N} \delta_{t-1,k}+\prod_{k\neq n}\delta_{t-1,k}+\delta_{t-1,n}}{1+(M-1)\prod_{k=1}^{N} \delta_{t-1,k}}.
\end{equation}
We now examine the three cases with respect to the values of $\delta_{t=1,n}$.
\begin{enumerate}[label=(\alph*)]
\item If \(0 < \delta_{0,n} <1 \) for all \(n \in [N]\), it is obvious that $0 < \tilde{\epsilon}_n < \epsilon_n$ and hence
\[
\delta_{0,n} < \delta_{1,n} < 1.
\]
It is straightforward that this inequality will hold for every subsequent iteration $t > 1$.
% \subsubsection{Proof of Theorem~\ref{theorem:main_theorem_boundary_conditions}}
% \label{sec:proof_of_theorem:main_theorem_boundary_conditions}
% \proof We apply Eq~\eqref{eq:delta_relations_t} to discuss these two cases.
    \item Suppose there exists an $n\in[N]$ such that $\delta_{0,n}=0$ and for any other $n'\neq n,~n'\in[N]$ we have $\delta_{0,n'}>0$. Then, $\prod_{k=1}^{N} \delta_{0,k}=0$,   $\prod_{n'\neq n}\delta_{0,n'} >0$  and $\prod_{k\neq n'}\delta_{0,k} =0$ for any $k\neq n', n'\neq n, ~k,n'\in[N]$. One now sees that
\begin{align*}
\delta_{1,n}&=\frac{0+\prod_{n'\neq n}\delta_{0,n'}+0}{1+0}=\prod_{n'\neq n}\delta_{0,n'} > 0, \\
\delta_{1,n'}&=\frac{0+0+\delta_{0,n'}}{1+0}=\delta_{0,n'} > 0.
\end{align*}
After the first iteration, apply the analysis in  $(a)$.
    \item Suppose there exist two (or more than two) steps $n,n'\in[N]$ satisfying $n'\neq n$, $\delta_{0,n}=0$  and $\delta_{0,n'}=0$.  We have $\prod_{k=1}^{N} \delta_{0,k}=0$,    and $\prod_{k\neq n}\delta_{0,k} =0$ for any $k\in[N]$. Then
$$
\delta_{1,n}=\frac{0+0+\delta_{0,n}}{1+0}=\delta_{0,n} \quad \text{for all $n \in [N]$}.
$$
It is obvious that above equation holds for every subsequent iteration $t > 1$.
\end{enumerate}
\qedhere

\subsubsection{Proof of Theorem~\ref{theorem:main_theorem_delta_speed}}
\label{sec:proof_of_theorem:main_theorem_delta_speed}
\proof For the case of $\delta_{0,n}>0$ for all $n\in[N]$, we can apply Proposition~\ref{proposition:convergence_speed}, where we set 
$$
\gamma=\frac{\vartheta_{0}}{M-(M-1) \vartheta_{0}} =\frac{1-\prod_{k=1}^{N}\delta_{0,k}}{(M-1)\prod_{k=1}^{N}\delta_{0,k}+1},
$$
and let $\varepsilon=\frac{\varepsilon'}{N}$ for $\varepsilon' \in\left(0, \frac{M}{M+1}\right)$. Therefore,
\[
\max_{k} \delta_{t,k} \geq
\frac{\sum_{k=1}^{N}\delta_{t,k}}{N} \geq 1 - \frac{\varepsilon' }{N}
\]
holds for all
\[
\adjustbox{max width=\linewidth}{$\displaystyle
t \geq\left\lceil\frac{\log \frac{M+1}{M \gamma}+\log \left(\theta_{0}\right)}{\log (1 / \gamma)}\right\rceil_{+}+\left\lceil\log _{2}\left(\frac{\log M+\log \frac{1-\varepsilon'}{\varepsilon'}}{\log \frac{(M+1)-M \gamma}{\gamma}}\right)\right\rceil_{+} .
$}
\]
On the other hand, if there exists only one $n\in [N]$ such that \(\delta_{0,n} = 0\),  
an additional iteration $t=1$ is needed to ensure \(\delta_{1,n} > 0\). Once \(\delta_{1,n}\) becomes strictly positive for all $n\in [N]$, we set
$$
\gamma = \frac{1-\prod_{k=1}^{N}\delta_{1,k}}{(M-1)\prod_{k=1}^{N}\delta_{1,k}+1},
$$
and then the argument proceeds as before, completing the proof.
\qedhere

% \subsubsection{Proof of Theorem~\ref{theorem:main_theorem_boundary_conditions}}
% \label{sec:proof_of_theorem:main_theorem_boundary_conditions}
% \proof We apply Eq~\eqref{eq:delta_relations_t} to discuss these two cases.
% \begin{enumerate}[label=(\alph*)]
%     \item There exists $n\in[N]$ satisfying $\delta_{t-1,n}=0$ and for any other $n'\neq n,~n'\in[N]$ satisfying $\delta_{t-1,n}>0$. We have $\prod_{k=1}^{N} \delta_{t-1,k}=0$,   $\prod_{n'\neq n}\delta_{t-1,n'} >0$  and $\prod_{k\neq n'}\delta_{t-1,k} =0$ for any $k\neq n', n'\neq n, ~k,n'\in[N]$, then
% \begin{align*}
% \delta_{t,n}&=\frac{(M-2)\prod_{k=1}^{N} \delta_{t-1,k}+\prod_{n'\neq n}\delta_{t-1,n'}+\delta_{t-1,n}}{1+(M-1)\prod_{k=1}^{N} \delta_{t-1,k}}=\frac{0+\prod_{n'\neq n}\delta_{t-1,n'}+0}{1+0}=\prod_{n'\neq n}\delta_{t-1,n'} > 0, \quad\text{and} \\
% \delta_{t,n'}&=\frac{(M-2)\prod_{k=1}^{N} \delta_{t-1,k}+\prod_{k\neq n'}\delta_{t-1,k}+\delta_{t-1,n'}}{1+(M-1)\prod_{k=1}^{N} \delta_{t-1,k}}=\frac{0+0+\delta_{t-1,n'}}{1+0}=\delta_{t-1,n'} > 0. 
% \end{align*}
%     \item If there exist tw steps $n,n'\in[N]$ satisfying $n'\neq n$, $\delta_{t-1,n}=0$  and $\delta_{t-1,n'}=0$.  We have $\prod_{k=1}^{N} \delta_{t-1,k}=0$,    and $\prod_{k\neq n}\delta_{t-1,k} =0$ for any $k\in[N]$, then
% $$
% \delta_{t,n}=\frac{(M-2)\prod_{k=1}^{N} \delta_{t-1,k}+\prod_{k\neq n}\delta_{t-1,k}+\delta_{t-1,n}}{1+(M-1)\prod_{k=1}^{N} \delta_{t-1,k}}=\frac{0+0+\delta_{t-1,n}}{1+0}=\delta_{t-1,n} \quad \text{for all $n \in [N]$}.
% $$
% \end{enumerate}
% \qedhere

\subsubsection{Proof of Corollary~\ref{theorem:main_theorem_policy_improvement}}
\label{sec:proof_of_theorem:main_theorem_policy_improvement}
% \todo{Proof}
\proof By applying Proposition~\ref{proposition:epsilon_decreasing} to the $t$-th iteration of RL-STaR, we abuse the notation of $P_{t,n}$ to denote the transition matrix in the $t$-th iteration of RL-STaR, at the $n$-th CoT step. Recall that
\[
\adjustbox{max width=\linewidth}{$\displaystyle
\prod_{k=1}^{N}P_{t,k} = U \left(\prod_{k=1}^{N}\delta_{t,k} I_M + \big(1- \prod_{k=1}^{N}\delta_{t,k}\big)e_1 e_1^\top \right)U^T,
$}
\]
where $U = [u_1, u_2, \dots, u_M]$, and $u_1 = \frac{1}{\sqrt{M}}\mathbf{1}$. One sees that
\[
\adjustbox{max width=\linewidth}{$\displaystyle
\big[\prod_{k=1}^{N}P_{t,k}\big]_{i,i}= \big[\prod_{k=1}^{N}P_{(t,k)}\big]_{1,1} = \prod_{k=1}^{N}\delta_{t,k}+\frac{1 - \prod_{k=1}^{N}\delta_{t,k}}{M} .
$}
\]
Since the initial state is chosen uniformly,
\[
\adjustbox{max width=\linewidth}{$\displaystyle
J(P_t) = \sum_{m=1}^{M}\frac{1}{M}\big[\prod_{k=1}^{N}P_{(t,k)}\big]_{m,m} = \frac{1}{M} + \tfrac{M-1}{M}\prod_{k=1}^{N}\delta_{t,k}.
$}
\]
Assuming Theorem~\ref{theorem:main_theorem}~\ref{theorem:item_d1} or \ref{theorem:item_d2} hold, we conclude as $\delta_{1,n} \geq \delta_{0,n}$ and $\delta_{t,n} > \delta_{t-1,n}$ for all $t > 1$ and $n \in [N]$.\qedhere

\subsubsection{Proof of Corollary~\ref{theorem:main_theorem_optimal_policy}}\label{sec:proof_of_theorem:main_theorem_optimal_policy}
\proof
Theorem~\ref{theorem:main_theorem_delta_speed} implies that 
$$
\lim_{t\to\infty}\delta_{t,n}=1 \quad \text{for all $n \in [N]$}.
$$ 
By Proposition~\ref{proposition:epsilon_decreasing}, we know that $P_{t,n}$ has diagonal elements $\alpha(\delta_{t,n})$ and off-diagonal elements $\beta(\delta_{t,n})$ such that
%With Sec.~\ref{sec:proof_of_theorem:main_theorem_policy_improvement}, by the definition of %$\alpha_{t,n}$ and $\beta_{t,n}$,
\begin{align*}
\lim_{t\to\infty}\alpha(\delta_{t,n}) &= \lim_{t\to\infty}\frac{1+(M-1)\delta_{t,n}}{M}=1, \\
\lim_{t\to\infty}\beta(\delta_{t,n}) &= \lim_{t\to\infty}\frac{1-\delta_{t,n}}{M}  =0,
\end{align*}
for all $n \in [N]$.
%By Proposition~\ref{proposition:epsilon_decreasing}, we know that $P_{t,n}$ has diagonal elements %$\alpha_{t,n}$ and non-diagonal $\beta_{t,n}$ , and it is obvious that
Every row of $P_{t,n}-I_{M}$ has one entry equal to $\alpha(\delta_{t,n})-1$ and $M-1$ entries equal to $\beta(\delta_{t,n})$, so
$\|P_{t,n}-I_{M}\|_{\infty}=\left(1-\alpha(\delta_{t,n})\right)+(M-1)\beta(\delta_{t,n})$. Hence, 
$$
\lim_{t\rightarrow \infty}\| P_{t,n}-I_{M} \|_{\infty}=0 \quad \text{for all} \quad n \in [N].
$$\qed

\subsubsection{Proof of Corollary~\ref{theorem:main_theorem_incorrect_reasoning_step}}\label{sec:proof_of_theorem:main_theorem_incorrect_reasoning_step}
% \todo{}
\proof This is a simple consequence of convergence to the optimal policy. Since the incorrect reasoning steps may appear in any CoT steps, we see that
\[p(\tau_{t,k}) = \frac{1}{M}\left(\beta(\delta_{t,n_1}) \cdots (\beta(\delta_{t,n_k})\right)\prod_{j \neq n_i \forall i}\alpha(\delta_{t,j})\]
where $2 \leq k\leq N$, for some subsequence $n_1 < n_2 < \cdots < n_k$. Using $\lim_{t \to \infty}\alpha(\delta_{t,n}) = 1$ and $\lim_{t \to \infty}\beta(\delta_{t,n}) = 0$ for each $n$, we have
$0 \leq p(\tau_{t,k}) \xrightarrow{t \to \infty} 0$. Because $\vert\bigcup_{k}\tau_{t,k}\vert < \infty$, we obtain the desired result. 
\qedhere

\subsection{Toy Example without the Markov Assumption}
\label{sec:non_markov_toy_example}
In the toy example of Sec.~\ref{sec:toy_example}, let \(P_{0,n}\) denote the pre-trained LLM’s estimated transition at step \(n\), which fails to satisfy the Markov property. Specifically, we assume  
\begin{align*}
&P_{0,1}(S_1 | S_0 ) =
\begin{cases}
\frac{1+ \delta_{0,n}}{2}  & \text{if } (S_1,S_0) = (s_{1, m},s_{0, m}) \text{ for all } n, m \in [2], \\
\frac{1- \delta_{0,n} }{2} & \text{otherwise. } 
\end{cases}
\end{align*}
And that
\begin{align*}
&P_{0,2}(S_2 | S_1, S_0 ) =
\begin{cases}
\frac{(1+ \delta_{0,1})^2}{4} &\text{if } (S_2,S_1,S_0) 
= (s_{2,m},s_{1, m},s_{0, m}) \text{ for all }  m \in [2],\\
\frac{(1+ \delta_{0,1})(1-\delta_{0,2}) }{4} &\text{if } (S_2,S_1,S_0) = (s_{2,m'\neq m},s_{1, m},s_{0, m}) \text{ for all }  m',m \in [2], \\
\frac{(1- \delta_{0,1})(1+ \delta_{0,2})}{4} &\text{if } (S_2,S_1,S_0) 
 = (s_{2,m},s_{1, m},s_{0, m'\neq m}) \text{ for all } m', m \in [2],\\
 \frac{(1- \delta_{0,1})^2 }{2}&\text{if } (S_2,S_1,S_0)   = (s_{2,m},s_{1, m'\neq m},s_{0, m})\text{ for all }  m',m \in [2].
\end{cases}
\end{align*}
To derive \(P_1(S_1\mid S_0)\) and \(P_2(S_2\mid S_1, S_0)\), we follow the same procedure as in the proof of Theorem~\ref{theorem:toy2_nter} in~\ref{sec:proof_of_theorem:toy2_nter}. In the inner loop \(\ell=1,\dots,L\) of Algorithm~\ref{algo:rl_star}, we sample trajectories \(\tau\), and
each occurring with probability \(p(\tau)\) as
$$
p(\tau)=
\begin{cases}
 .125\left({1+\delta_{0,1}}\right)\left({1+\delta_{0,2}}\right) & \text{if }\tau=(s_{0,1},s_{1,1},s_{2,1}), \\
 .125\left({1+\delta_{0,1}}\right)\left({1-\delta_{0,2}}\right) & \text{if }\tau=(s_{0,1},s_{1,1},s_{2,2}), \\
 .125\left({1-\delta_{0,1}}\right)\left({1-\delta_{0,2}}\right) & \text{if }\tau=(s_{0,1},s_{1,2},s_{2,1}),  \\
 .125\left({1-\delta_{0,1}}\right)\left({1+\delta_{0,2}}\right) & \text{if }\tau=(s_{0,1},s_{1,2},s_{2,2}),  \\
 .125\left({1-\delta_{0,1}}\right)\left({1+\delta_{0,2}}\right) & \text{if }\tau=(s_{0,2},s_{1,1},s_{2,1}),  \\
 .125\left({1-\delta_{0,1}}\right)\left({1-\delta_{0,2}}\right) & \text{if }\tau=(s_{0,2},s_{1,1},s_{2,2}),  \\
 .125\left({1+\delta_{0,1}}\right)\left({1-\delta_{0,2}}\right) & \text{if }\tau=(s_{0,2},s_{1,2},s_{2,1}),  \\
 .125\left({1+\delta_{0,1}}\right)\left({1+\delta_{0,2}}\right) & \text{if }\tau=(s_{0,2},s_{1,2},s_{2,2}).  \\
\end{cases}
$$
Conditioned on \(\tau\in\mathcal{D}_1\), the sampling probability is 
$
p(\tau\mid\tau\in\mathcal{D}_1)
$
as 
$$
p(\tau|\tau\in \mathcal{D}_1)=
\begin{cases}
 \frac{(1+\delta_{0,1})(1+\delta_{0,2})}{4(1+\delta_{0,1}\delta_{0,2})} & \text{if }\tau=(s_{0,1},s_{1,1},s_{2,1}), \\
 \frac{(1-\delta_{0,1})(1-\delta_{0,2})}{4(1+\delta_{0,1}\delta_{0,2})} & \text{if }\tau=(s_{0,1},s_{1,2},s_{2,1}),  \\
 \frac{(1-\delta_{0,1})(1-\delta_{0,2})}{4(1+\delta_{0,1}\delta_{0,2})} & \text{if }\tau=(s_{0,2},s_{1,1},s_{2,2}),  \\
 \frac{(1+\delta_{0,1})(1+\delta_{0,2})}{4(1+\delta_{0,1}\delta_{0,2})} & \text{if }\tau=(s_{0,2},s_{1,2},s_{2,2}).  \\
\end{cases}
$$
If the LLM perfectly learns these conditional transitions from \(\tau\sim p(\tau\mid\tau\in\mathcal{D}_1)\), then
$$
\begin{aligned}
&P_{1,1}(S_1|S_0) =\begin{cases}
 \frac{(1+\delta_{0,1})(1+\delta_{0,2})}{2(1+\delta_{0,1}\delta_{0,2})} &\text{if $(S_1,S_0)=(s_{1,m},s_{0,m})$ for all $m \in[2]$}, \\
 \frac{(1-\delta_{0,1})(1-\delta_{0,2})}{2(1+\delta_{0,1}\delta_{0,2})} &\text{if $(S_1,S_0)=(s_{1,m'\neq m},s_{0,m})$ for all $m',m \in[2]$}.
\end{cases}\\
\end{aligned}
$$
And that
$$
\begin{aligned}
&P_{1,2}(S_2|S_1,S_0) =
\begin{cases}
 1 & \text{if $(S_2,S_1,S_0)\in$} \text{$\left\{(s_{2,m},s_{1,m},s_{0,m}),(s_{2,m},s_{1,m'\neq m},s_{0,m})\right\}$} \text{ for all $m',m \in[2]$}, \\
 0 
 & \text{if $(S_2,S_1,S_0)\in$}\text{$\left\{(s_{2,m\neq m'},s_{1,m},s_{0,m}),(s_{2,m},s_{1, m},s_{0,m'\neq m})\right\}$} \text{ for all $m',m \in[2]$}. \\
\end{cases}
\end{aligned}
$$
Based on these equations, since \(P_{1,2}\) is deterministic, repeating the outer loop of Algorithm~\ref{algo:rl_star} for \(T\) iterations yields
$\mathcal{D}_T =\mathcal{D}_{T-1}=\cdots=\mathcal{D}_1$.
As a result, the trajectories 
$\tau = (s_{0,m},s_{1,m'\neq m},s_{2,m})$, which contain an incorrect reasoning step, remain in the dataset and cannot be removed by RL-STaR.

\end{document}